\documentclass[11pt]{article}
\usepackage{array}
\newcolumntype{P}[1]{>{\centering\arraybackslash}p{#1}}
\usepackage[a4paper, total={6in, 8in}]{geometry}
\usepackage[T1]{fontenc}
\usepackage{graphicx}
\usepackage{multirow}
\usepackage{booktabs}
\usepackage{amsmath}
\usepackage{booktabs,caption}
\usepackage[flushleft]{threeparttable}
\usepackage[displaymath]{lineno}
\usepackage{setspace}
\usepackage{comment}
\usepackage{hyperref}
\usepackage{cleveref}
\usepackage{placeins}
\usepackage[font=small,labelfont=bf]{caption} 
\usepackage[subrefformat=parens]{subcaption}
\usepackage{soul}
\usepackage{color}
\usepackage{amsthm}
\usepackage{mathtools}
\usepackage{pdflscape}
\usepackage{rotating}

\begin{document}

\title{\textbf{Equality of opportunity in travel behavior prediction with deep neural networks and discrete choice models}}

\author{Yunhan Zheng \\
  Shenhao Wang\footnote{Corresponding Author; Email: shenhao@mit.edu}\\
  Jinhua Zhao \\
  \small{Massachusetts Institute of Technology} \\
  \small{77 Mass Ave, Cambridge, Massachusetts, U.S.} \\
  \\}
\date{}              

\renewcommand{\thefootnote}{\fnsymbol{footnote}}
\singlespacing
\maketitle

\vspace{-.2in}
\begin{abstract}
\noindent
Although researchers increasingly adopt machine learning to model travel behavior, they predominantly focus on prediction accuracy, ignoring the ethical challenges embedded in machine learning algorithms. This study introduces an important missing dimension - computational fairness - to travel behavior analysis. It highlights the accuracy-fairness tradeoff instead of the single dimensional focus on prediction accuracy in the contexts of deep neural network (DNN) and discrete choice models (DCM). We first operationalize computational fairness by \textit{equality of opportunity}, then differentiate between the bias inherent in data and the bias introduced by modeling. The models inheriting the inherent biases can risk perpetuating the existing inequality in the data structure, and the biases in modeling can further exacerbate it. We then demonstrate the prediction disparities in travel behavior modeling using the 2017 National Household Travel Survey (NHTS) and the 2018-2019 My Daily Travel Survey in Chicago. Empirically, DNN and DCM reveal consistent prediction disparities across multiple social groups: both over-predict the false negative rate of frequent driving for the ethnic minorities, the low-income and the disabled populations, and falsely predict a higher travel burden of the socially disadvantaged groups and the rural populations than reality. Comparing DNN with DCM, we find that DNN can outperform DCM in prediction disparities because of DNN's smaller misspecification error. To mitigate prediction disparities, this study introduces an absolute correlation regularization method, which is evaluated with synthetic and real-world data. The results demonstrate the prevalence of prediction disparities in travel behavior modeling, and the disparities still persist regarding a variety of model specifics such as the number of DNN layers, batch size and weight initialization. Since these prediction disparities can exacerbate social inequity if prediction results without fairness adjustment are used for transportation policy making, we advocate for careful consideration of the fairness problem in travel behavior modeling, and the use of bias mitigation algorithms for fair transport decisions.

\hfill\break%
\end{abstract}
\medskip

\thispagestyle{empty}
\clearpage

\onehalfspacing
\setcounter{footnote}{0}
\renewcommand{\thefootnote}{\arabic{footnote}}
\setcounter{page}{1}

\section{Introduction}
In recent years, a growing body of literature has adopted machine learning models, particularly deep neural networks (DNNs), to predict travel behavior. The common practice of machine learning is to identify the best model by fitting the training data and being evaluated on the test data, with the objectives of performance optimization and output prediction in various scenarios \cite{paredes2017machine,cheng2019applying}. Comparing DNN with traditional logit models, previous studies have shown that DNN has higher predictive power and typically makes fewer mistakes in predictions compared to multinomial logit models (MNL) \cite{fernandez2014we,karlaftis2011statistical}. Specifically, DNN is powerful owing to factors such as the relaxation of linear relationships among variables \cite{wang2020deep}, automatic feature learning \cite{wang2020deep} and the ability to accommodate various data structures \cite{krizhevsky2012imagenet,goodfellow2016deep}.\\ 

\noindent However, machine learning also poses tremendous ethical challenges. Many studies have found that machine learning models can produce much worse prediction results for disadvantaged groups such as black people, women and low-income populations, leading to unfair treatment of these populations. For example, software based on machine learning algorithms to predict future criminals is biased against people of color \cite{Angwin16}.
Research on online advertisement showed that ads (e.g. PeopleSmart ads, public records ads) for public records on a person appear disproportionately higher for black-identifying names \cite{Sweeney13}. Literature concerning text classification demonstrated gender bias in word embeddings trained on Google News, which systematically associates men with computer programmers and women with homemakers \cite{bolukbasi2016man}. Although the fairness problem in machine learning has been revealed across a large number of contexts, thus far no study has examined the computational fairness issue in travel behavior modeling. In fact, fairness has been an enduring topic in the transportation field \cite{Ramjerdi2006,Litman2002,Martens11}. The traditional transport fairness research either adopts a highly qualitative approach or presents quantitative metrics without integrating fairness into algorithms. In this computational era, it is critical to demonstrate the risks of naively adopting models without fairness concerns and showcase the integration of fairness metrics into modeling practices and policy decisions. \\ 


\noindent Motivated by these research gaps, this research investigates how to measure, evaluate, and mitigate prediction disparities of travel behavior models regarding protected status - race, gender, income, medical condition and region. 
We take the following three steps. First, we introduce \textit{equality of opportunity} to measure computational fairness in travel behavioral modeling based on the fairness theory in machine learning research. Second, we identify and measure prediction disparities in travel behavior modeling using binary logistic regression (BLR) and DNN. Third, building upon the approach by Beutel et al. \cite{Alex19}, we adopt an absolute correlation regularization method to mitigate the prediction biases, and evaluate the performance of the new model with bias mitigation. The second and the third steps are deployed on both synthetic and real-world datasets. Experiments conducted on the synthetic datasets show how the prediction disparity varies with data structure, number of predictors, sample size and the degree of model misspecification. The fairness analysis is then deployed on the 2017 National Household Travel Survey (NHTS) dataset which has wide coverage of geographic areas and populations with different characteristics, as well as the large sample size and the diversity of input features. It is also deployed on a regional dataset---the 2018-19 Chicago travel survey data. Our analyses show that the prediction disparity issues are prevalent in travel behavior predictions, and our findings are robust to change of hyperparameters such as the number of DNN layers, batch size and weight initialization. Also, our analyses on the nation-wide travel data (NHTS) and the regional data (Chicago travel survey) suggest similar prediction disparity issues and both demonstrate the effectiveness of our bias mitigation approach. The code for our experiments is available online at: \url{https://github.com/zhengyunhan/Equality-of-Opportunity-Travel-Behavior.git}.\\

\noindent Prediction disparities could bias transport policy decisions unfavorably against socially disadvantaged groups, such as low-income and ethnic minority populations. For example, when the estimate of African-American communities' demand for transit has a higher error rate, transit agencies would make more mistakes when considering adding bus routes and investing in new transit stations for minority neighborhoods. Echoing Title VI, the Department of Transportation regulations in 49 CFR Part 21 are designed to ensure that "no person in the United States, based on race, color, or national origin, is excluded from participation in, denied the benefits of, or otherwise subjected to discrimination under any program that DOT financially assists" \cite{dot-2019}. Therefore, people from different groups of interest (such as race, gender, income) should be treated fairly by the models and algorithms which have been widely deployed to inform transportation project planning and policy making. \\

\noindent The paper is organized as follows: the next section reviews travel behavior modeling and fairness in machine learning. Section 3 introduces data and models, where we introduce methods for measuring and mitigating prediction disparities, and also illustrate how the methods are tied to fairness theories in machine learning. Section 4 presents the results, which include the simulation experiments on the synthetic data, the quantification of prediction disparity across various dependent variables and protected variables in the NHTS data and the Chicago travel survey data, as well as the results of bias mitigation for both synthetic and real-world data. Section 5 and 6 summarize the key findings and discuss the study limitations and future research directions.

\section{Literature Review}
Discrete choice modeling has been used in numerous travel behavior studies. With the development of artificial intelligence, machine learning techniques such as DNN have become increasingly popular and been adopted widely to achieve higher prediction accuracy \cite{Cui18,Wang20,Allahviranloo13}. Researchers adopted DNN to predict travel mode choices \cite{wang2020deep,cantarella2005multilayer}, car ownership \cite{paredes2017machine}, traffic accidents \cite{zhang2018deep}, travellers' decision rules \cite{van2019artificial}, driving behaviors \cite{huang2018car}, traffic flows \cite{polson2017deep,wu2018hybrid} and many other decisions in the transportation field. Some researchers focused on comparing the performance of various machine learning models such as DCM, DNN, SVM, and random forest on travel behavior prediction \cite{Cheng19,wang2021deep,Hichem15,Matthew11,Sekhar16}. However, nearly all of these studies evaluate the performance of different models in terms of accuracy, stability and interpretability. \\

\noindent The previous machine learning literature has presented a great body of evidence showing that a model can act discriminatorily on one population in a variety of settings including, but not limited to, criminal risk assessment \cite{Angwin16,chouldechova2017fair}, clinical care \cite{Rajkomar18,goodman2018machine}, online advertisement delivery \cite{datta2015automated,Sweeney13}, text classification \cite{Lucas18,bolukbasi2016man}  and credit risk evaluation \cite{deku2016access,lee2019context}. Approaches to understand and address unfairness in machine learning models include formalizations of fairness in machine learning \cite{DBLP:journals/corr/abs-1808-00023,DBLP:journals/corr/abs-1710-03184}, designing fairness-enhancing algorithms \cite{friedler_comparative_2019,6137441,zhang_mitigating_2018} and solving fairness concerns in real-world industries \cite{kenneth_holstein_improving_2019,gunduz_machine_2019}. In contrast to the rich literature in computer science focusing on fairness analysis, the authors cannot identify any study that has investigated the fairness issue in travel behavior prediction. \\

\noindent On the other hand, fairness has long been a crucial concern in transportation planning, and there has been numerous transport equity analysis \cite{Bills2017,Golub2010,Linovski18,Delbosc11}. Among the variety of equity-focused transportation studies, most carried out fairness discussions in terms how transportation infrastructure, mobility services, opportunities and rights are distributed among the population \cite{Ramjerdi2006,Litman2002}. Research in this domain usually involves the cost and benefit analysis for specific groups or individuals related to a specific transportation project \cite{Litman2002}. The analysis of cost may include transportation cost \cite{Martens11} and environmental cost \cite{Schweitzer2004}, such as the air and noise pollution produced by transportation-related construction. The benefit analysis focuses on the benefits people receive in terms of accessibility, mobility and economic vitality \cite{Bills2017,Martens12}. In these studies, fairness is usually evaluated based on the distributions of costs and benefits among different demographics, neglecting the fact that the machine-learned models deployed to estimate travel demand - which is critical for cost and benefit analysis - itself can be unfair \cite{barocas2016big,DBLP:journals/corr/abs-1710-03184}. Therefore, instead of focusing on substantive fairness which emphasizes resource allocation and decision making, we take a step back and examine bias exhibited in the model estimation results. As such, this study aims to enrich the fairness discussion by focusing on machine-learned models, investigating the fairness issue in the modeling process and analyzing the biases that arise in the modeling results. \\

\noindent Among a variety of fairness definitions, this paper uses equality of opportunity to define fairness. Equality of opportunity is a relaxation of the fairness measure equality of odds, and equality of odds states that the protected and unprotected groups should have equal rates for true positives and false positives \cite{mehrabi2019survey,Hardt16}. Since achieving equal rates for both measures (true positives and false positives) in practice is usually hard, equality of opportunity is adopted instead, stating that protected and unprotected groups should have equal true positive rates \cite{mehrabi2019survey,Hardt16}. A similar fairness definition -  ``reciprocal equality of opportunity'' - is also adopted in this study, which requires that the protected and unprotected groups have equal true negative rates \cite{beutel2017data}. \\

\noindent Equality of opportunity is a type of ``disparate impact'' analysis which evaluates fairness based on model impact (results) - specifically whether policies or practices have a disproportionately adverse impact on protected classes \cite{barocas2016big}. This fairness notion is chosen as it is inherently connected with the notion of equality of opportunity in the traditional transportation equity literature. In the traditional transport equity literature, equality of opportunity focuses on applications and resource allocations. It asserts that the education, employment and consumer opportunities accessible to residents should be equal between different groups \cite{Litman2002}. Violation of equality of opportunity in travel behavior predictions can consequently affect the transportation resources accessible by different populations, thus perpetuating inequality of opportunity in reality. \\

\noindent In addition to equality of opportunity, another widely-adopted fairness measure focusing on disparate impact is ``demographic parity'' \cite{mehrabi2019survey,Hardt16}, which is achieved when the likelihood of a positive outcome is the same regardless of whether the person is in the protected group. For example, when studying gender fairness in predicting the usage of public transit, demographic parity is achieved when the proportion of people predicted as using public transit frequently is the same between male and female. Equality of opportunity is preferred to demographic parity since the latter fails to account for discrimination which is explainable in terms of legitimate grounds \cite{binns2018fairness}.\\

\noindent Apart from disparate impact analysis, another strand of research - called ``disparate treatment'' analysis - evaluates fairness in terms of treatment rather than modeling result to see if the decisions are made (partly) based on the subject’s sensitive attribute information \cite{pmlr-v54-zafar17a}. This type of fairness includes ``fairness through unawareness'' and ``counterfactual fairness''. In the definition of fairness through unawareness, an algorithm is fair as long as any protected attributes are not explicitly used in the decision-making process \cite{grgic2016case,dwork2012fairness}. On the contrary, counterfactual fairness deems a predictor to be fair if its output remains the same when the protected attribute is flipped to its counterfactual value \cite{DBLP:journals/corr/abs-1710-03184,kusner2017counterfactual}. Disparate treatment emphasizes explicit formal classification and intentional discrimination. Therefore, in many machine learning modeling cases where there is no discriminatory intent, disparate impact doctrine is more suited to analyzing unintended biases in data mining compared with disparate treatment doctrine.\\

\noindent The various fairness definitions can also be categorized based on whether they are individual-focused or group-focused. Group fairness, such as equality of opportunities, requires a fair model to treat different groups equally, whereas individual fairness refers to the rule that deems a predictor fair if it produces similar outputs for similar individuals \cite{DBLP:journals/corr/abs-1710-03184,john2020verifying}. Though we use equality of opportunities as the fairness definition in this research, the above-mentioned fairness definitions can be adopted for future studies in this area.


\section{Data and Methods}
\subsection{Equality of opportunity as the definition of fairness}
This study measures fairness by equality of opportunity, mathematically denoted as $P(\hat{y}=1|z=0, y=1)=P(\hat{y}=1|z=1, y=1)$, in which $y$ represents the binary travel behavioral outcomes, $\hat{y}$ represents the predicted values, $z$ represents the protected variable such as race and gender. Intuitively, equality of opportunity requires predicted travel behavior to be conditionally independent of the protected attributes given that the real outcome is positive \cite{Hardt16}. Taking racial disparity as an example, equality of opportunity implies that the predicted travel demand is independent of the travelers being in minority or majority groups, thus achieving a socially non-discriminatory predictive performance \cite{Christina18}. A related concept is referred to as reciprocal equality of opportunity, denoted as $P(\hat{y}=1|z=0, y=0)=P(\hat{y}=1|z=1, y=0)$, implying that the predicted travel behavioral outcome is conditionally independent of the protected attributes given that the real outcome is negative \cite{beutel2017data}.

\subsection{Data and Variables}
In this study, numerical experiments are conducted on two datasets: a group of synthetic datasets and the 2017 National Household Travel Survey data. For each of these two datasets, both BLR and DNN are deployed for model estimations. The experiments on the synthetic data are used to show how the covariance between a protected variable and an explanatory variable may lead to disparate results, and how this prediction disparity varies with the covariance of these two variables, the sample size, and the number of input variables. We then test our bias mitigation method on the synthetic data. For the NHTS data, we examine the prediction disparity for a series of protected (sensitive) variables (e.g. race, gender, income, medical conditions and urban-rural divide) and different dependent variables. Our bias mitigation method is later tested on this real-world dataset as well.

\subsection{Models and Bias Measurement}
In this study, we adopt two models, BLR and DNN, for model predictions, which are later evaluated by fairness metrics for different demographics.

\subsubsection{Binary Logistic Regression (BLR)}
As a classic travel behavior modeling method, BLR has been widely deployed to predict the probability of a certain outcome. The outcome probability is defined as follows:
\setlength{\abovedisplayskip}{10pt}
\setlength{\belowdisplayskip}{10pt}\begin{linenomath}
\begin{gather*}
    P(y_{i}=1|\boldsymbol{x_{i}}) = \frac{1}{1 + exp(-(\alpha+\boldsymbol{x_{i}^\top\beta}))}
\end{gather*}
\end{linenomath}
where $y_{i}$ identifies the dependent variable for individual $i$, $\boldsymbol{x_{i}}$ represents the vector of all the independent variables, $\alpha$ is the intercept, $\boldsymbol{\beta}$ is the vector of parameters associated with attribute $\boldsymbol{x_{i}}$, which is estimated by the negative log likelihood loss function.
\subsubsection{Deep Neural Network Modeling (DNN)}
\begin{figure}[t!]
\centering
\includegraphics[width=5 in]{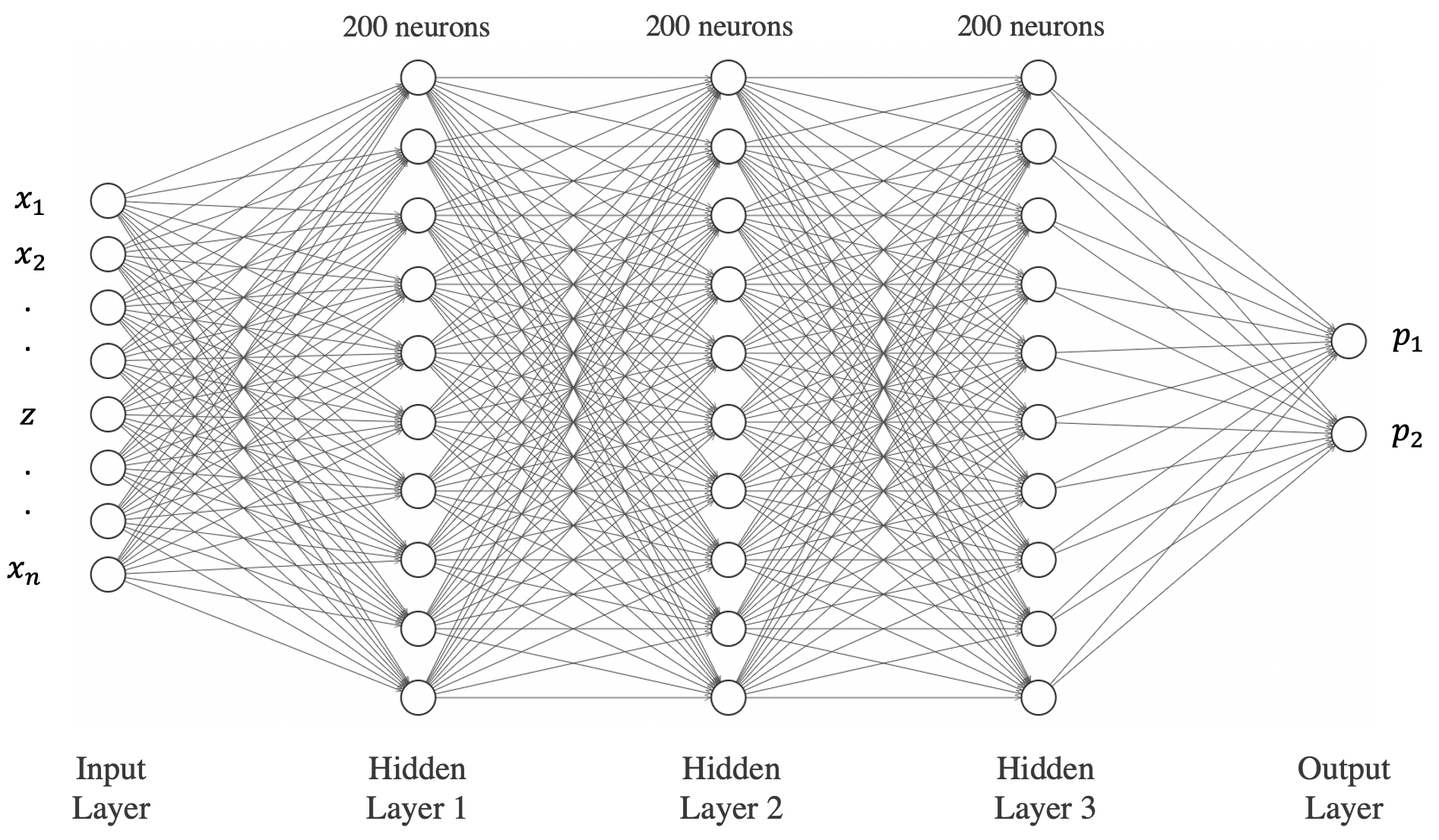}
    \caption{DNN architecture}
\label{DNN_arch}
\end{figure}

DNN usually outperforms traditional methods regarding prediction accuracy, because of the non-linear transformation. The outcome probability using DNN can be expressed as follows \cite{Hichem15,wang2021deep}:
\setlength{\abovedisplayskip}{10pt}
\setlength{\belowdisplayskip}{10pt}\begin{linenomath}
\begin{gather*}
    P(y_{i}=1|\boldsymbol{x_{i}}) = \frac{1}{1 + exp(-\Phi(\boldsymbol{x_{i}},\boldsymbol{\beta}))}
\end{gather*}
\end{linenomath}
 where $\Phi(\boldsymbol{x_{i}},\boldsymbol{\beta})$ represents a layer-by-layer transformation: $\Phi(\boldsymbol{x_{i}},\boldsymbol{\beta})=(g_{m}\circ ...g_{2}\circ g_{1})(\boldsymbol{x_{i}};\boldsymbol{\beta}) $, in which each $g_{l}(\boldsymbol{x_{i}^\top\beta})=ReLU(\boldsymbol{x_{i}^\top\beta}+b_{l})$ denotes one standard module in DNN which consists of linear and rectified linear unit (ReLU) transformations. A dropout layer with a dropout rate of 0.01 is applied after each standard module to prevent overfitting. The architecture of the DNNs used in this study is shown in Figure \ref{DNN_arch}, which includes 3 hidden layers and 200 neurons in each layer. Noted that the protected variable $z$ is also included in $\boldsymbol{x}$ as an explanatory variable.

\subsubsection{Bias Measurement}
\noindent After applying BLR and DNN models for a specific prediction task, we measure the fairness metrics for the prediction result. Based on the fairness definition of equality of opportunity, unfairness occurs when the machine-learned models offer much worse results for some demographic groups than others \cite{Hardt16}. The degree of unfairness is measured by the false positive rate (FPR) gap or the false negative rate (FNR) gap between two groups depending on the specific context. 
The two fairness metrics are calculated as:
\setlength{\abovedisplayskip}{10pt}
\setlength{\belowdisplayskip}{10pt}
\begin{linenomath}
\begin{gather}
    False\:Positive\:Rate\:(FPR)\:Gap=\frac{FP_{z=0}}{TN_{z=0}+FP_{z=0}}-\frac{FP_{z=1}}{TN_{z=1}+FP_{z=1}} \label{FPR}\\
    False\:Negative\:Rate\:(FNR)\:Gap=\frac{FN_{z=0}}{TP_{z=0}+FN_{z=0}}-\frac{FN_{z=1}}{TP_{z=1}+FN_{z=1}} \label{FNR}
\end{gather}
\end{linenomath}
\noindent In the above expressions, $TP_{z}$, $FP_{z}$, $TN_{z}$, and $FN_{z}$ represent the number of true positives, false positives, true negatives and false negatives in class $z$, with $z=0$ representing the disadvantaged group. For example, when examining racial bias in predicting frequent usage of rideshare, $FN_{z=0}$ represents the number of individuals in the minority group who frequently use rideshare but are wrongly categorized as not doing so, and $FP_{z=0}$ represents the number of individuals in the minority group who do not frequently use rideshare but are wrongly categorized as doing so. Higher false negative rate or false positive rate for the minority group intuitively suggests that the algorithm makes more mistakes on the ethnic minority group with regard to predicting whether an individual uses rideshare frequently, which might lead to significant mismatch between demand and supply of TNC service in minority neighborhoods. Therefore, our definition captures the essential intuition in transport equity discussions. For these two fairness metrics, lower absolute value is better. 

\subsection{Bias Mitigation}\label{sec_mitigation}
Adapted from the work of Beutel et al. \cite{Alex19}, we mitigate prediction disparity by adding a regularization term to the loss function. While Beutel et al. \cite{Alex19} used correlation between the output distributions of two groups as their regularization term since the outputs in their studies are continuous scores, we use the correlation between the predicted probability distributions of two groups instead, which makes the regularization term differentiable in our classification tasks. This regularization loss term helps shrink the difference of prediction disparity across groups towards zero. Compared with other approaches which generally come with notable engineering concerns, this approach is lightweight, can be easily adapted to real-world systems and has achieved good empirical results \cite{Alex19}. The loss function is specified as:

\begin{gather}
\min_{p} \quad (1-\lambda)L_{primary}+\lambda |Corr(p(\boldsymbol{x}),z|y=q)|
\shortintertext{where}
L_{primary}=\sum_{i=1}^{N}[-y_{i} log(p(\boldsymbol{x_{i}}))-(1-y_{i})log(1-p(\boldsymbol{x_{i}}))] \\
Corr(p(\boldsymbol{x}),z|y=q)=\frac{\sum_{i \in S_{q}} (p(\boldsymbol{x_i})-\overline{p(\boldsymbol{x_i}}))(z_i-\overline{z_i})}{(\sqrt{\sum_{i \in S_{q}}  (p(\boldsymbol{x_i})-\overline{p(\boldsymbol{x_i})})^2}+\epsilon)*
(\sqrt{\sum_{i \in S_{q}}  (z_i-\overline{z_i})^2}+\epsilon)}\\
S_{q}={\{i|y_i=q\}\: ; \:\epsilon=e^{-20}}
\qedhere
\end{gather}

\noindent In the above equation, $\boldsymbol{x_i}$ is the vector representing the explanatory variables. $y_i$ denotes the true outcome and $p(\boldsymbol{x_i})$ represents the estimated probability of the output $y_i$=1 using the explanatory variables. $q$ is equal to 1 if the false negative bias is examined and is equal to 0 if the false positive bias is examined.  $z_i$ denotes the value of the protected variable. $\epsilon$ is added to prevent the denominator from becoming zero. $S_q$ represents the set of samples with $y_{i}=q$, which is used to compute the correlation loss.\\

\noindent $L_{primary}$ is a negative log likelihood loss function for the DNN. The correlation term is a penalty added to the model based on the distribution of the predictions. By reducing the correlation between $p(\boldsymbol{x_i})$ and $z_i$ conditioning on $y_i=q$, it minimizes the conditional dependence between the distribution of the predicted probabilities and the group membership determined by the protected variable. $\lambda$ is a parameter controlling the tradeoff between primary loss and fairness loss. When $\lambda=0$, no bias mitigation is employed. In this research, we demonstrate the effectiveness of using this fairness-adjusted loss function in both BLR and DNN.\\

\section{Experiments}
In this section, the experiment results for the synthetic dataset, the NHTS data and the Chicago travel survey data are reported. For these three datasets, we first conduct BLR and DNN for model estimations, then examine fairness issues in the prediction results, and lastly apply our bias mitigation methods. DNN is implemented using the TensorFlow library in Python. BLR is implemented using the scikit-learn library when examining the fairness issues while using the TensorFlow library for bias mitigation, since TensorFlow allows us to modify the loss function.\\

\noindent Experiments conducted in TensorFlow all use the mini-batch stochastic gradient descent method with a batch size of 1,000 and a step size of 0.0001 in each training. We draw samples without replacement to generate the mini-batches within an epoch. After a mini-batch is generated, the algorithm calculates the prediction loss and updates the coefficients. The model that produces the lowest training loss among the 50 epochs is chosen and later performs prediction over the test data. We ran 5 trials of 5-fold validation for each experiment.

\subsection{Synthetic Experiment}

In our simulations, we consider the type of bias that arises when the true predictor and the protected variable are highly correlated. In this case, the true predictor of the outcome also happens to serve as a reliable proxy for class membership in the training set \cite{barocas2016big,zliobaite2017fairness}. For example, if the usage of rideshare is positively associated with income, and ethnic minorities in the training data tend to have lower income, the algorithm will tend to predict low usage of rideshare for the minority population, even if the true contributing factor for rideshare usage is income rather than race. This type of bias is inherent in the existing population inequality and has been less emphasized in previous literature.\footnote{Besides this inherent bias, other sources of bias could exist, such as the imbalanced training data problem. This problem arises when the disadvantaged group has insufficient training data or is misrepresented, in which case the model will either fail to learn a correct statistical pattern or favor the majority group during the estimates, since the training data in the disadvantaged groups often misrepresent the true population when they are insufficient \cite{yao2017beyond,gu2020self,Rajkomar18,ferryman2018fairness,calders2013unbiased}.}
\\


\subsubsection{Data generation process}
\noindent In the synthetic dataset, each data point can be represented by a tuple ($z_{i}$,$x_{i}$,$\boldsymbol{k_{i}}$,$y_{i}$), where $z$ represents the protected variable (e.g. race, gender), $x$ is the explanatory variable that is correlated with $z$ (e.g. income), $\boldsymbol{k}$ is a vector of explanatory variables which does not include $x$ and $z$. $y$ represents the binary outcome.\\

\noindent First, $x$ and all the elements in $\boldsymbol{k}$ are drawn from the independent standard normal distributions. $z$ is generated as a binary variable that is positively correlated with $x$ and is independent with $\boldsymbol{k}$. The label $y$ is drawn from a binomial distribution with probability $Pr(y=1)=\frac{1}{1+\exp{(-V)}}$. The systematic utility function $V$ takes $x$ and $\boldsymbol{k}$ as the input variables.\\

\noindent We test two scenarios with the true utility function taking a linear form and a quadratic form respectively. Let $V=\alpha+\boldsymbol{w} \phi(x,\boldsymbol{k})$. In the first scenario, $\phi(x,\boldsymbol{k})$ takes the linear transformation: $\phi(x,\boldsymbol{k})=[x,k_{1},k_{2},...,k_{d}]$. The weight for $x$ is set to 1, and the weights for other explanatory variables are chosen from the set \{-0.5,0.5\} with equal probabilities. In the second scenario, $\phi(x,\boldsymbol{k})$ takes the quadratic transformation: $\phi(x,\boldsymbol{k})=[x,k_{1},...,k_{d},x^2,k_{1}^2,...,k_{d}^2]$. The weights for $x$ and $x^2$ are set as 1 and 0.5, and the weights for other explanatory variables take \{-0.5,0.5\} with equal probabilities. Our data generation process makes sure that the mean value of $z$ is 0.5, which means that there are approximately equal numbers of $z=0$ and $z=1$, thus giving us a balanced dataset. Detailed descriptions of the data generation process can be found in Appendix~\ref{sec:DGP}.\\

\noindent We examine the cases where $Cov(z, x)$ is non-negative and $x$ positively affects $V$. We therefore use $z=0$ to mimic the disadvantaged population and use $z=1$ to mimic the privileged population, since in the real world the privileged population (e.g. the ethnic majority group) is often positively correlated with the factor (e.g. income) that has a positive effect on the utility of an advanced mobility service (e.g. the utility of using a ride-hailing service).\\

\subsubsection{Fairness measurement results}
\noindent In the prediction phase, we use $z$, $x$ and $\boldsymbol{k}$ as the explanatory variables for both the logit model and the DNN model, so these two choice models are defined as $Logit(z,x, \boldsymbol{k})$ and $DNN(z,x, \boldsymbol{k})$.\\

\noindent First, we want to examine how fairness measures and accuracy vary regarding the correlation between the sensitive attribute $z$ and the explanatory variable $x$,
the sample size and the number of explanatory variables in the data generation process. Therefore, we run experiments along these three dimensions, and when each dimension is examined, we set the other two dimensions as the default values. The default values for $Cov(z, x)$, sample size and number of explanatory variables are 0.2, $10^6$ and 5. For each experiment, three datasets are randomly generated based on the above data generation process.\footnote{Occasionally, the data generation process in Scenario 2 produced datasets with highly imbalanced outcomes (e.g. when the minority class accounts for less than 30\% of the total samples). In that case, we would drop that imbalanced dataset and generate another one. We iterated this process until all the datasets are roughly balanced (when the minority class accounts for 40\%-50\% of the total samples)}\\

\begin{figure}
\begin{subfigure}[b]{.32\linewidth}
\centering
\includegraphics[width=\linewidth]{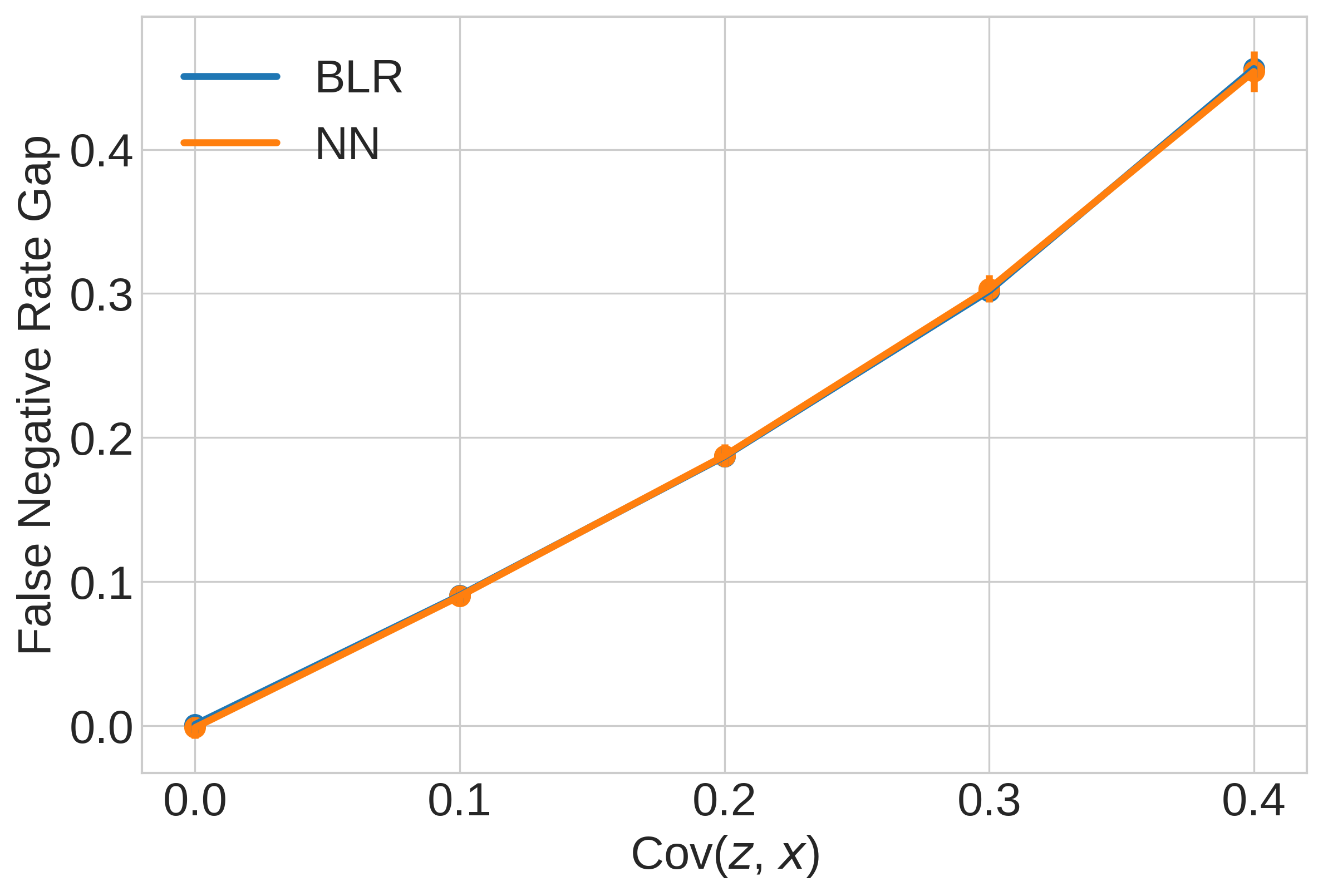}
\caption{FNR gap vs. Cov$(z, x)$}\label{1a}
\end{subfigure}\hfill
\begin{subfigure}[b]{.32\linewidth}
\centering
\includegraphics[width=\linewidth]{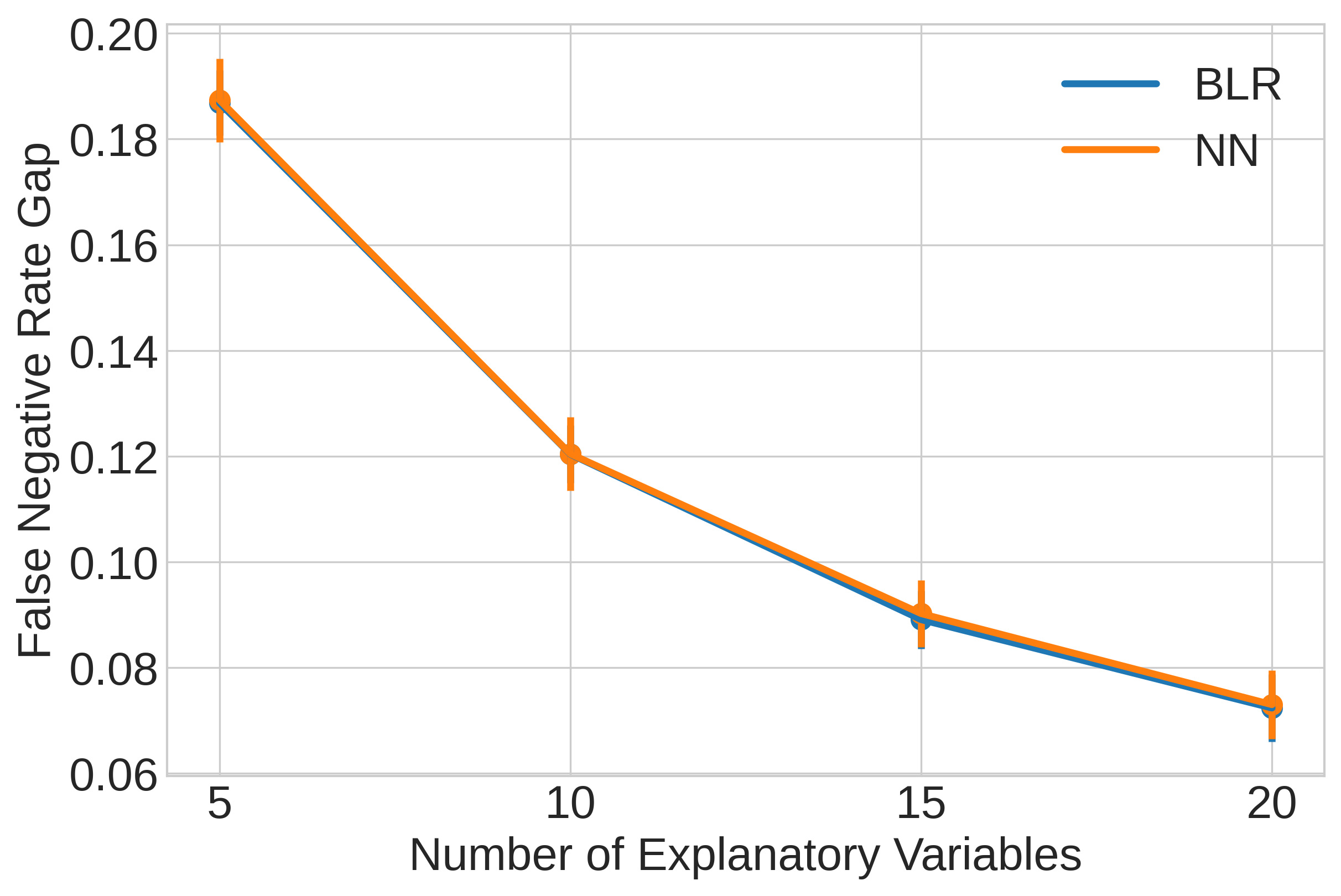}
\caption{FNR gap vs. \# predictors}\label{1b}
\end{subfigure}\hfill
\begin{subfigure}[b]{.32\linewidth}
\centering
\includegraphics[width=\linewidth]{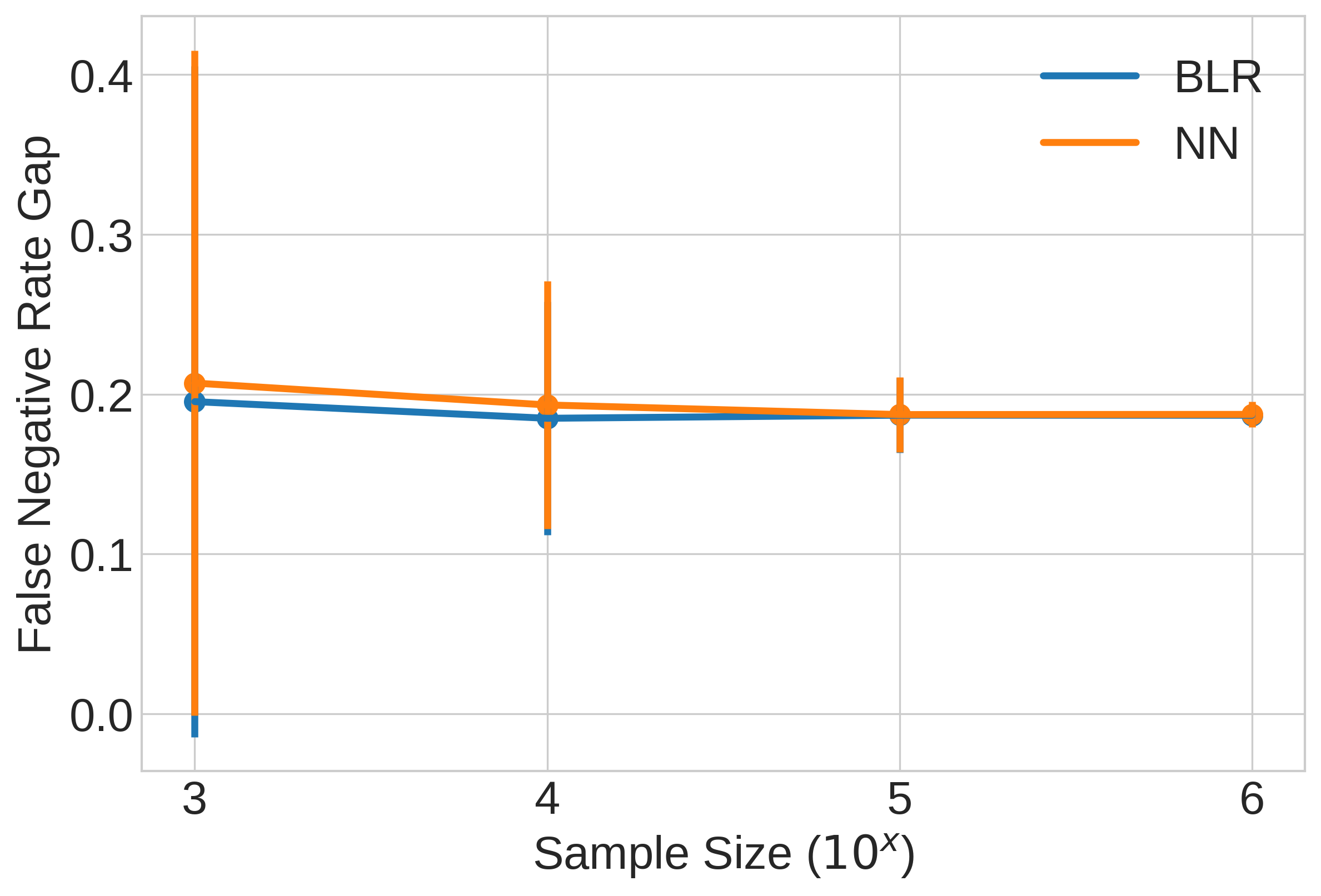}
\caption{FNR gap vs. Sample Size}\label{1c}
\end{subfigure}\\
\begin{subfigure}[b]{.32\linewidth}
\centering
\includegraphics[width=\linewidth]{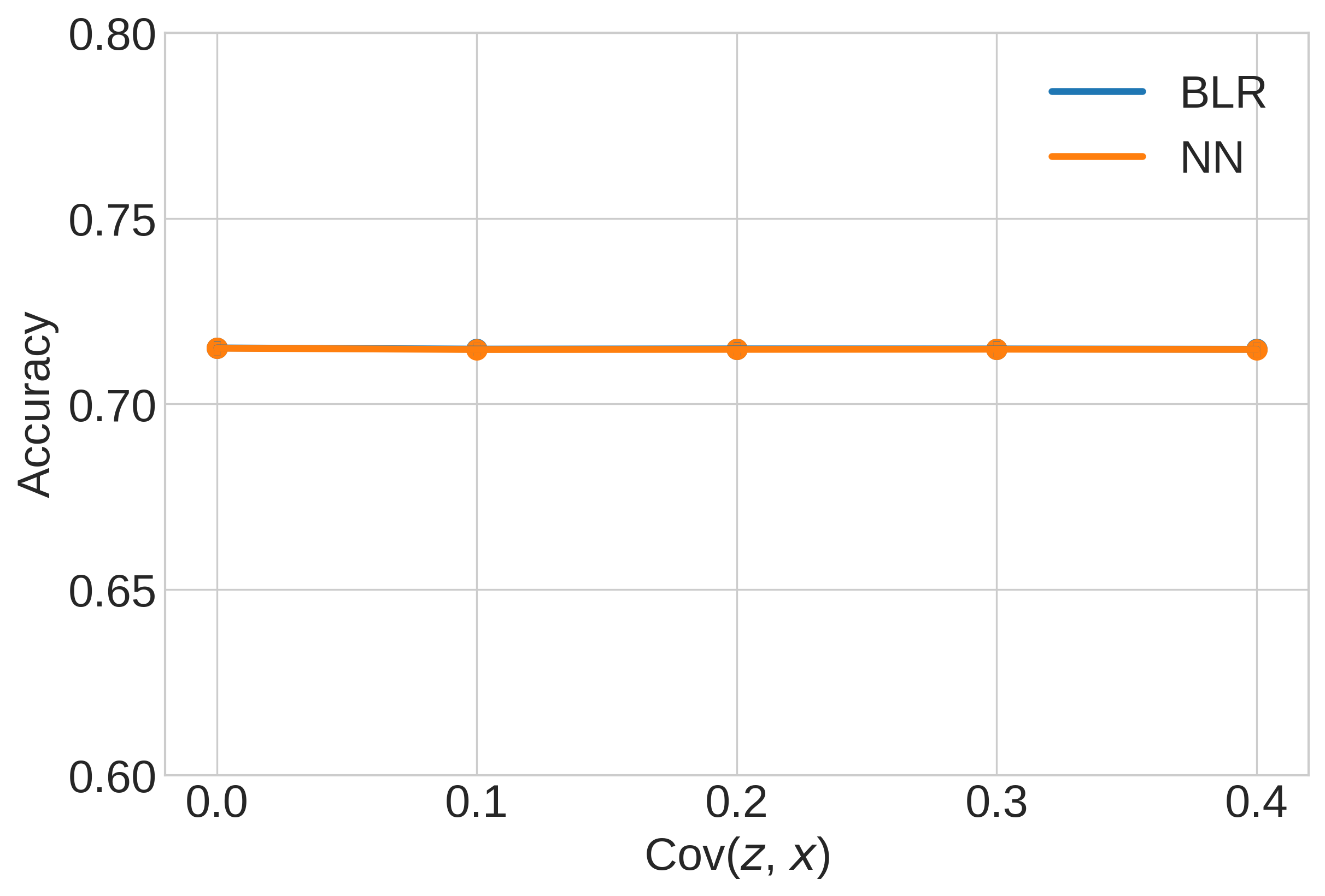}
\caption{Accuracy vs. Cov$(z, x)$}\label{1d}
\end{subfigure}\hfill
\begin{subfigure}[b]{.32\linewidth}
\centering
\includegraphics[width=\linewidth]{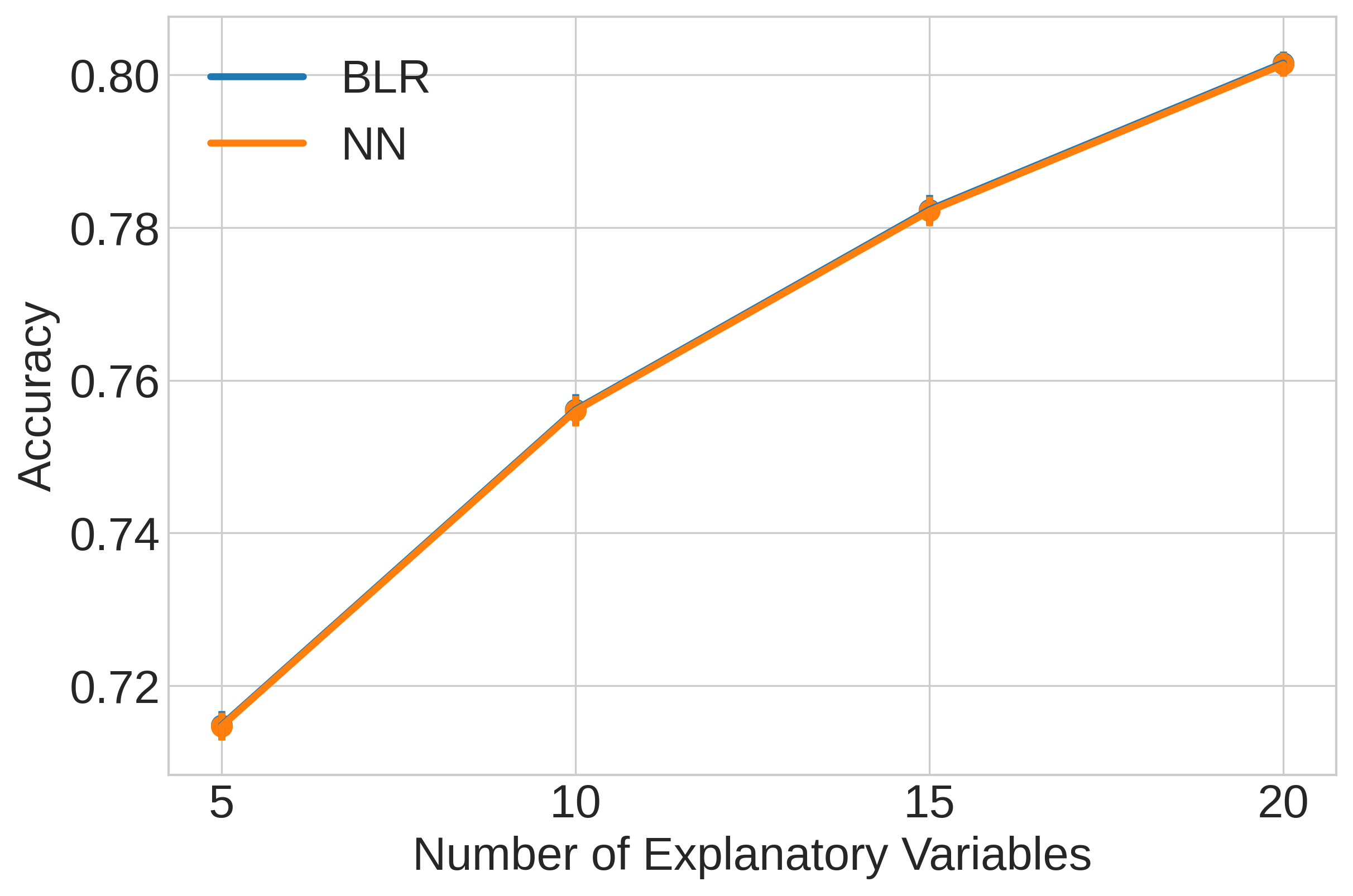}
\caption{Accuracy vs. \# predictors}\label{1e}
\end{subfigure}\hfill
\begin{subfigure}[b]{.32\linewidth}
\centering
\includegraphics[width=\linewidth]{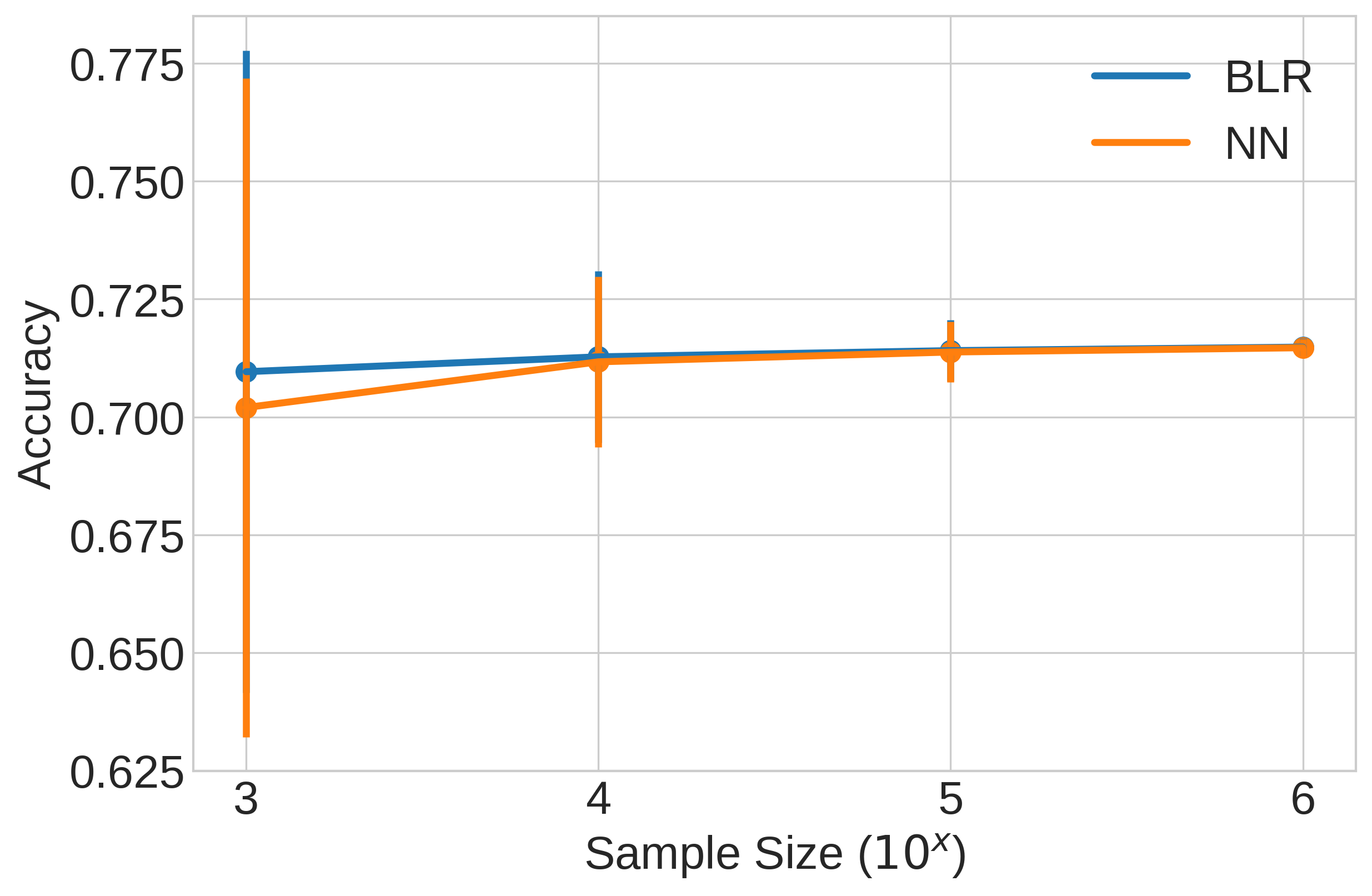}
\caption{Accuracy vs. Sample Size}\label{1f}
\end{subfigure}

\caption{Fairness metric and accuracy with different parameters (BLR vs. DNN): true model taking the linear form; estimation models: $Logit(z,x, \boldsymbol{k})$ and $DNN(z,x, \boldsymbol{k})$. $Logit(z,x, \boldsymbol{k})$ adopts a linear specification, so both models contain the true model.}
\label{figure_syn1}
\end{figure}

\begin{figure}
\begin{subfigure}[b]{.32\linewidth}
\centering
\includegraphics[width=\linewidth]{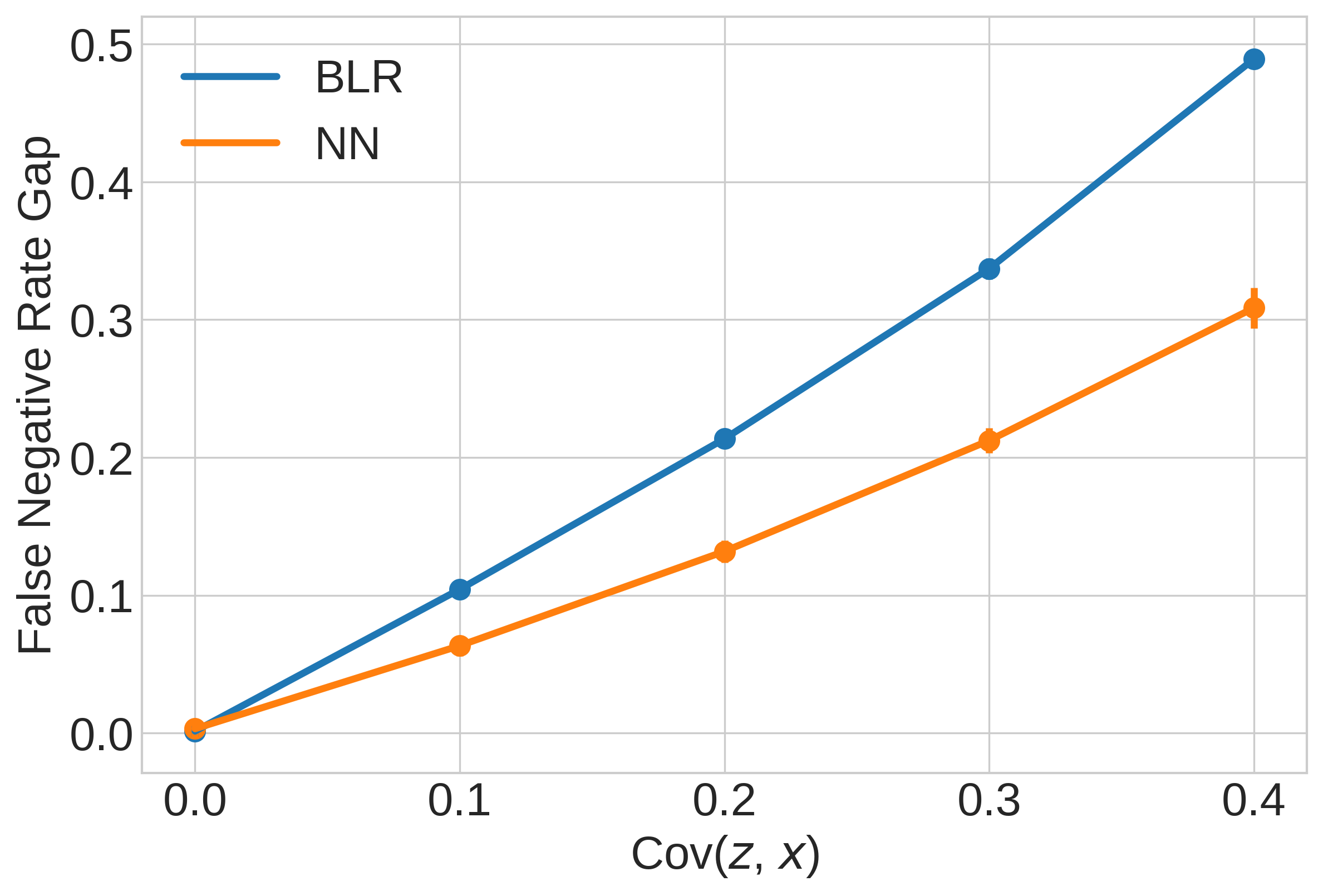}
\caption{FNR gap vs. Cov$(z, x)$}\label{2a}
\end{subfigure}\hfill
\begin{subfigure}[b]{.32\linewidth}
\centering
\includegraphics[width=\linewidth]{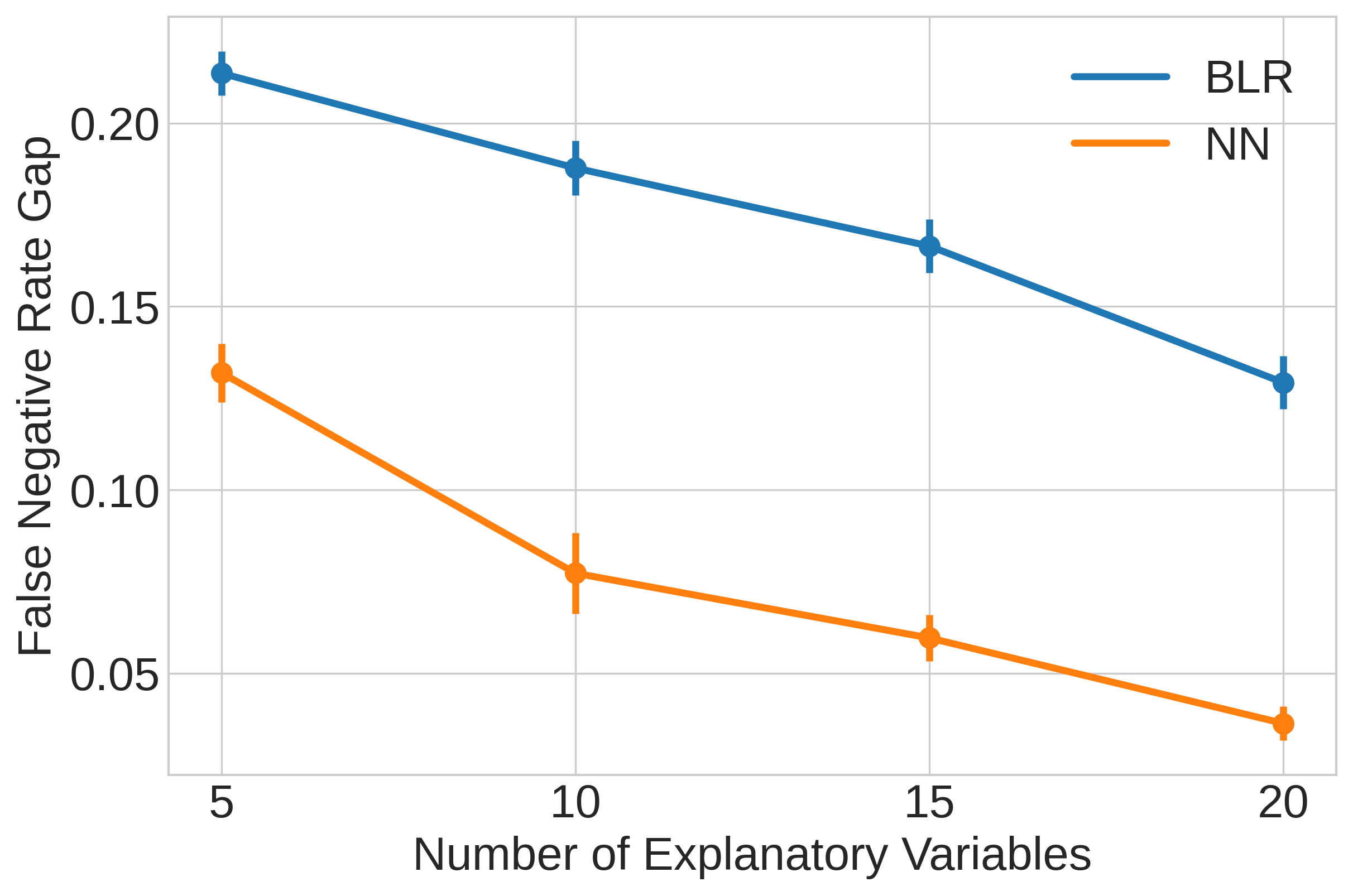}
\caption{FNR gap vs. \# predictors}\label{2b}
\end{subfigure}\hfill
\begin{subfigure}[b]{.32\linewidth}
\centering
\includegraphics[width=\linewidth]{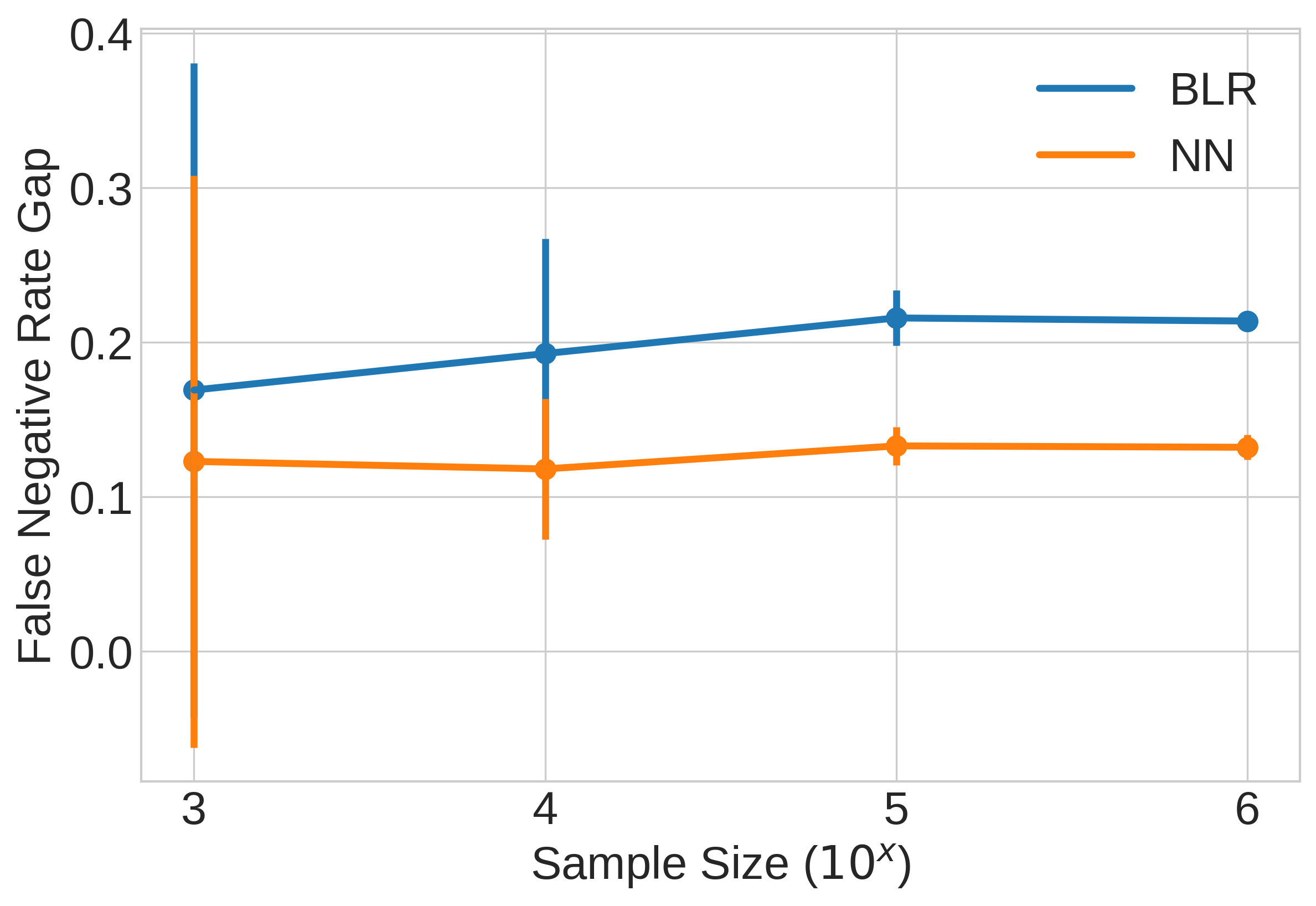}
\caption{FNR gap vs. Sample Size}\label{2c}
\end{subfigure}\\
\begin{subfigure}[b]{.32\linewidth}
\centering
\includegraphics[width=\linewidth]{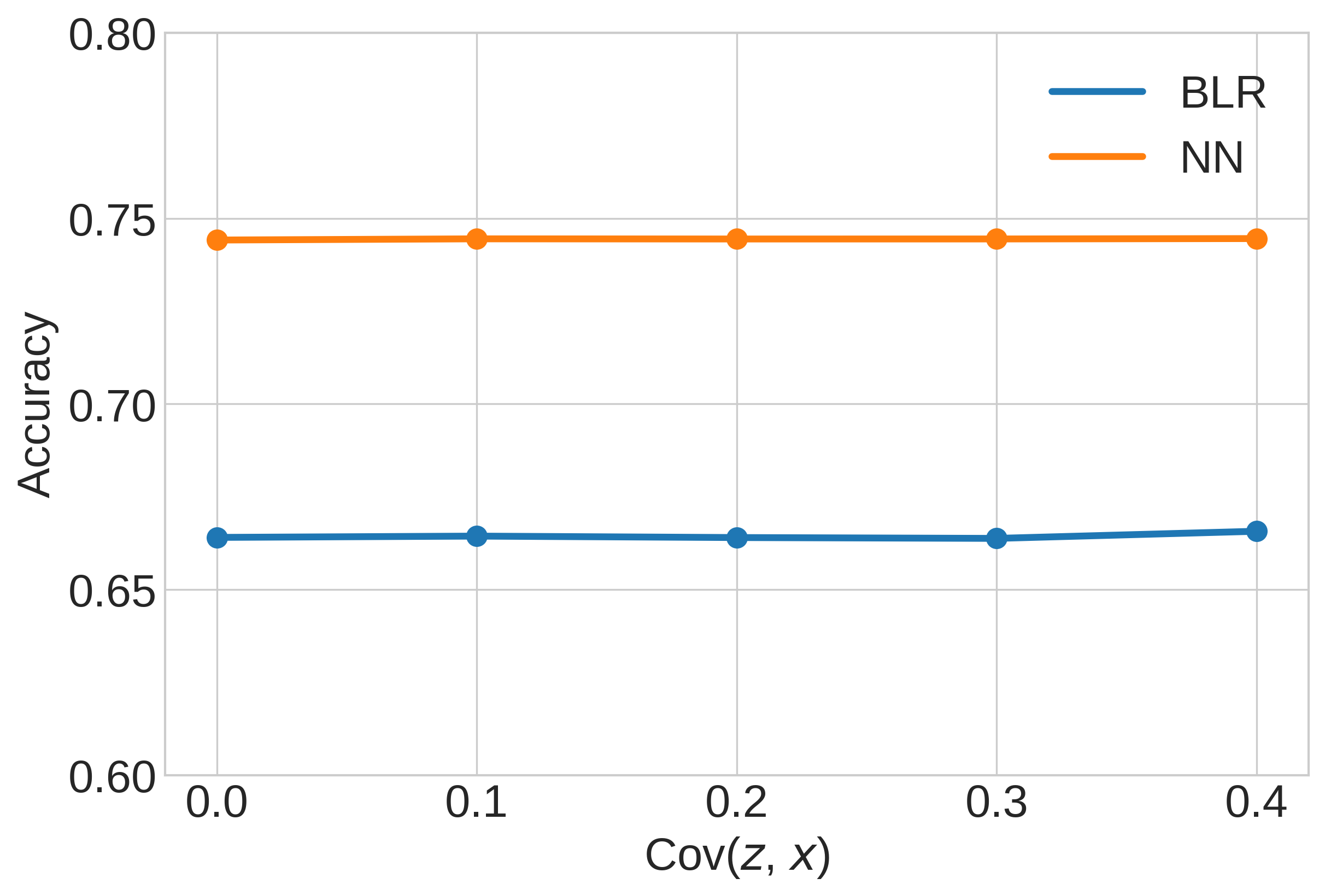}
\caption{Accuracy vs. Cov$(z, x)$}\label{2d}
\end{subfigure}\hfill
\begin{subfigure}[b]{.32\linewidth}
\centering
\includegraphics[width=\linewidth]{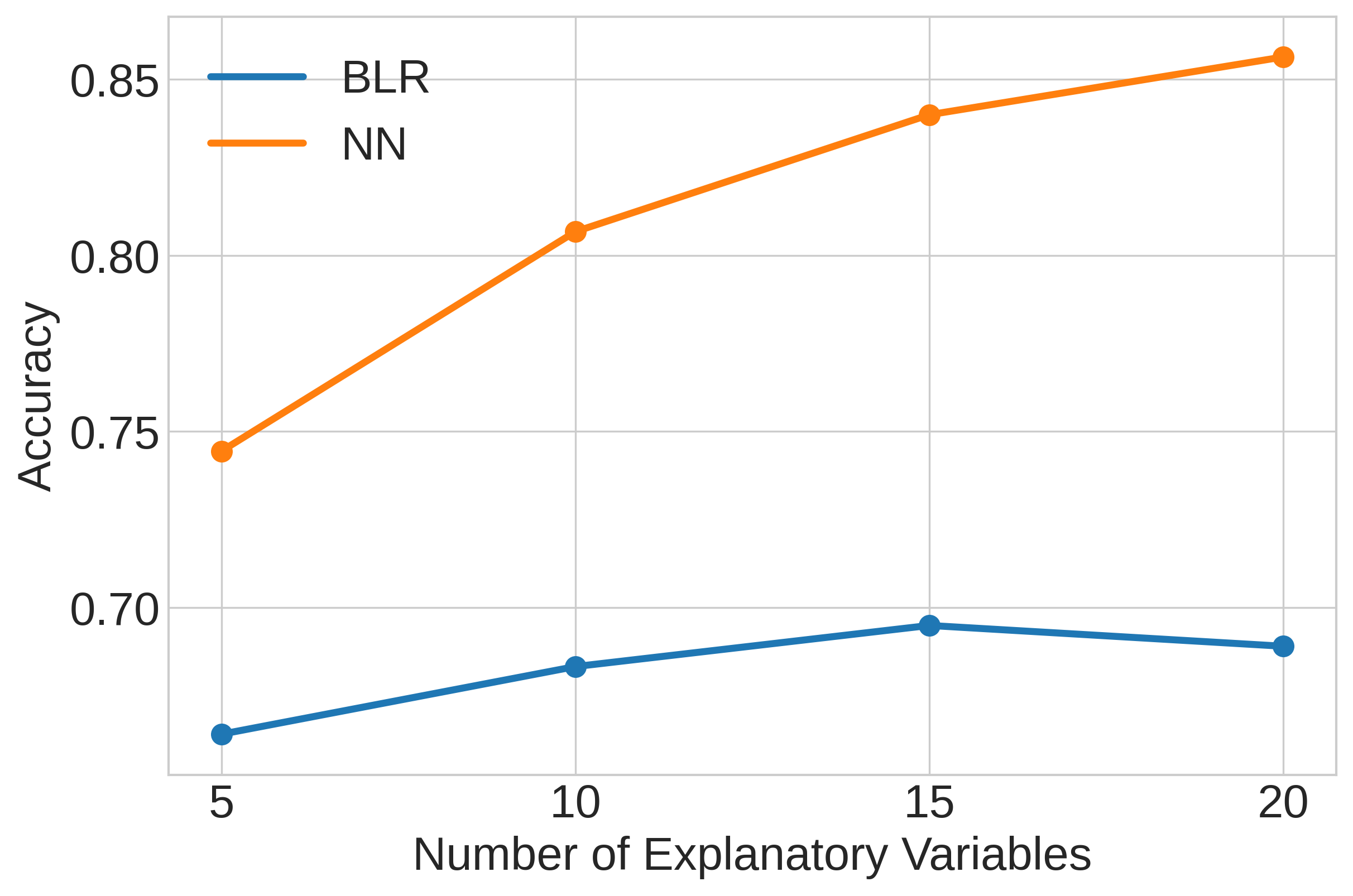}
\caption{Accuracy vs. \# predictors}\label{2e}
\end{subfigure}\hfill
\begin{subfigure}[b]{.32\linewidth}
\centering
\includegraphics[width=\linewidth]{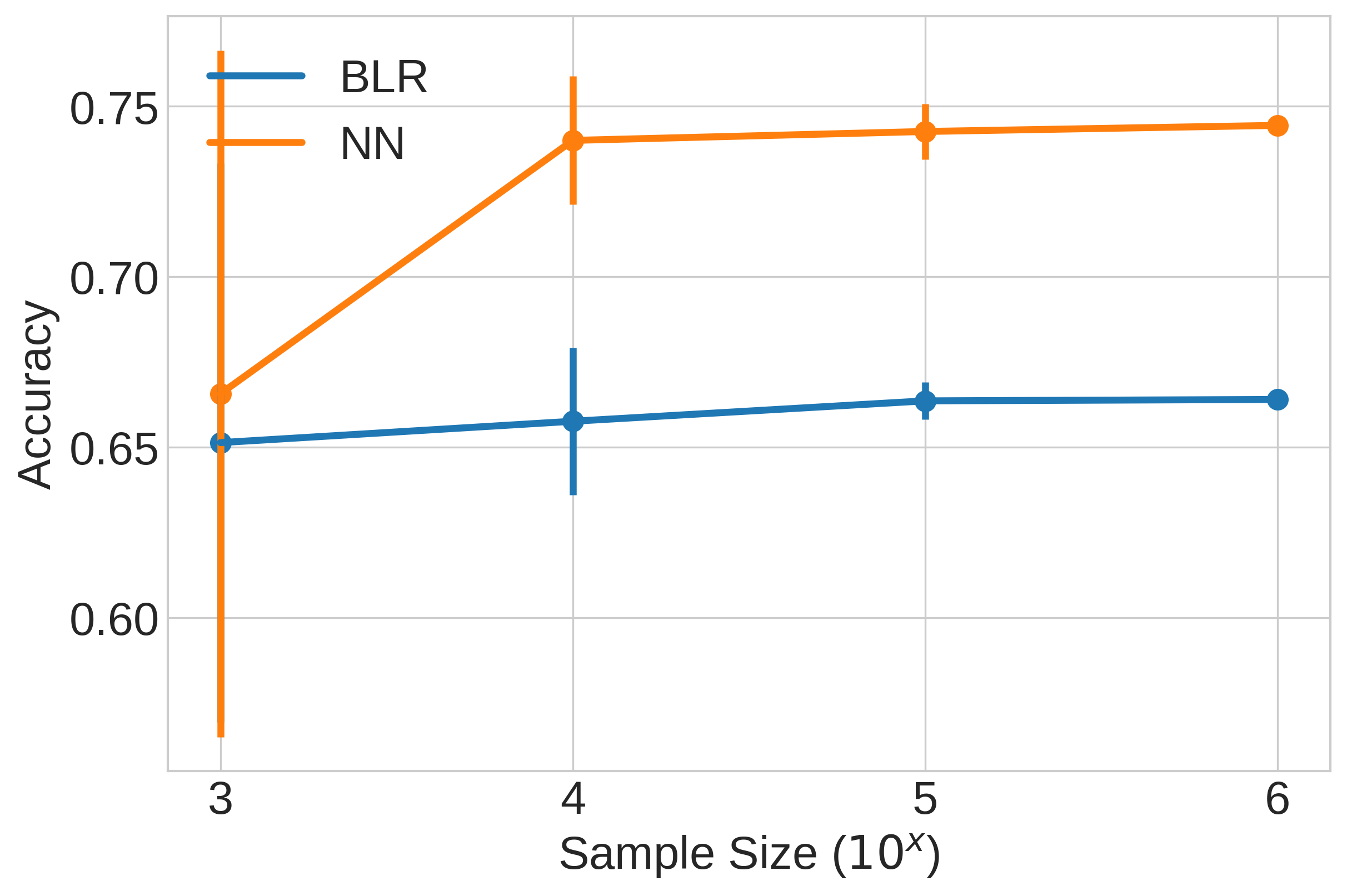}
\caption{Accuracy vs. Sample Size}\label{2f}
\end{subfigure}

\caption{Fairness metric and accuracy with different parameters (BLR vs. DNN): true model taking the quadratic form; estimation models: $Logit(z,x, \boldsymbol{k})$ and $DNN(z,x, \boldsymbol{k})$. $Logit(z,x, \boldsymbol{k})$ follows a linear model specification, so it has the misspecification error.}
\label{figure_syn2}
\end{figure}

\noindent Figure \ref{figure_syn1} presents the results of the linear data generating process, while in Figure \ref{figure_syn2}, the true data generating process is quadratic in variables. In both figures, the first row shows the FNR gap (see Equation \ref{FNR}) between the disadvantaged group (defined as $z=0$) and the privileged group (defined as $z=1$) as the measure of fairness and the second row shows the prediction accuracy. The x-axis of the first, second and third columns respectively represent the covariance between $z$ and $x$, the number of explanatory variables in the data generation process and the sample size. Each figure plots both the BLR and DNN results, which are represented by the blue and orange colors. The figures plot the values averaged across 5 trials of 5-fold validation in 3 datasets for each experiment; the error bar indicates the standard deviation multiplied by 1.96, which approximates the confidence interval of the estimations.\\

\noindent As shown in Figure \ref{figure_syn1}, the results of BLR and DNN mostly overlap since they both recover the true linear model. Figure \ref{1a} shows that the FNR gap increases with $Cov(z, x)$, indicating that as $x$ becomes more positively correlated with $z$, the algorithm is more likely to falsely associate the disadvantaged population ($z$=0) with negative outcomes, even if their real outcomes are actually positive. Prediction disparity is a metric relatively independent of the predictive performance, as there is no difference in prediction accuracy with different $Cov(z,x)$ (Figure \ref{1d}); it is also not due to the imbalanced training data problem, as the outcome variable, the protected variable and all the explanatory variables are balanced in the training data. The prediction disparity is purely inherent in the relationship among variables in the data. Figure \ref{1b} shows that the FNR gap decreases with the number of explanatory variables, which is probably because increasing the number of predictors dilutes the influence of $x$ on the outcome. Figure \ref{1c} shows that the variance of the fairness and accuracy estimations decrease as the sample size increases.\\

\noindent Figure \ref{figure_syn2} shows the results of prediction fairness and accuracy when the true data generation model takes a quadratic form. In this case, the BLR with linear specification has the model misspecification error while DNN does not. Figure \ref{2a} shows that prediction disparity still increases with the increase of $Cov(z, x)$, and DNN is always associated with smaller FNR compared with BLR for $Cov(z, x)$>0. This result indicates that the model misspecification not only induces more prediction error, but also harms prediction fairness. Figure \ref{2b} shows that though increasing the number of explanatory variables can reduce the prediction disparity, the magnitude of the prediction disparity caused by model misspecification was not reduced. Figure \ref{2c} indicates that the fairness prediction result becomes more stable as the sample size increases.

\subsubsection{Bias mitigation results}
\begin{figure}
\begin{subfigure}[b]{.32\linewidth}
\centering
\end{subfigure}\hfill
\begin{subfigure}[b]{.4\linewidth}
\centering
\includegraphics[width=\linewidth]{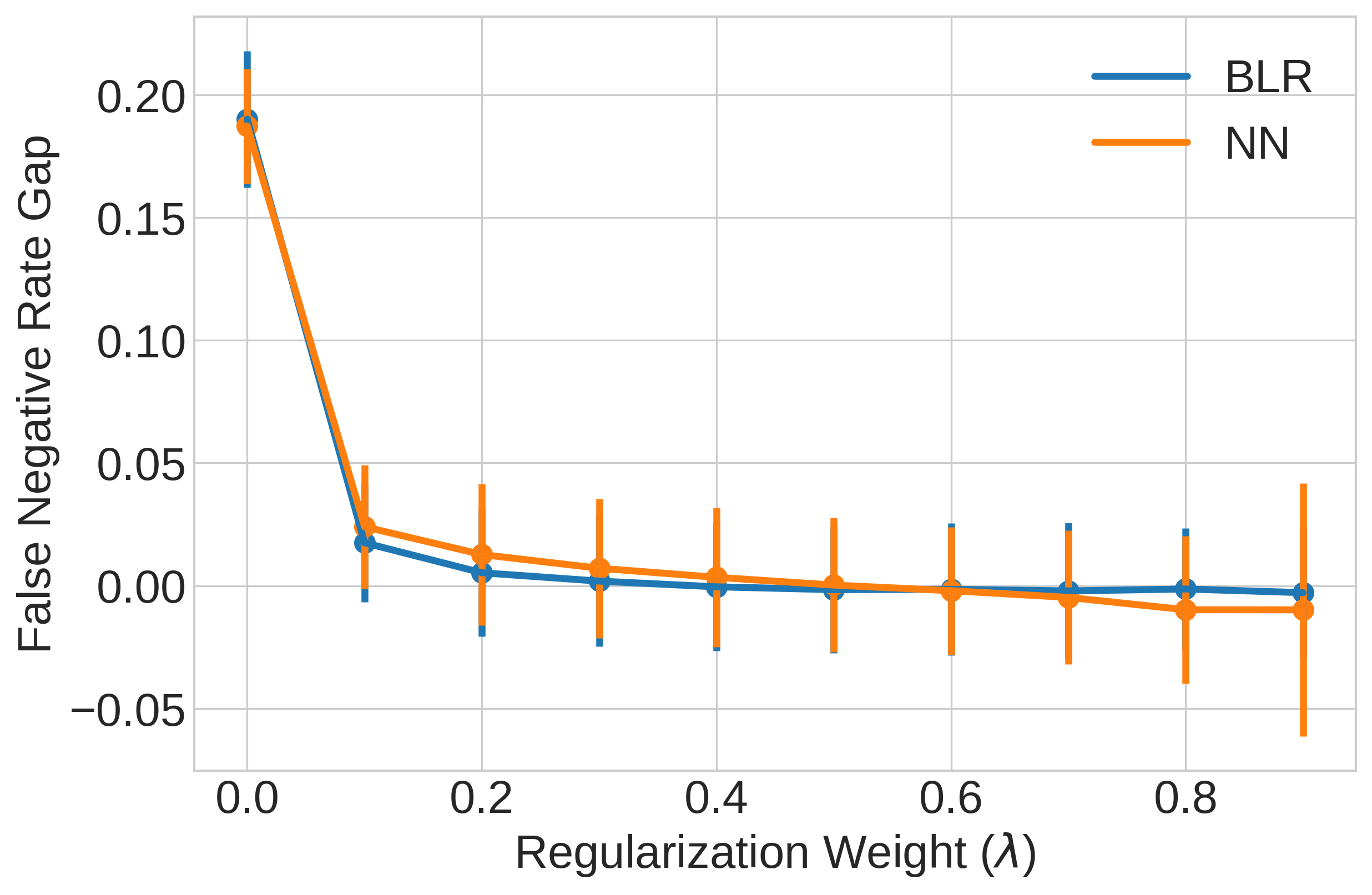}
\caption{FNR gap vs. Regularization Weight}\label{sync_mitigate_logit_fairness}
\end{subfigure}\hfill
\begin{subfigure}[b]{.4\linewidth}
\centering
\includegraphics[width=\linewidth]{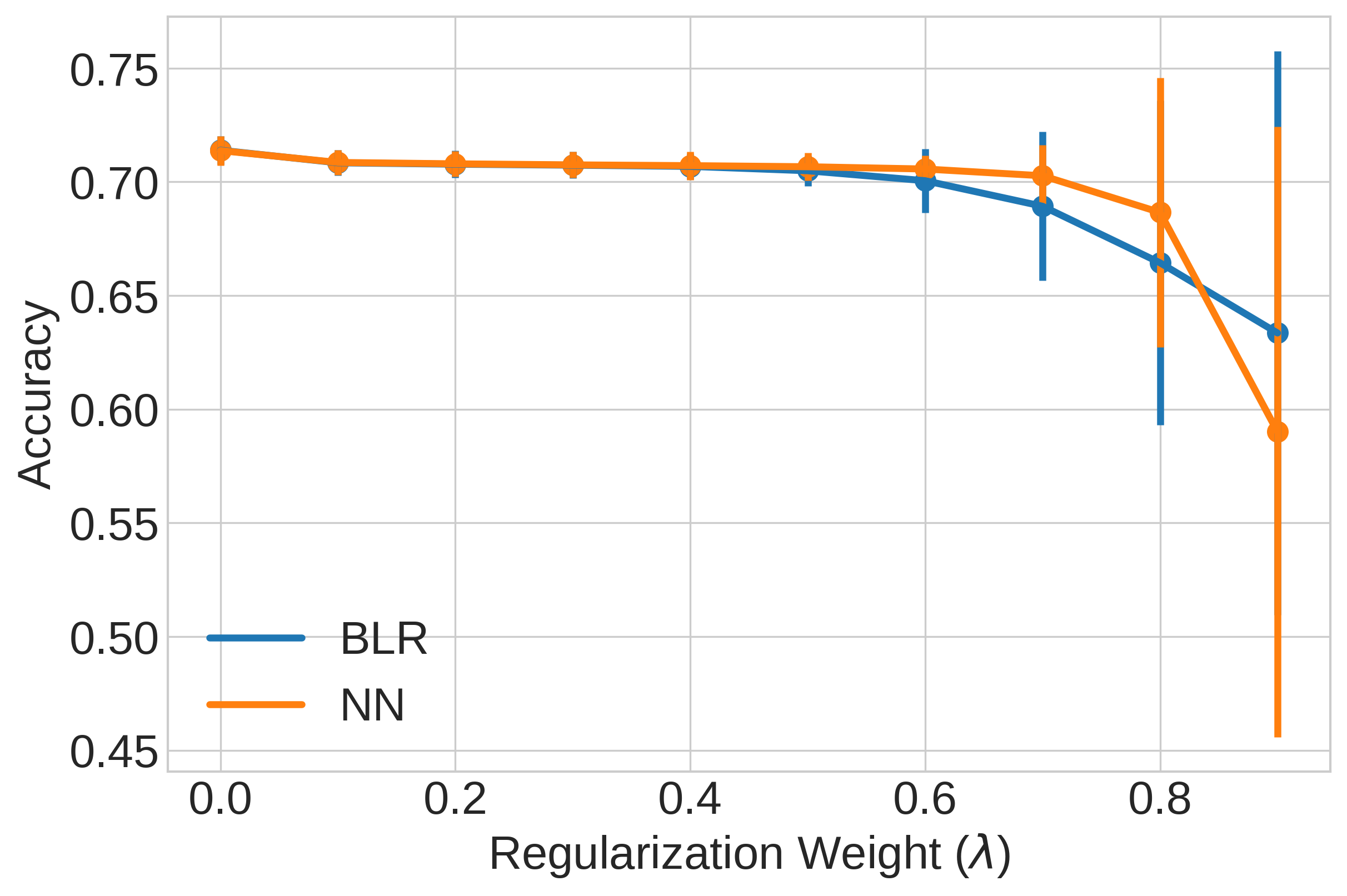}
\caption{Accuracy vs. Regularization Weight}\label{sync_mitigate_logit_accuracy}
\end{subfigure}\hfill
\begin{subfigure}[b]{.32\linewidth}
\centering
\end{subfigure}\hfill

\caption{Fairness and accuracy by bias mitigation weight ($\lambda$): true model taking the linear form (Scenario 1)}
\label{sync_mitigate_logit}
\end{figure}

\begin{figure}
\begin{subfigure}[b]{.32\linewidth}
\centering
\end{subfigure}\hfill
\begin{subfigure}[b]{.4\linewidth}
\centering
\includegraphics[width=\linewidth]{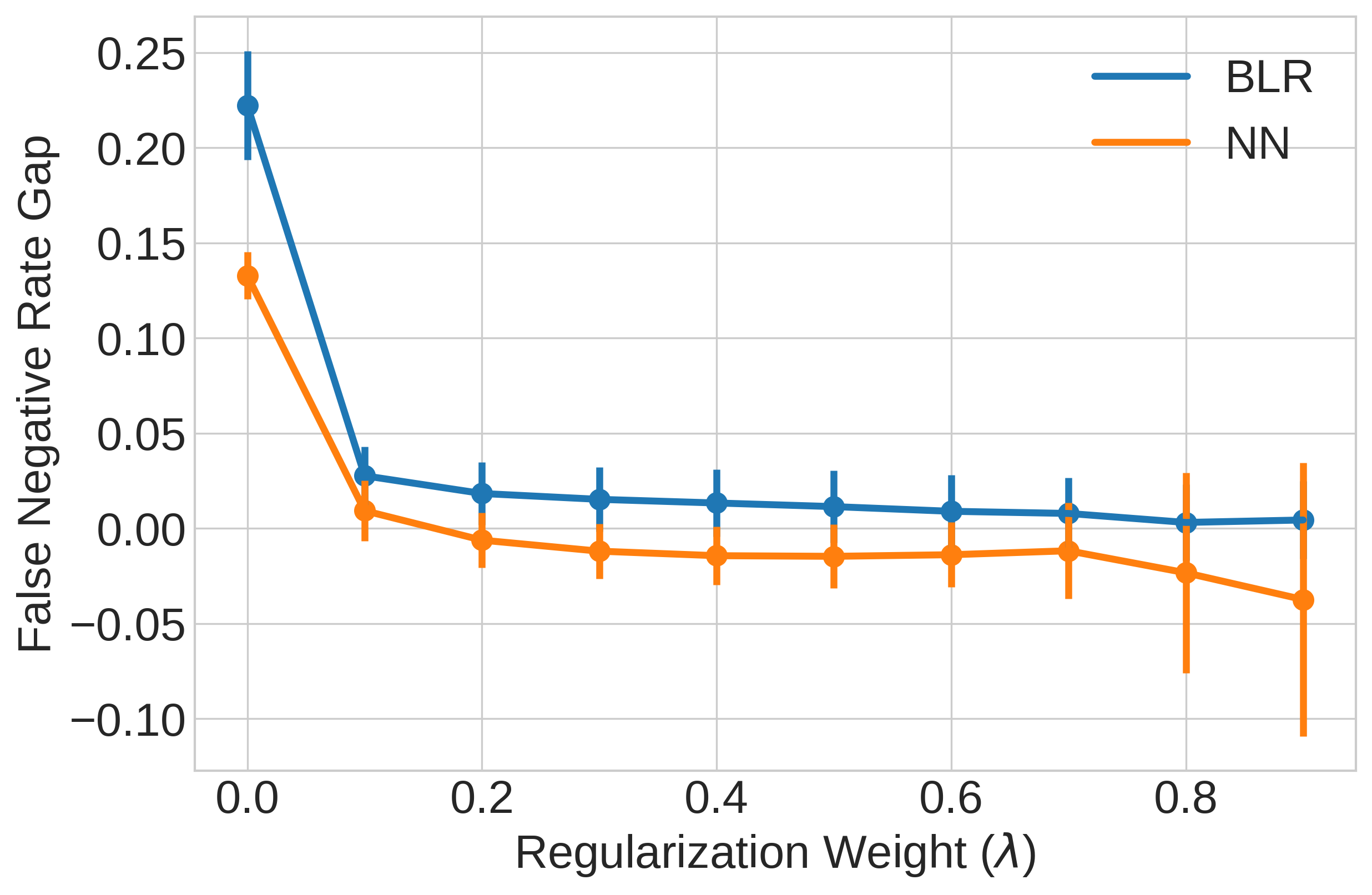}
\caption{FNR gap vs. Regularization Weight}\label{sync_mitigate_quadratic_fairness}
\end{subfigure}\hfill
\begin{subfigure}[b]{.4\linewidth}
\centering
\includegraphics[width=\linewidth]{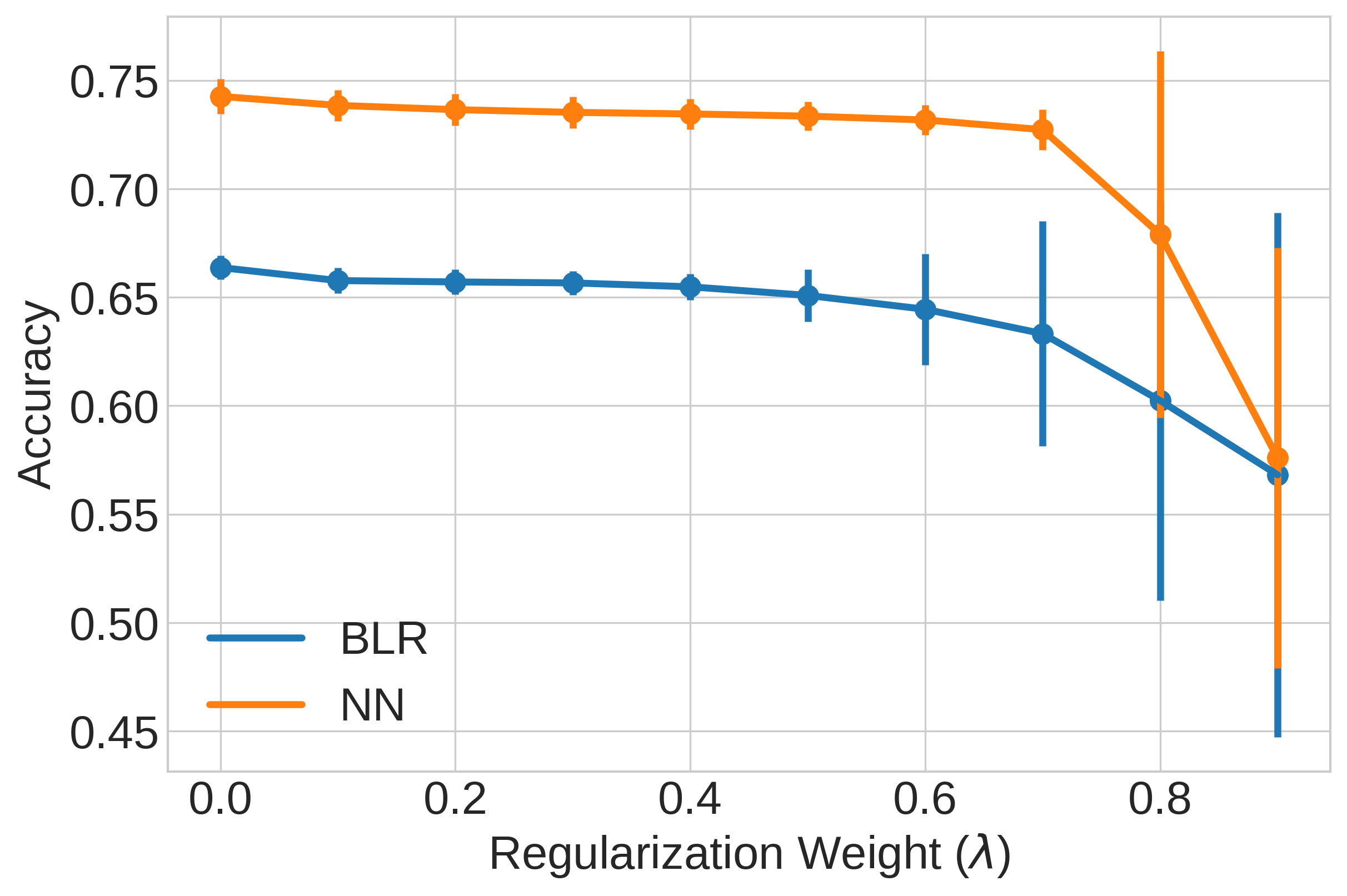}
\caption{Accuracy vs. Regularization Weight}\label{sync_mitigate_quadratic_accuracy}
\end{subfigure}\hfill
\begin{subfigure}[b]{.32\linewidth}
\centering
\end{subfigure}\hfill

\caption{Fairness and accuracy by bias mitigation weight ($\lambda$): true model taking the quadratic form (Scenario 2)}
\label{sync_mitigate_quadratic}
\end{figure}

To address prediction disparity, we apply the bias mitigation method illustrated in Section \ref{sec_mitigation} to the synthetic datasets with $Cov(z, x)=0.2$, sample size of 100,000 and number of predictors set to 5. For each regularization weight $\lambda$, we run the training procedure 5 times for each of the 3 datasets with 5-fold cross-validation and report the average results in Figure \ref{sync_mitigate_logit} for Scenario 1 and Figure \ref{sync_mitigate_quadratic} for Scenario 2. The error bars in the figures indicates the standard deviation multiplied by 1.96, which approximates the confidence interval. \\

\noindent Figure \ref{sync_mitigate_logit_fairness} and \ref{sync_mitigate_quadratic_fairness} show that in both scenarios, applying the regularization even with a small weight (e.g. $\lambda=0.1$) can substantially reduce the prediction disparity and this finding holds for both BLR and DNN. Given the model misspecification for BLR in Scenario 2, Figure \ref{sync_mitigate_quadratic_fairness} shows that our method is still effective in reducing the prediction bias to as low as zero. \\

\noindent Figure \ref{sync_mitigate_logit_accuracy} and \ref{sync_mitigate_quadratic_accuracy} report the corresponding model accuracy as the regularization weight varies. The results show that when $\lambda<0.7$, the accuracy decreases only slightly. These results suggest that improvement in prediction fairness can be achieved with a minimal cost of accuracy.\\

\begin{figure}
\begin{subfigure}[b]{.32\linewidth}
\centering
\end{subfigure}\hfill
\begin{subfigure}[b]{.4\linewidth}
\centering
\includegraphics[width=\linewidth]{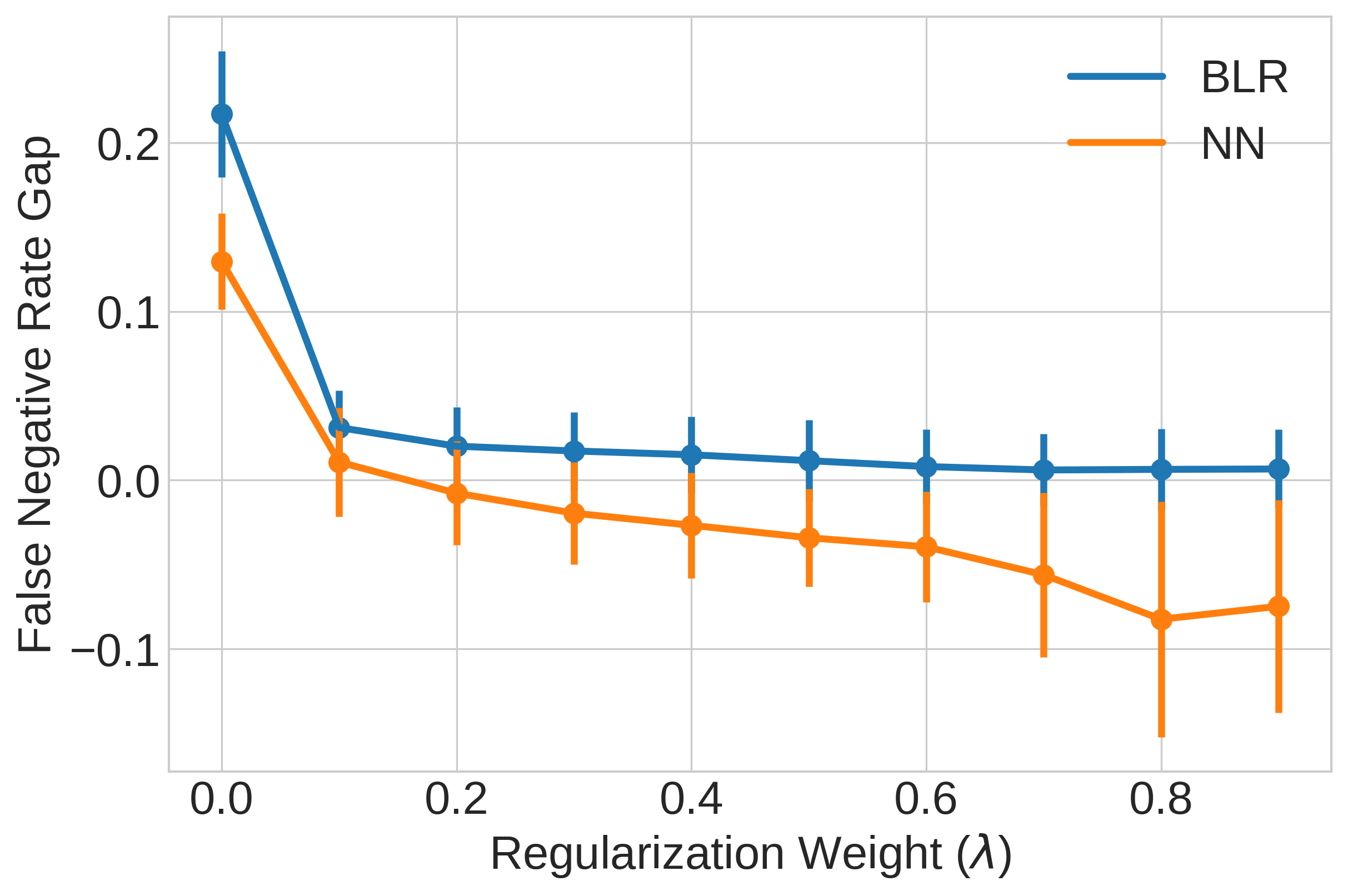}
\caption{FNR gap vs. Regularization Weight}\label{sync_mitigate_imbalance_fairness}
\end{subfigure}\hfill
\begin{subfigure}[b]{.4\linewidth}
\centering
\includegraphics[width=\linewidth]{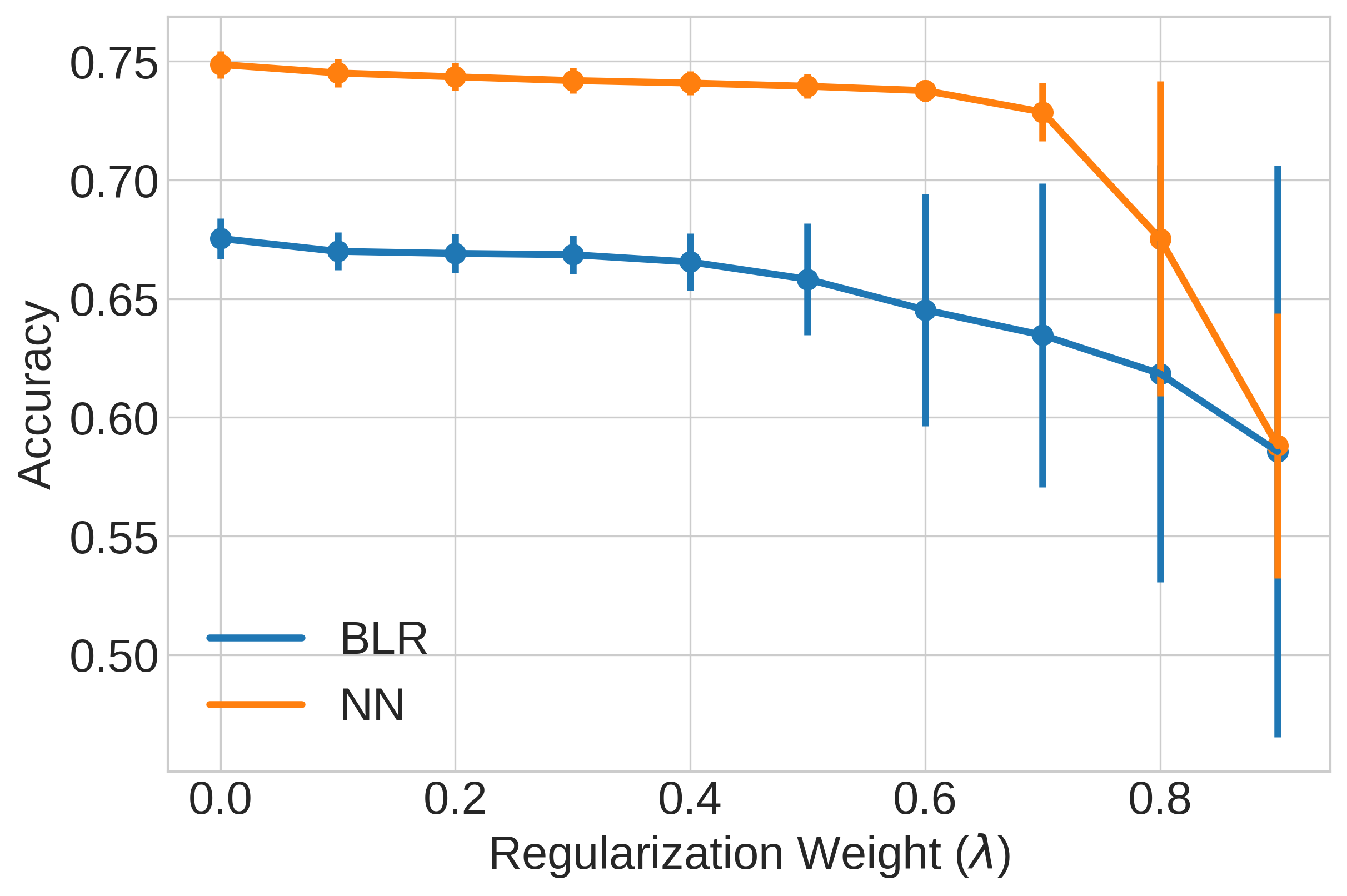}
\caption{Accuracy vs. Regularization Weight}\label{sync_mitigate_imbalance_accuracy}
\end{subfigure}\hfill
\begin{subfigure}[b]{.32\linewidth}
\centering
\end{subfigure}\hfill

\caption{Fairness and accuracy by bias mitigation weight ($\lambda$): imbalanced dataset}
\label{sync_mitigate_imbalance}
\end{figure}

\noindent We also test our mitigation method on the dataset with an imbalanced distribution over the protected variable. For the data generated in Scenario 2, we randomly drop half of the data in the disadvantaged group to make the ratio of number of samples in the disadvantaged and the privileged groups roughly 1:2. Figure \ref{sync_mitigate_imbalance} shows the bias mitigation results with this new dataset. It shows that the increase in bias mitigation weight leads to a reduction in the mean FNR gap for both DNN and BLR, and the fairness-accuracy trade-off still exists. However, for DNN, though the FNR gap continues to decrease with increasing $\lambda$, the absolute value of the FNR gap becomes larger when $\lambda>0.2$, indicating that the model may adjust too much to mitigate the bias when the bias mitigation weight is too large.


\subsection{The NHTS Dataset}
\label{nhts_exp}
\subsubsection{Data and variables}
The NHTS data are collected directly from a stratified random sample of U.S. households. The richness of the dataset enables us to examine fairness in predictions with varying dependent variables and protected attributes. Protected attributes are the variables we want to protect against in the model prediction process, which include race, gender, income, medical condition and urban-rural divide in this study. In terms of ``race'', we define the ethnic minority group as the non-white population. In terms of the variable ``income'', we identify low-income households based on the combination of household size and last year's household income following the 2017 Health and Human Services poverty guidelines \cite{cochran2017annual}. An individual is deemed to have a ``medical condition'' if he or she answered ``yes'' to the question ``do you have a condition or handicap that makes it difficult to travel outside of the home?'' in the survey. Regarding the protected variable ``region'', the question ``household in urban area? Answer `yes' or `no'.'' is used to identify whether the individual is an urban or rural resident. \\

\noindent The dependent variables examined in this study can be categorized into two groups: the first group contains four variables indicating the ``yes'' or ``no'' answers to ``usually work from home'', ``have the option of working from home'', ``agree that travel is a financial burden'' and ``agree that gas price affects travel''; the second group contains four variables indicating the high frequent usage of four travel modes: bike, car, bus and rideshare. These eight dependent variables are all binary variables, with ``yes'' taking the value 1. A detailed description of these variables can be found in Appendix~\ref{sec:descriptive_nhts}.\\

\noindent The distributions of the dependent variables except ``travel burden'' and ``gas price impact'' are highly skewed, and previous research has found that when the outcome class sizes are highly imbalanced, the classification algorithms tend to strongly favor the majority outcome class, resulting in very low or even no detection of the minority outcome class \cite{byon2010classification}. Therefore, for each of the six imbalanced dependent variables, we balance the data to facilitate training by downsampling the majority class. The summary statistics of the independent and dependent variables as well as the distributions of two groups of dependent variables by different protected attributes are reported in Appendix~\ref{sec:descriptive_nhts}. \\

\subsubsection{Bias mitigation method with sample weights}
\noindent As previously mentioned, one source of bias in the data is that the training data might not be representative of the overall population. Luckily, NHTS contains the sample weight\footnote{which is primarily calculated as the inverse of the probability of selection of the person in the given sampling stratum from the sampling frame} \cite{federal20182017} for each individual, which can be used to address the representation bias. We incorporate the sample weights in the training and evaluation phases. Weighted accuracy and weighted fairness metrics are used for model evaluation. To be specific, the weighted accuracy is calculated as:
\begin{linenomath}
\begin{equation}
\begin{aligned}
Weighted\:Accuracy= \frac{\sum_{i}^{N}1(\hat{y_{i}}=y_{i})w_i}{\sum_{i}^{N}w_i}
\end{aligned}
\label{equ}
\end{equation}
\end{linenomath}
where $w_{i}$ represents the sample weight for sample $i$, $y_{i}$ is the label and $\hat{y_{i}}$ is the predicted outcome. Similarly, the sample weight is applied for each sample when calculating fairness metrics (FNR and FPR). $N$ denotes the sample size.\\

\noindent Corresponding to the weighted evaluation metrics, the sample weights are also applied in the loss function during the training process. The weighted loss function is written as:
\begin{gather}
\min_{p} \quad (1-\lambda)L_{primary}+\lambda |Corr(p(\boldsymbol{x}),z|y=q)|\label{loss_weight}
\shortintertext{where}
L_{primary}=\sum_{i=1}^{N}\frac{w_{i}L_{i}}{\sum_{i}^{N}w_i}\: ; \:L_i=-y_{i} log(p(\boldsymbol{x_{i}}))-(1-y_{i})log(1-p(\boldsymbol{x_{i}})) \\
Corr(p(\boldsymbol{x}),z|y=q)=\frac{\sum_{i \in S_{q}} w_{i}(p(\boldsymbol{x_i})-m(p(\boldsymbol{x_i})))(z_i-m(z_i))}{(\sqrt{\sum_{i \in S_{q}} w_i (p(\boldsymbol{x_i})-m(p(\boldsymbol{x_i})))^2}+\epsilon)*
(\sqrt{\sum_{i \in S_{q}} w_i (z_i-m(z_i))^2}+\epsilon)}\\
m(p(\boldsymbol{x_i}))=\frac{\sum_{i \in S_{q}}w_i p(\boldsymbol{x_i})}{\sum_{i \in S_{q}}w_i}\: ; \:
m(z_i)=\frac{\sum_{i \in S_{q}}w_i z_i}{\sum_{i \in S_{q}}w_i}\\
S_{q}={\{i|y_i=q\}\: ; \:\epsilon=e^{-20}}
\qedhere
\end{gather}
BLR is implemented through the scikit-learn library when evaluating fairness issues and through TensorFlow when mitigating the bias. For DNN, all of the experiments are implemented in TensorFlow. Each of the experiments conducted in TensorFlow uses the mini-batch gradient descent method with a step size of 0.0001 during training. The best model among the 5000 epochs is chosen and later performs prediction over the test data. 5-fold validation is conducted for each experiment.\\

\subsubsection{Fairness issues in the adoption of BLR and DNN}
\begin{figure}
\begin{subfigure}[b]{.32\linewidth}
\centering
\includegraphics[width=\linewidth]{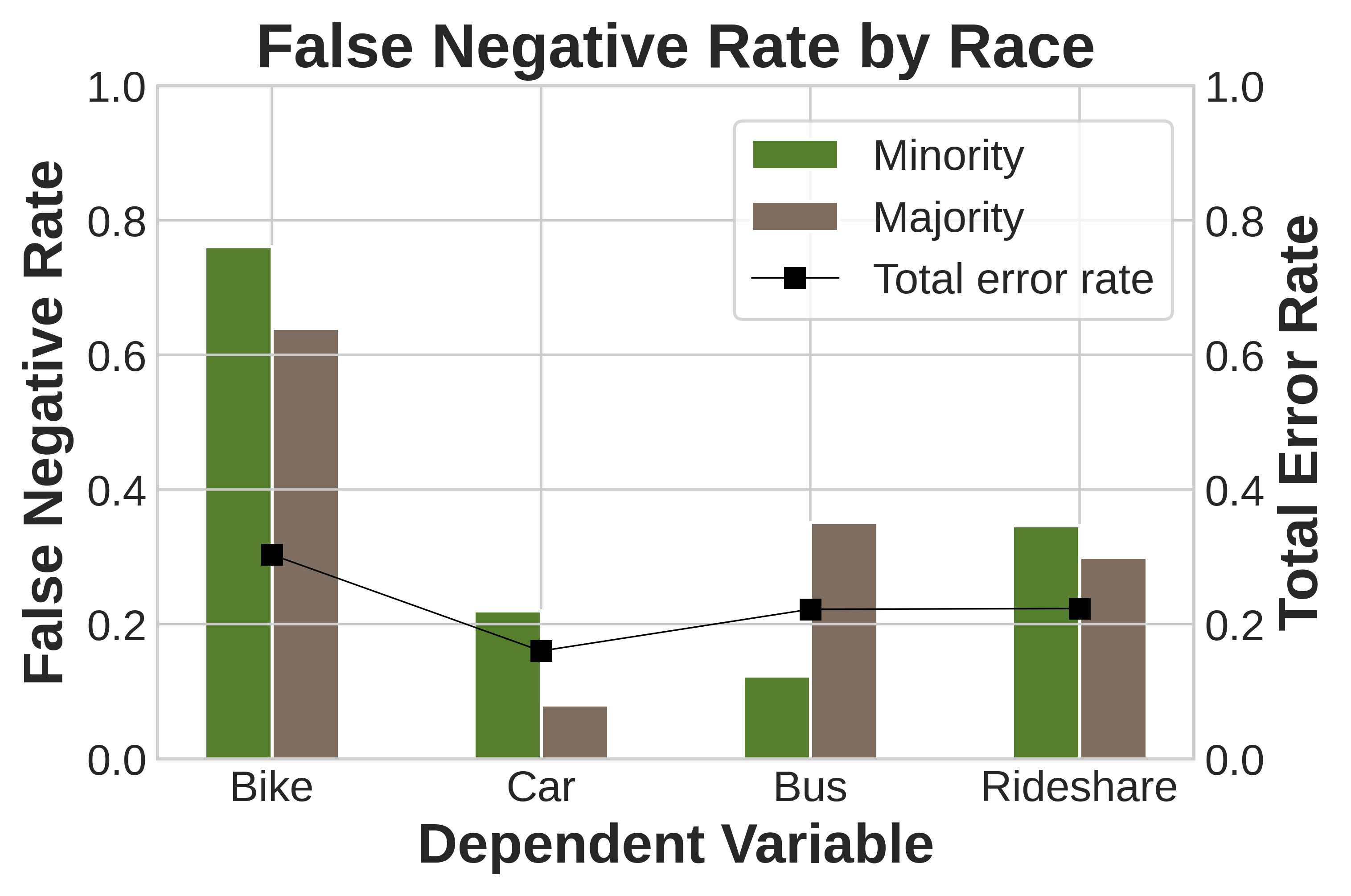}
\caption{}\label{hist_mode_logit_race}
\end{subfigure}\hfill
\begin{subfigure}[b]{.32\linewidth}
\centering
\includegraphics[width=\linewidth]{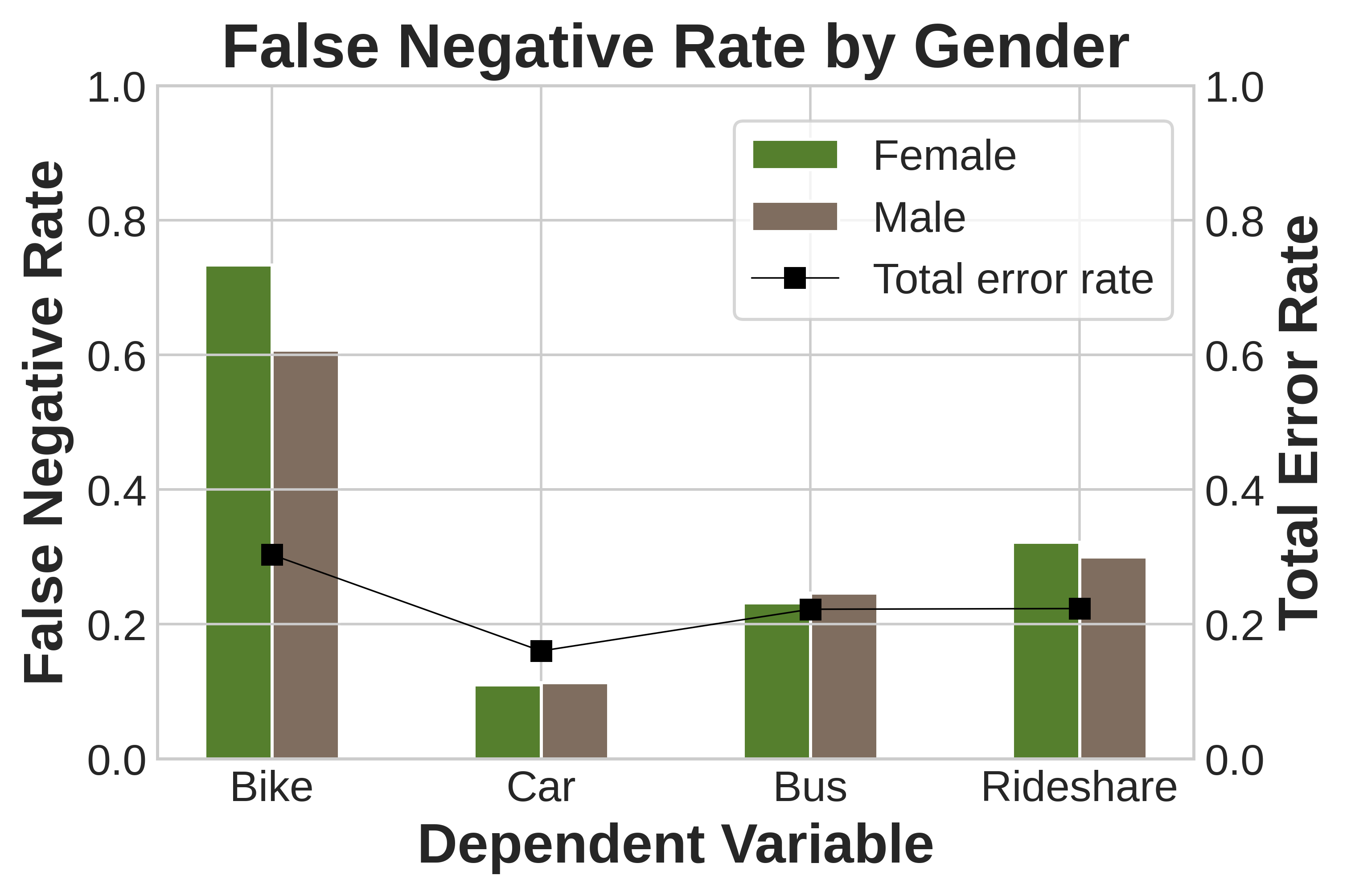}
\caption{}\label{hist_mode_logit_female}
\end{subfigure}\hfill
\begin{subfigure}[b]{.32\linewidth}
\centering
\includegraphics[width=\linewidth]{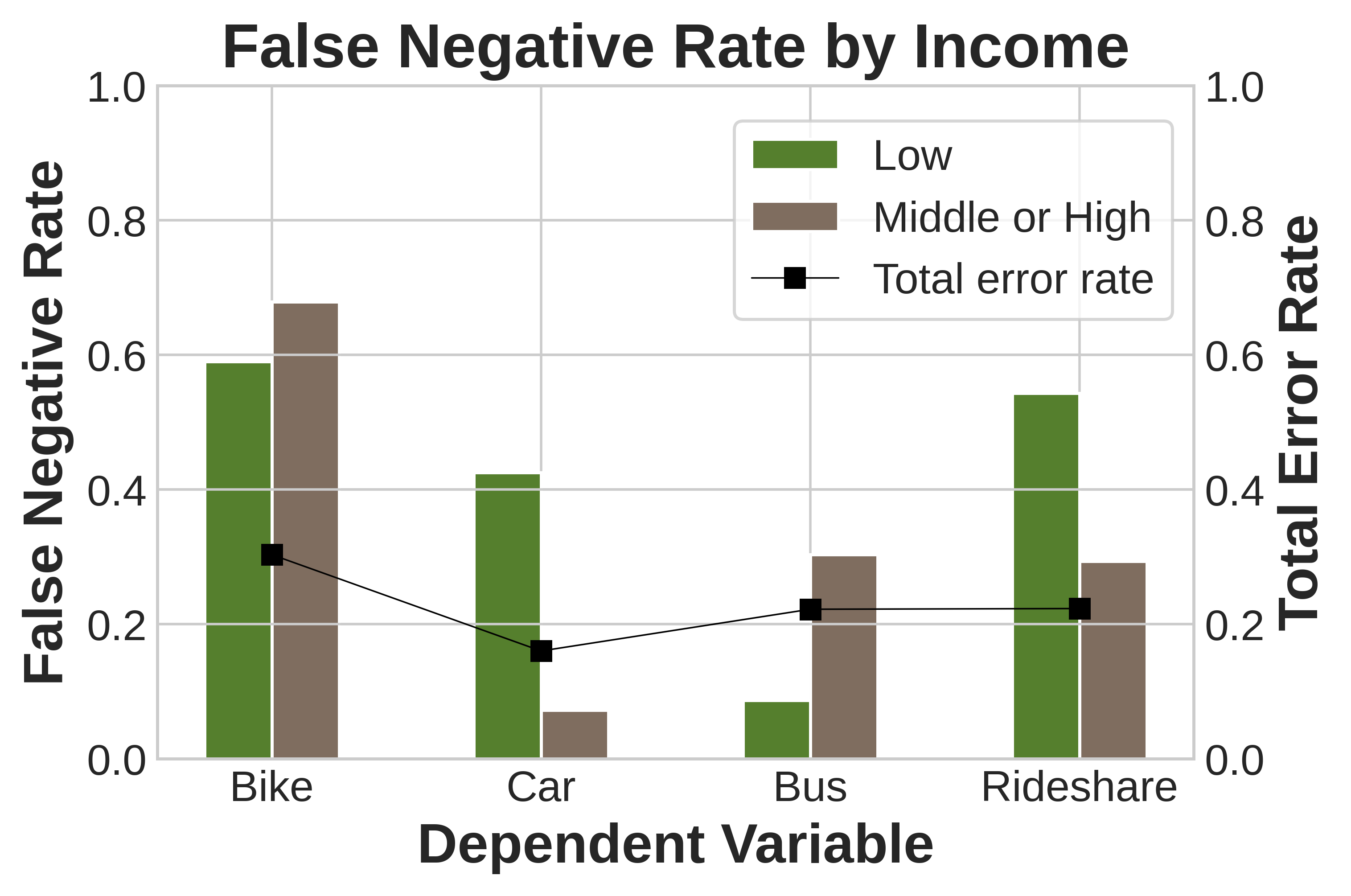}
\caption{}\label{hist_mode_logit_poverty}
\end{subfigure}\\
\begin{subfigure}[b]{.32\linewidth}
\centering
\end{subfigure}\hfill
\begin{subfigure}[b]{.32\linewidth}
\centering
\includegraphics[width=\linewidth]{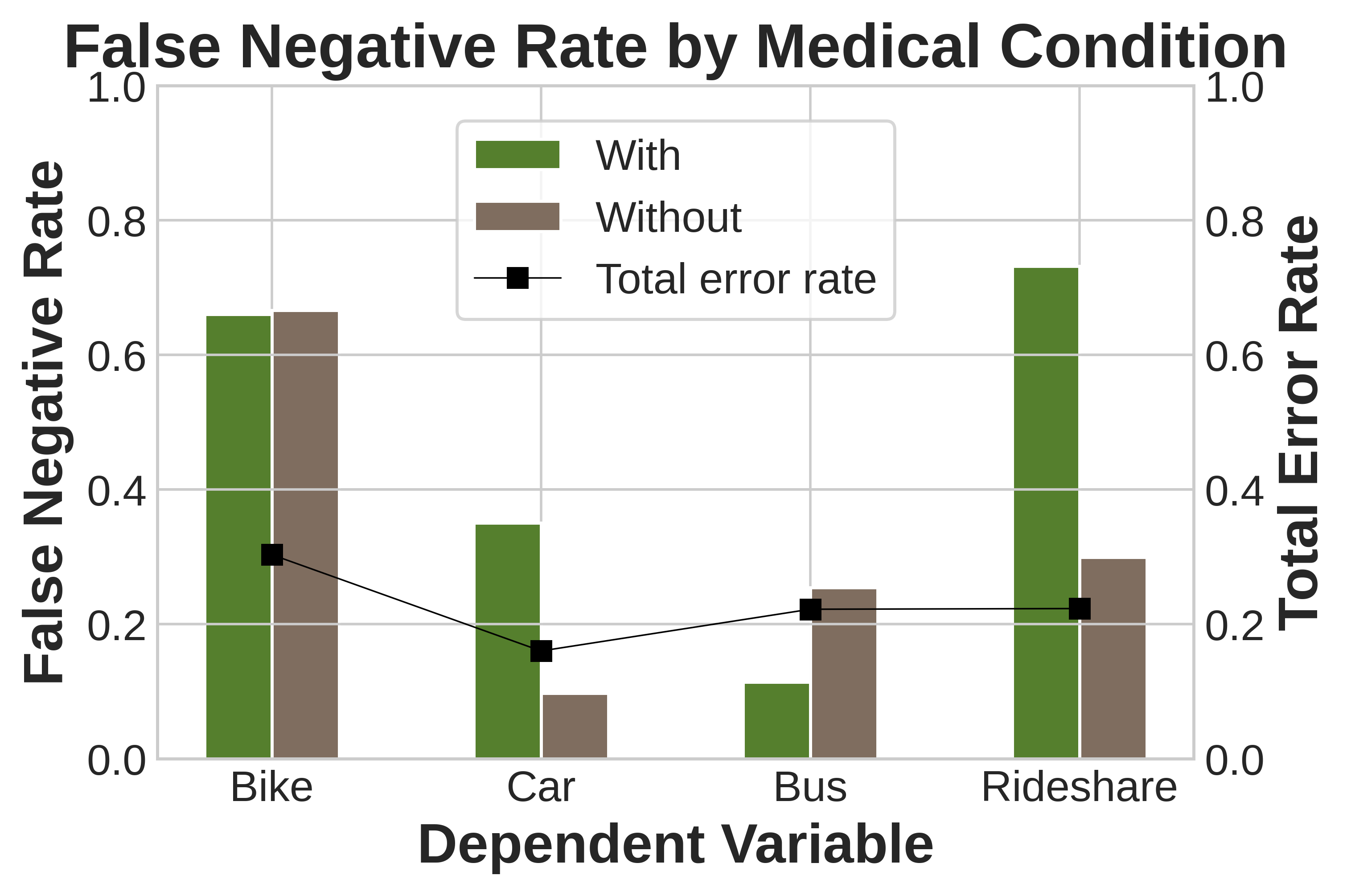}
\caption{}\label{hist_mode_logit_med}
\end{subfigure}\hfill
\begin{subfigure}[b]{.32\linewidth}
\centering
\includegraphics[width=\linewidth]{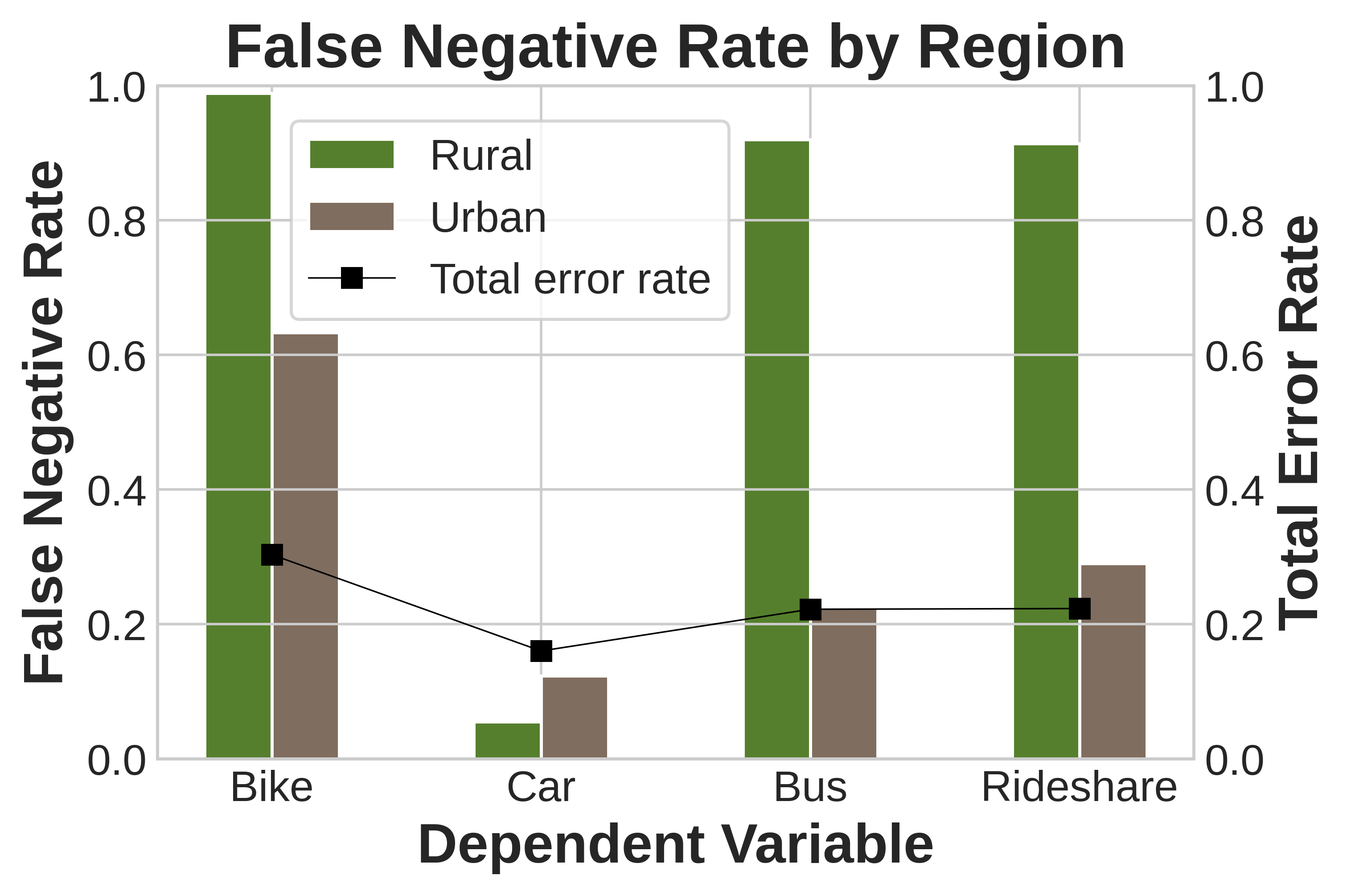}
\caption{}\label{hist_mode_logit_rural}
\end{subfigure}\hfill
\begin{subfigure}[b]{.32\linewidth}
\centering
\end{subfigure}\hfill

\caption{Disparity of prediction accuracy (BLR): frequent usage of bike, car, bus and rideshare}
\label{figure_hist_logit_mode}
\end{figure}

\begin{figure}
\begin{subfigure}[b]{.32\linewidth}
\centering
\includegraphics[width=\linewidth]{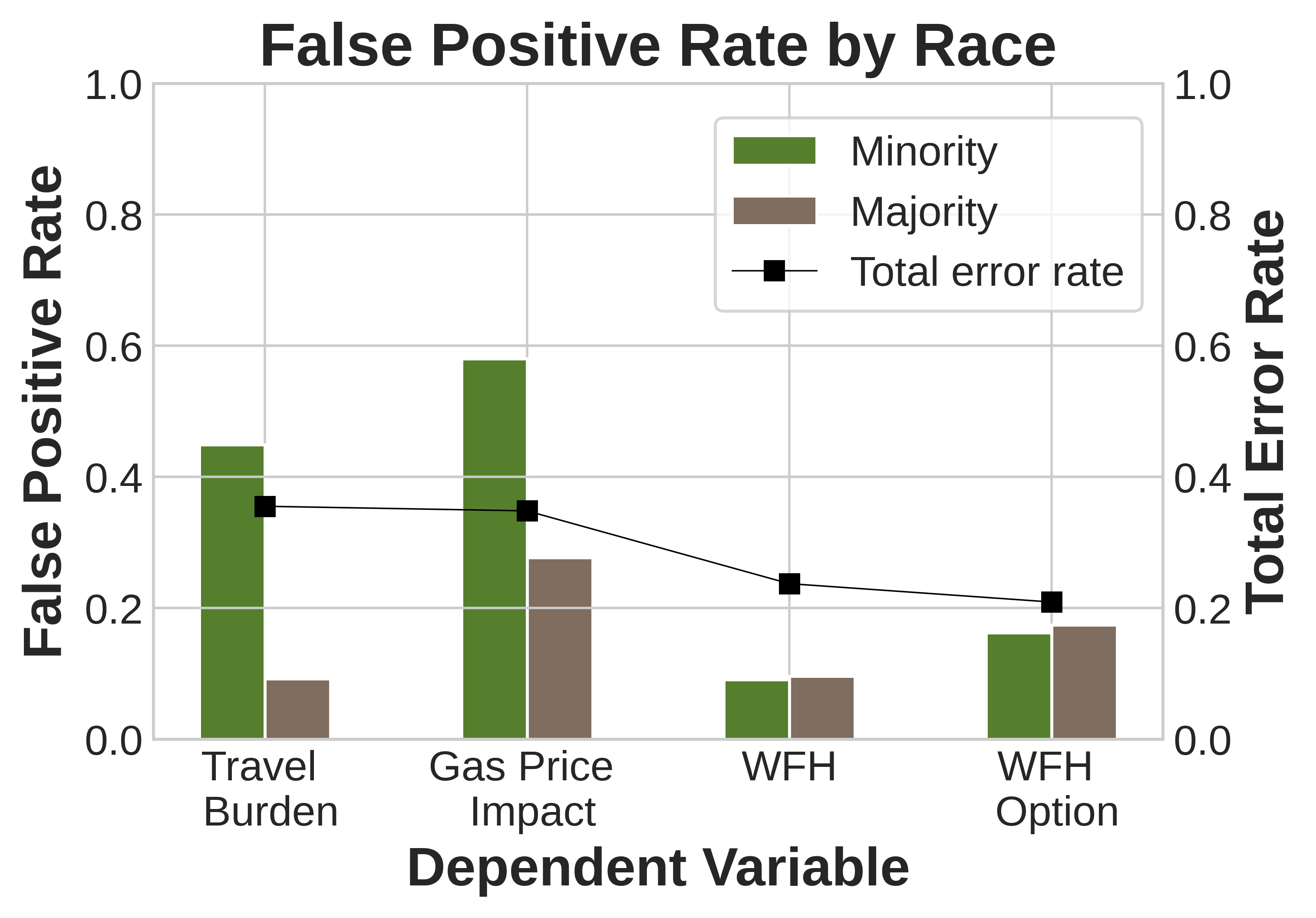}
\caption{}\label{hist_wfh_logit_race}
\end{subfigure}\hfill
\begin{subfigure}[b]{.32\linewidth}
\centering
\includegraphics[width=\linewidth]{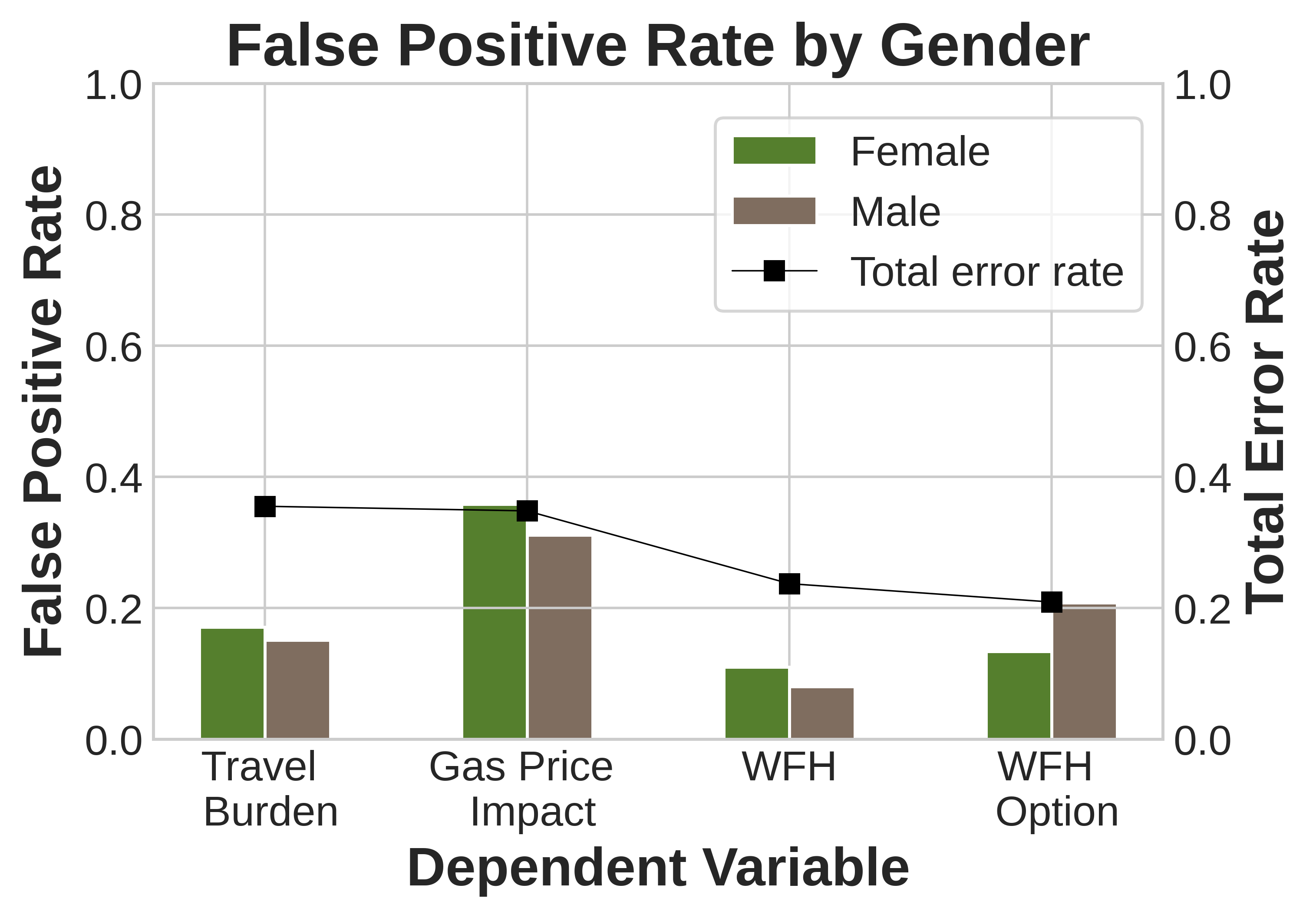}
\caption{}\label{hist_wfh_logit_female}
\end{subfigure}\hfill
\begin{subfigure}[b]{.32\linewidth}
\centering
\includegraphics[width=\linewidth]{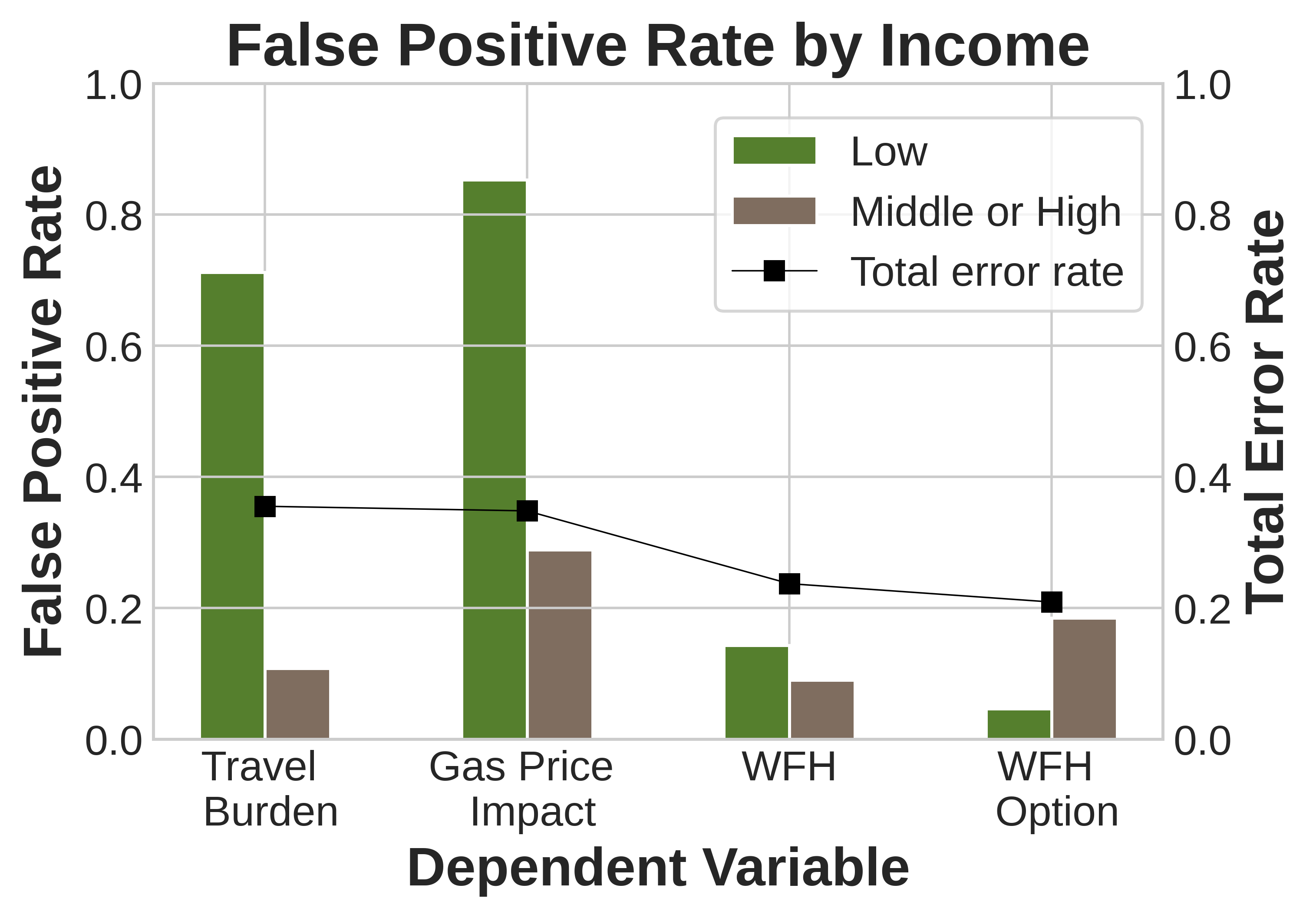}
\caption{}\label{hist_wfh_logit_poverty}
\end{subfigure}\\
\begin{subfigure}[b]{.32\linewidth}
\centering
\end{subfigure}\hfill
\begin{subfigure}[b]{.32\linewidth}
\centering
\includegraphics[width=\linewidth]{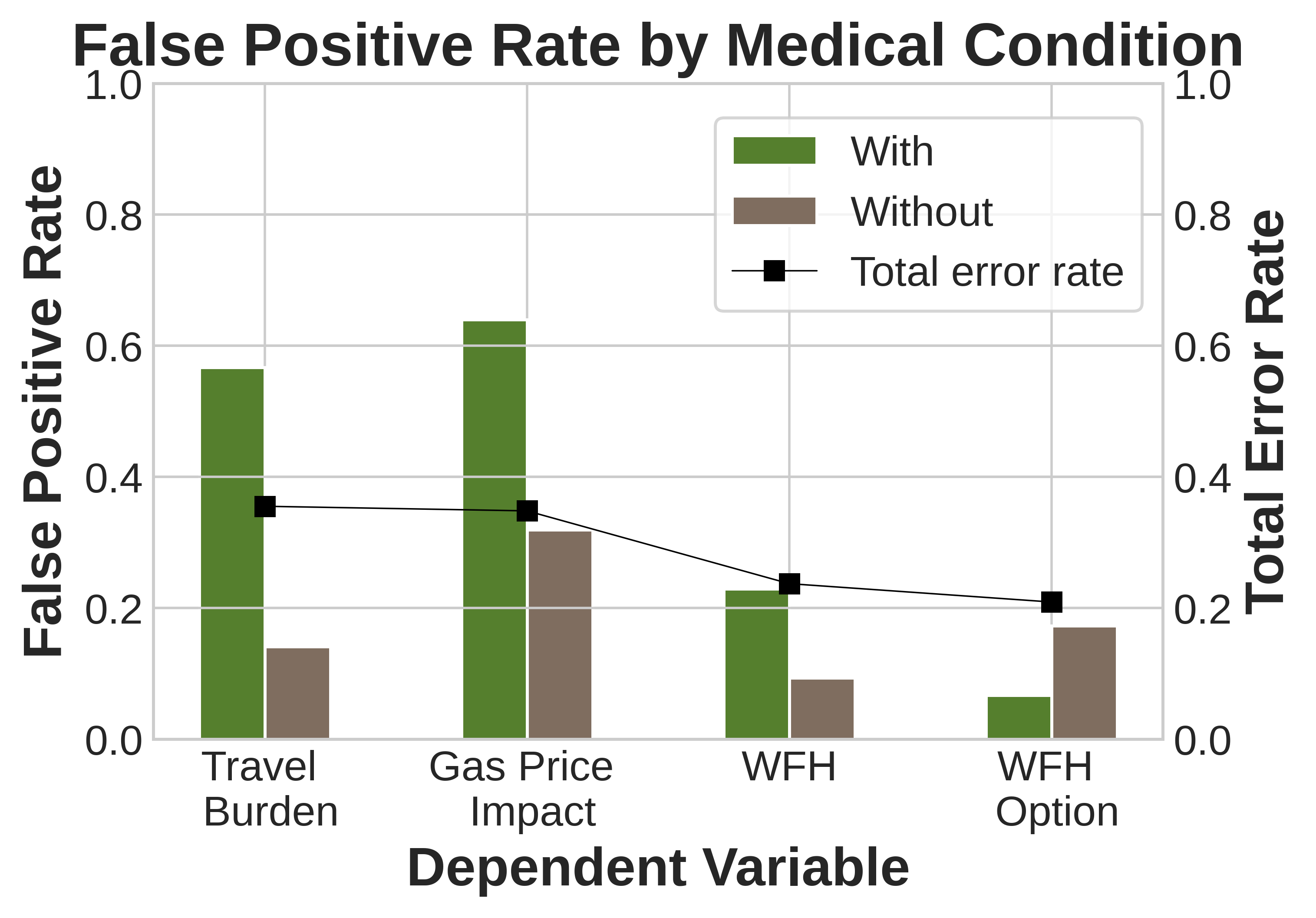}
\caption{}\label{hist_wfh_logit_med}
\end{subfigure}\hfill
\begin{subfigure}[b]{.32\linewidth}
\centering
\includegraphics[width=\linewidth]{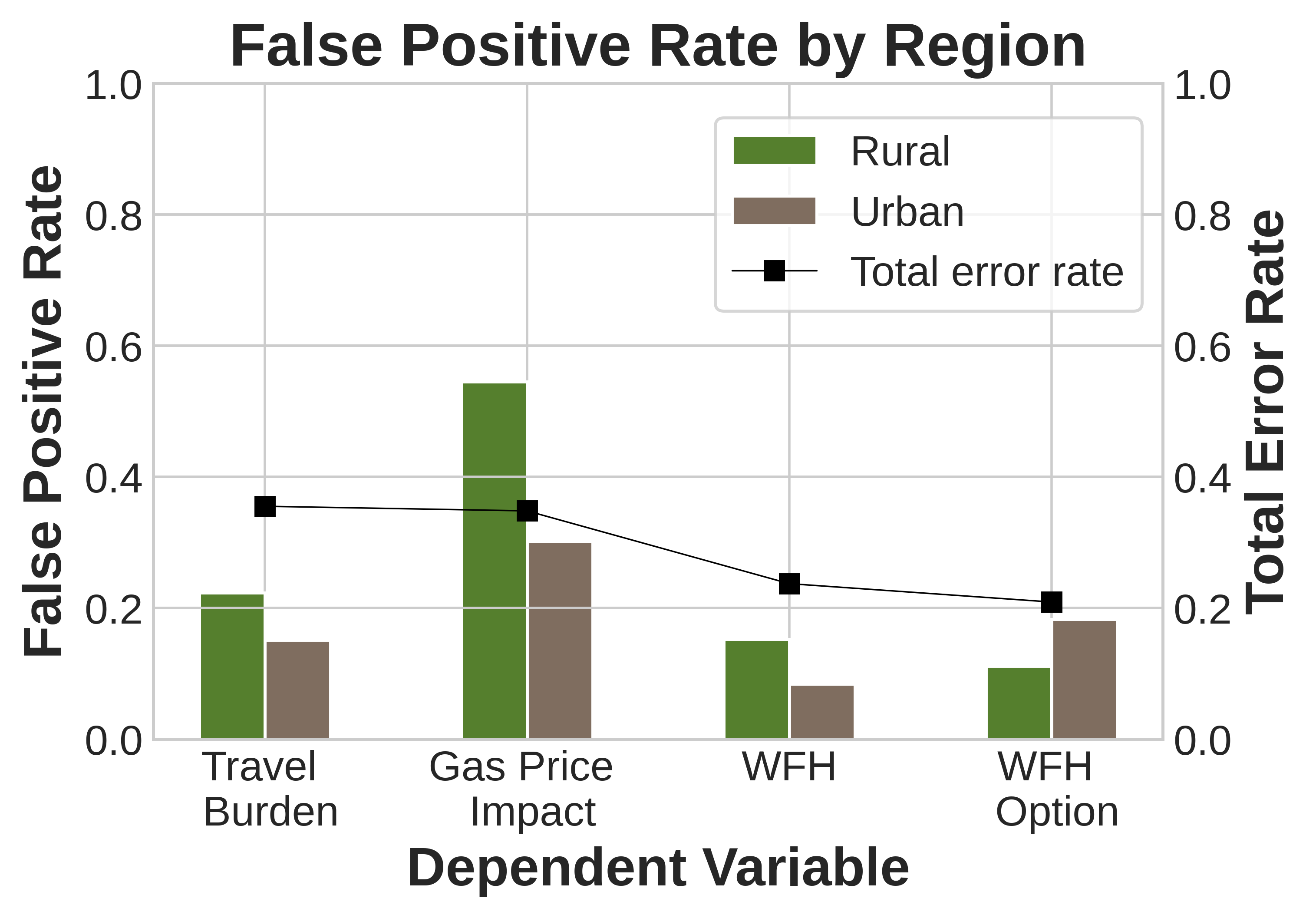}
\caption{}\label{hist_wfh_logit_rural}
\end{subfigure}\hfill
\begin{subfigure}[b]{.32\linewidth}
\centering
\end{subfigure}\hfill

\caption{Disparity of prediction accuracy (BLR): work from home, work from home option, travel burden, gas price impact}
\label{figure_hist_logit_wfh}
\end{figure}

\begin{figure}
\begin{subfigure}[b]{.32\linewidth}
\centering
\includegraphics[width=\linewidth]{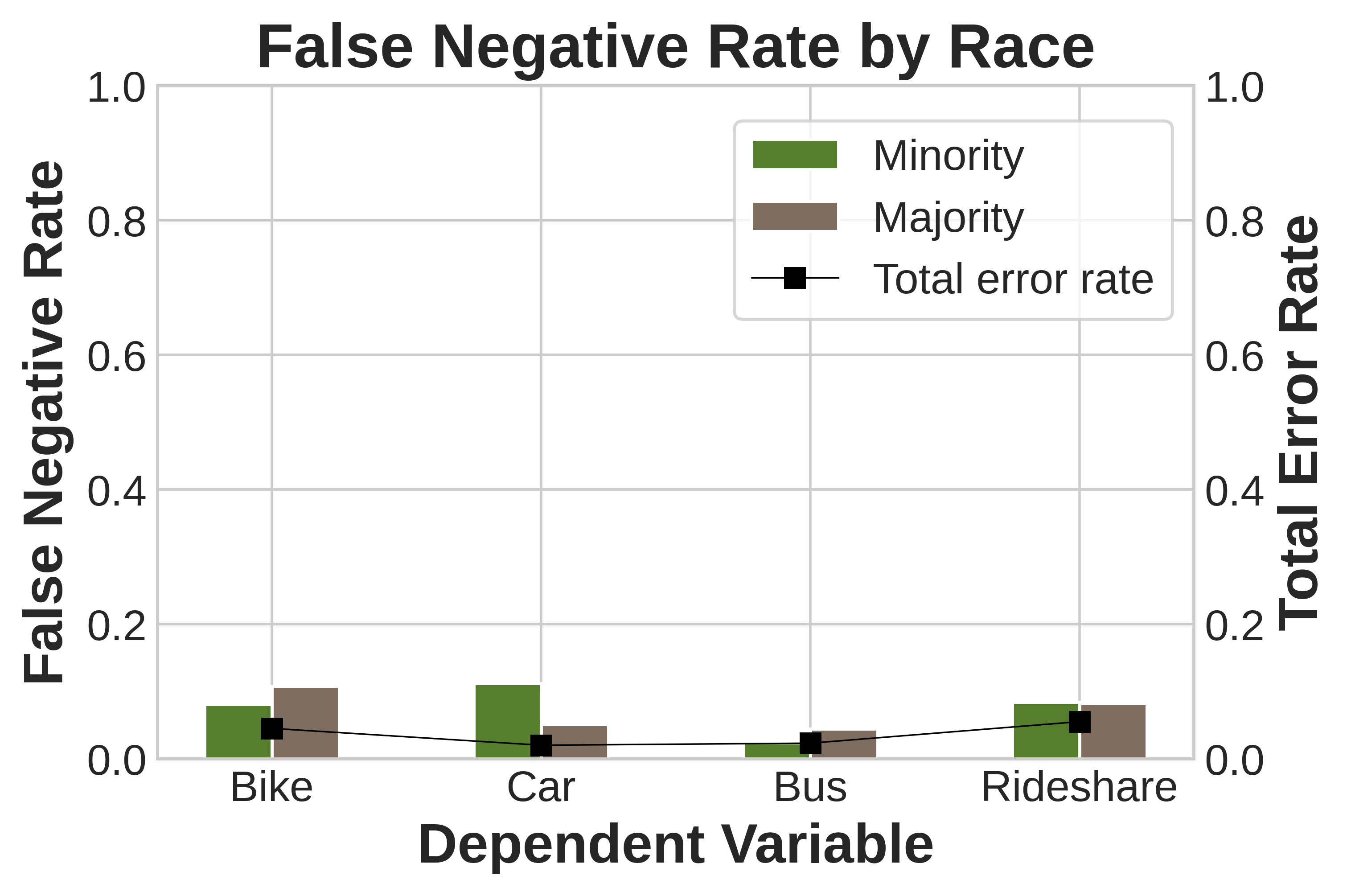}
\caption{}\label{hist_mode_dnn_race}
\end{subfigure}\hfill
\begin{subfigure}[b]{.32\linewidth}
\centering
\includegraphics[width=\linewidth]{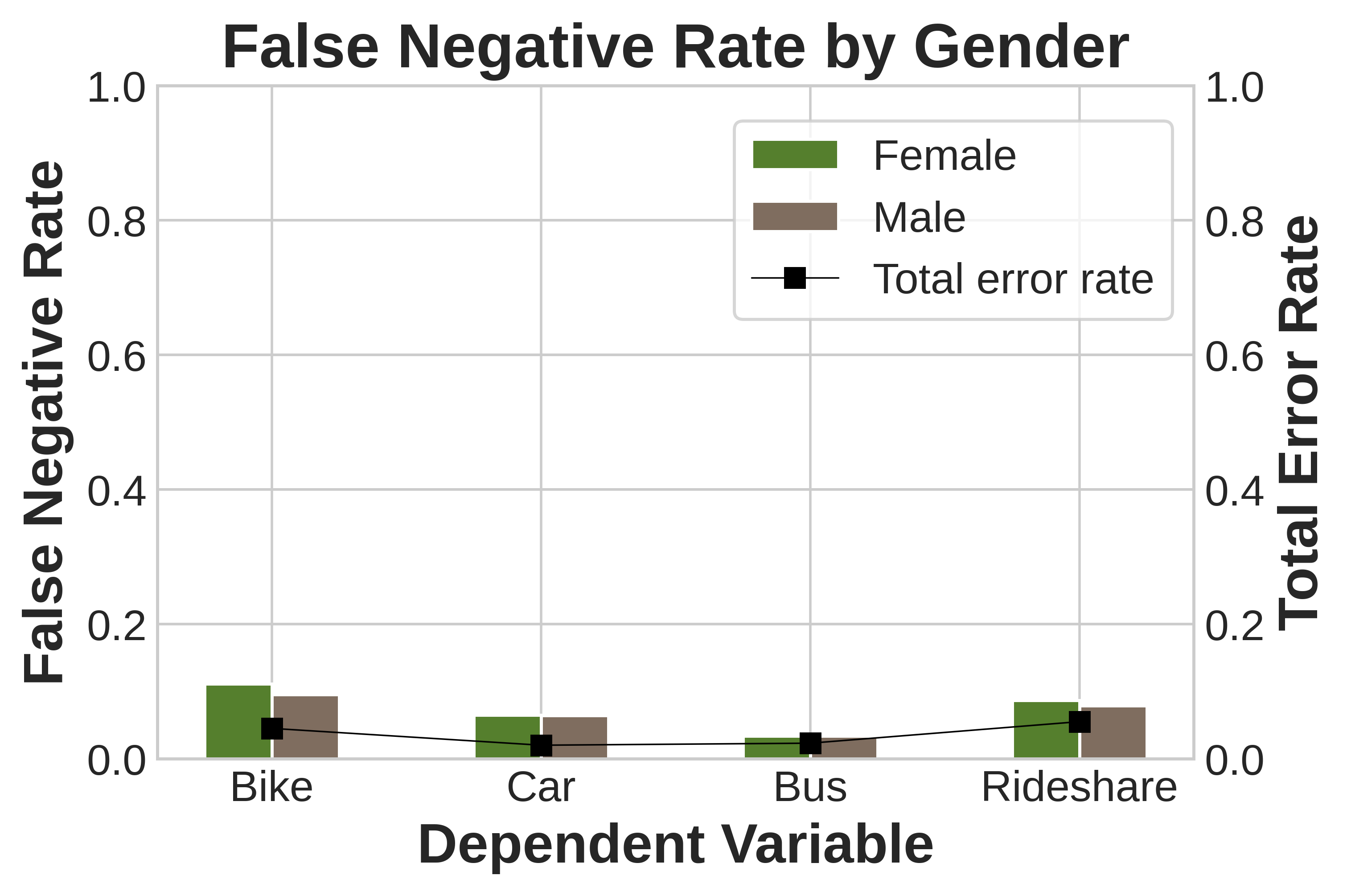}
\caption{}\label{hist_mode_dnn_female}
\end{subfigure}\hfill
\begin{subfigure}[b]{.32\linewidth}
\centering
\includegraphics[width=\linewidth]{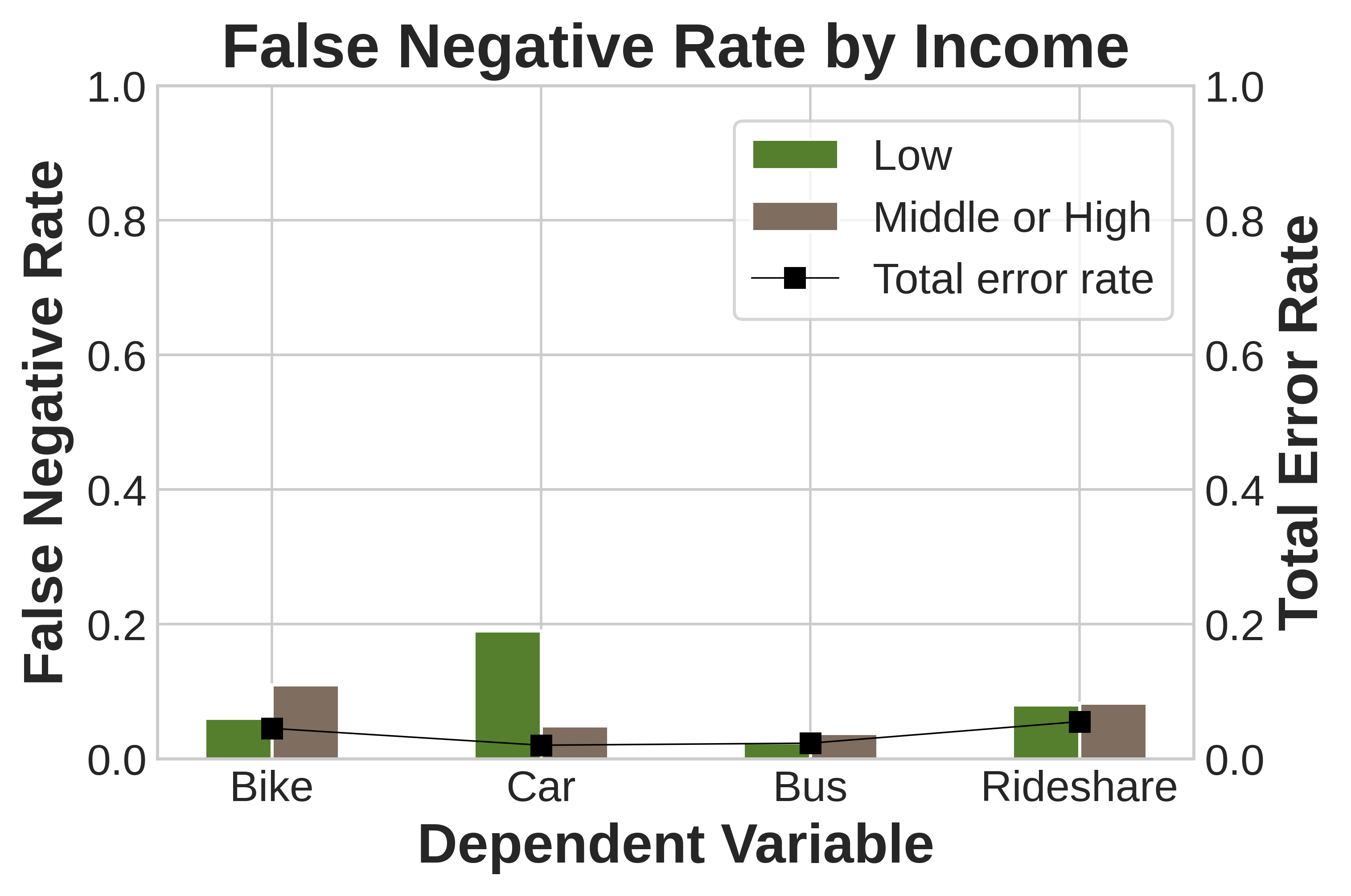}
\caption{}\label{hist_mode_dnn_poverty}
\end{subfigure}\\
\begin{subfigure}[b]{.32\linewidth}
\centering
\end{subfigure}\hfill
\begin{subfigure}[b]{.32\linewidth}
\centering
\includegraphics[width=\linewidth]{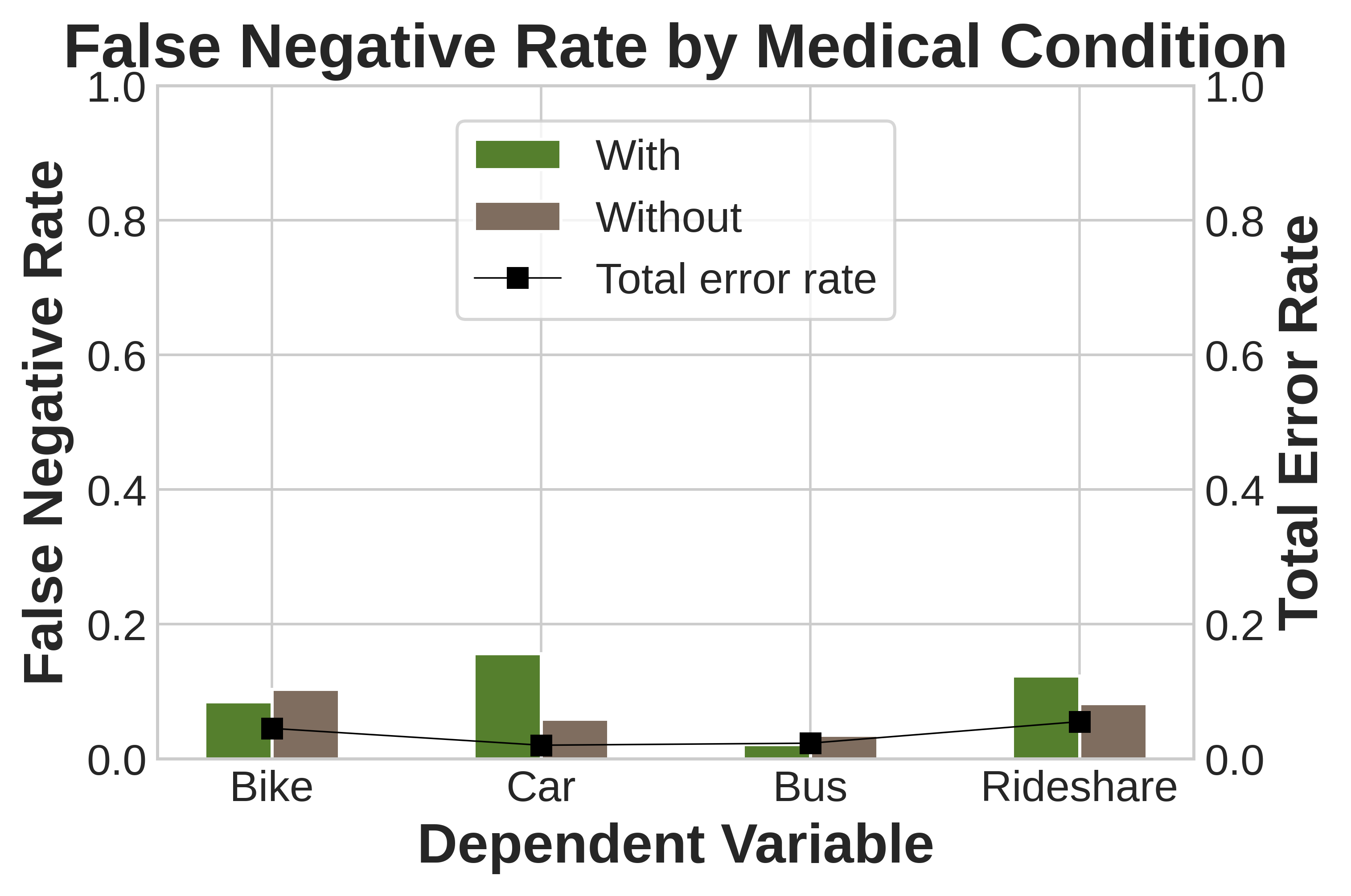}
\caption{}\label{hist_mode_dnn_med}
\end{subfigure}\hfill
\begin{subfigure}[b]{.32\linewidth}
\centering
\includegraphics[width=\linewidth]{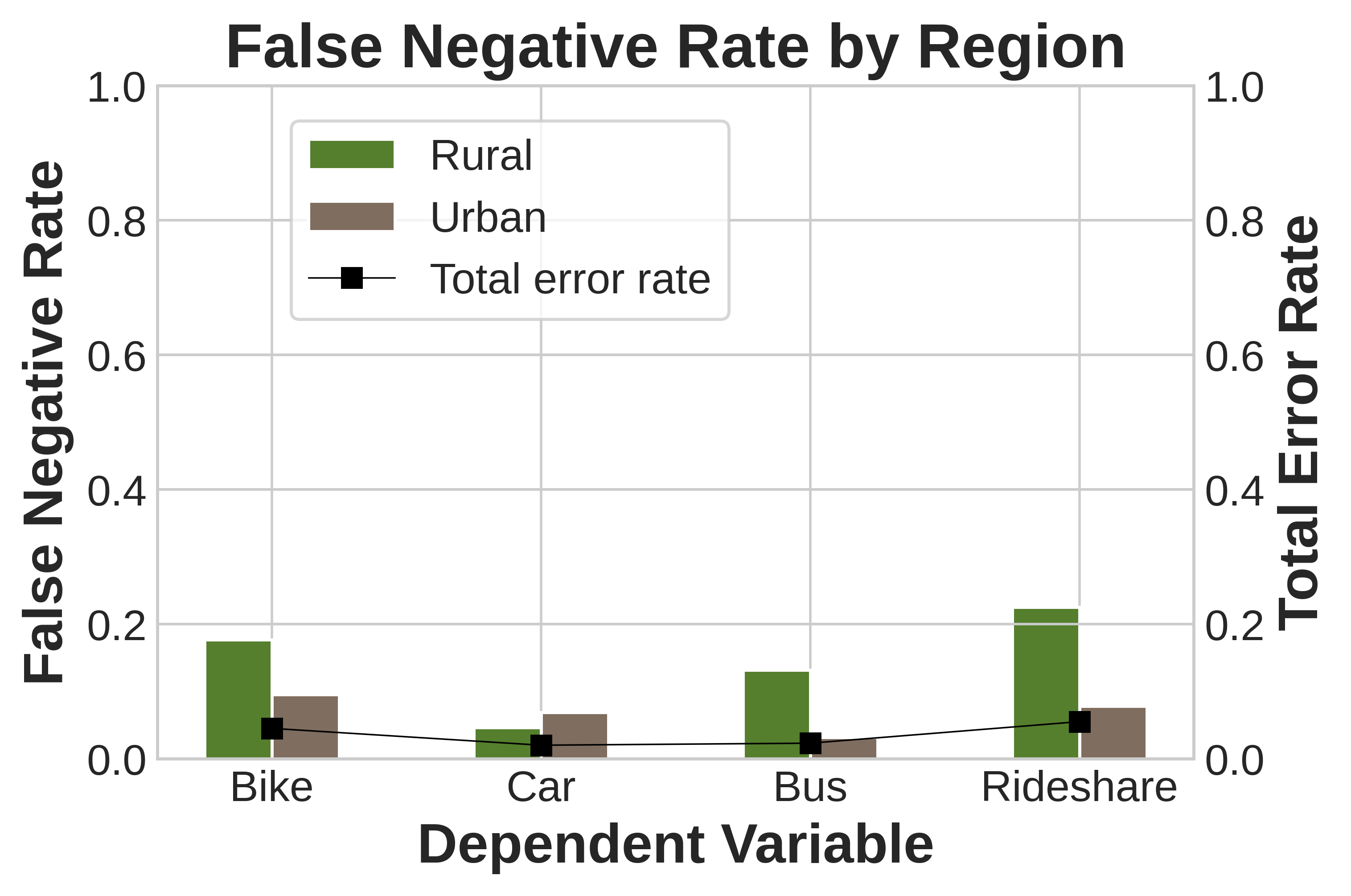}
\caption{}\label{hist_mode_dnn_rural}
\end{subfigure}\hfill
\begin{subfigure}[b]{.32\linewidth}
\centering
\end{subfigure}\hfill

\caption{Disparity of prediction accuracy (DNN): frequent usage of bike, car, bus and rideshare}
\label{figure_hist_dnn_mode}
\end{figure}

\begin{figure}
\begin{subfigure}[b]{.32\linewidth}
\centering
\includegraphics[width=\linewidth]{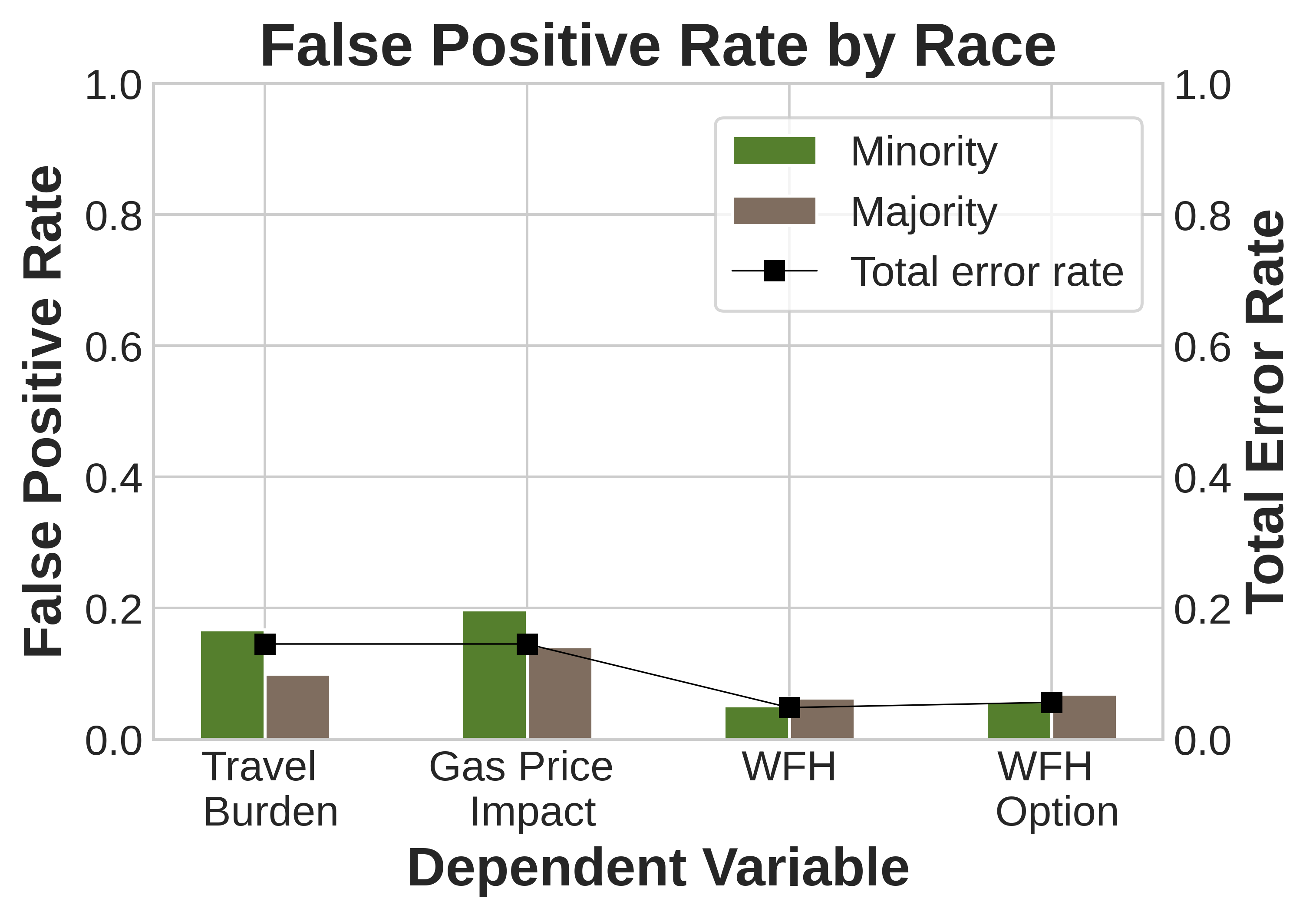}
\caption{}\label{hist_wfh_dnn_race}
\end{subfigure}\hfill
\begin{subfigure}[b]{.32\linewidth}
\centering
\includegraphics[width=\linewidth]{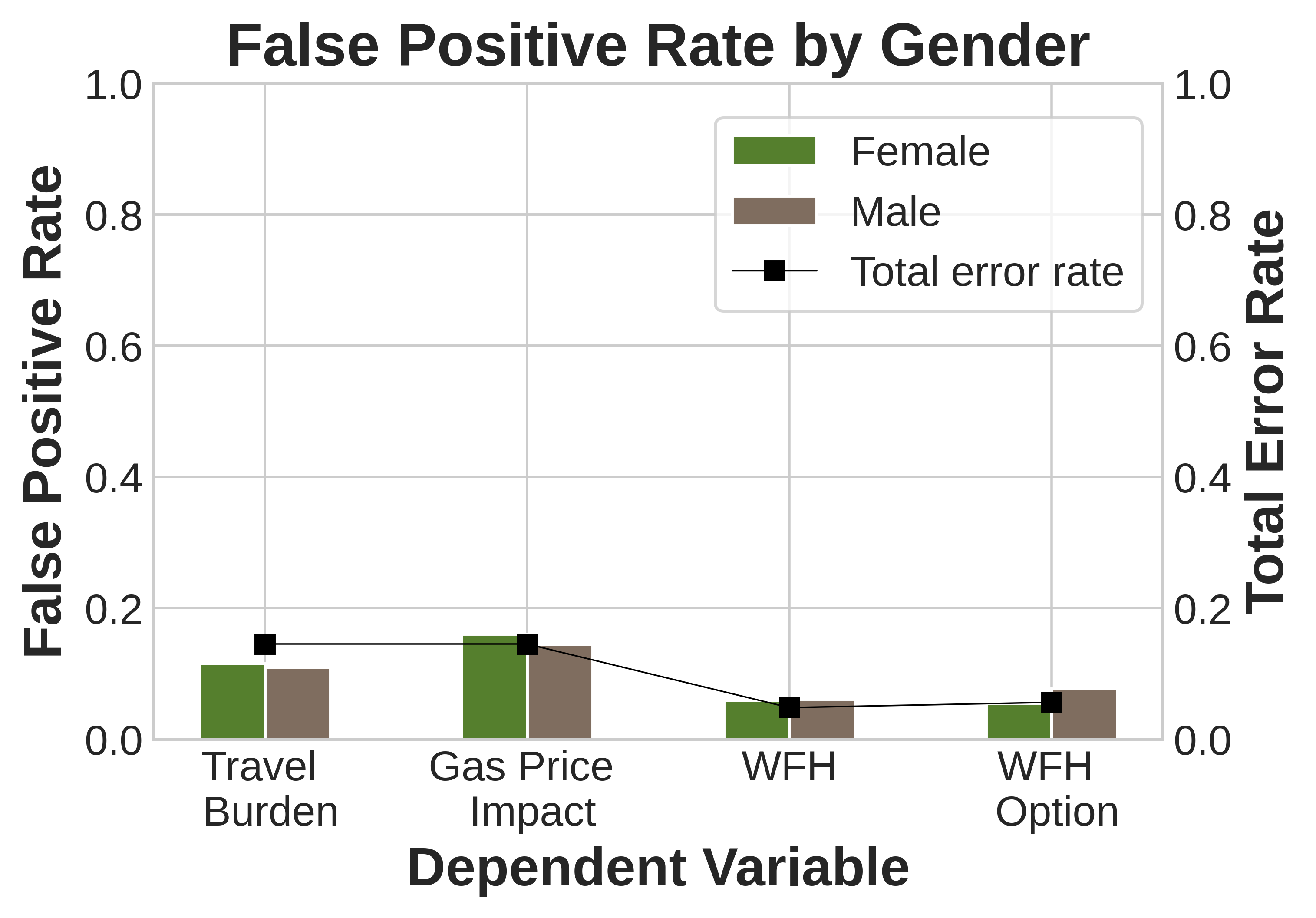}
\caption{}\label{hist_wfh_dnn_female}
\end{subfigure}\hfill
\begin{subfigure}[b]{.32\linewidth}
\centering
\includegraphics[width=\linewidth]{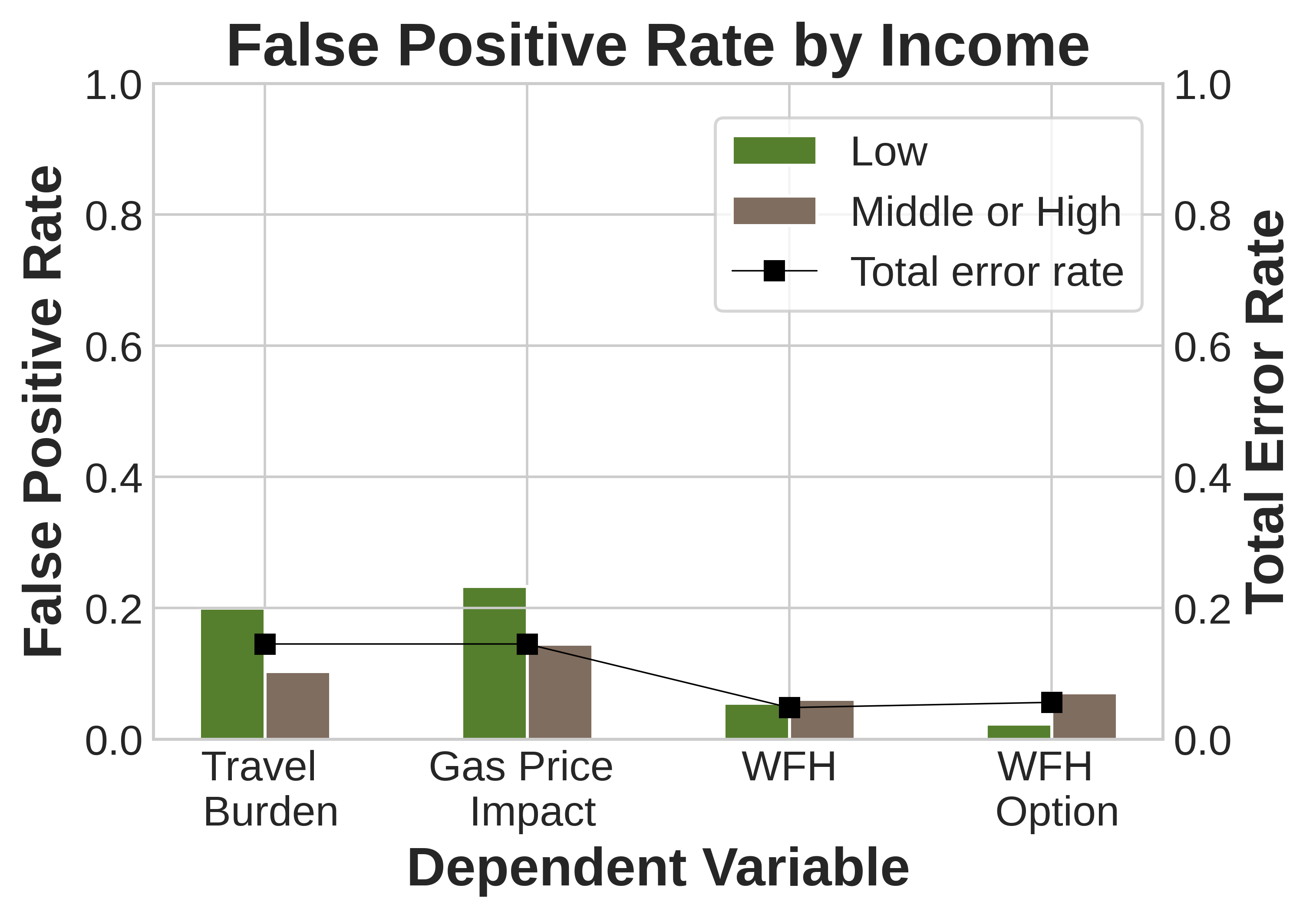}
\caption{}\label{hist_wfh_dnn_poverty}
\end{subfigure}\\
\begin{subfigure}[b]{.32\linewidth}
\centering
\end{subfigure}\hfill
\begin{subfigure}[b]{.32\linewidth}
\centering
\includegraphics[width=\linewidth]{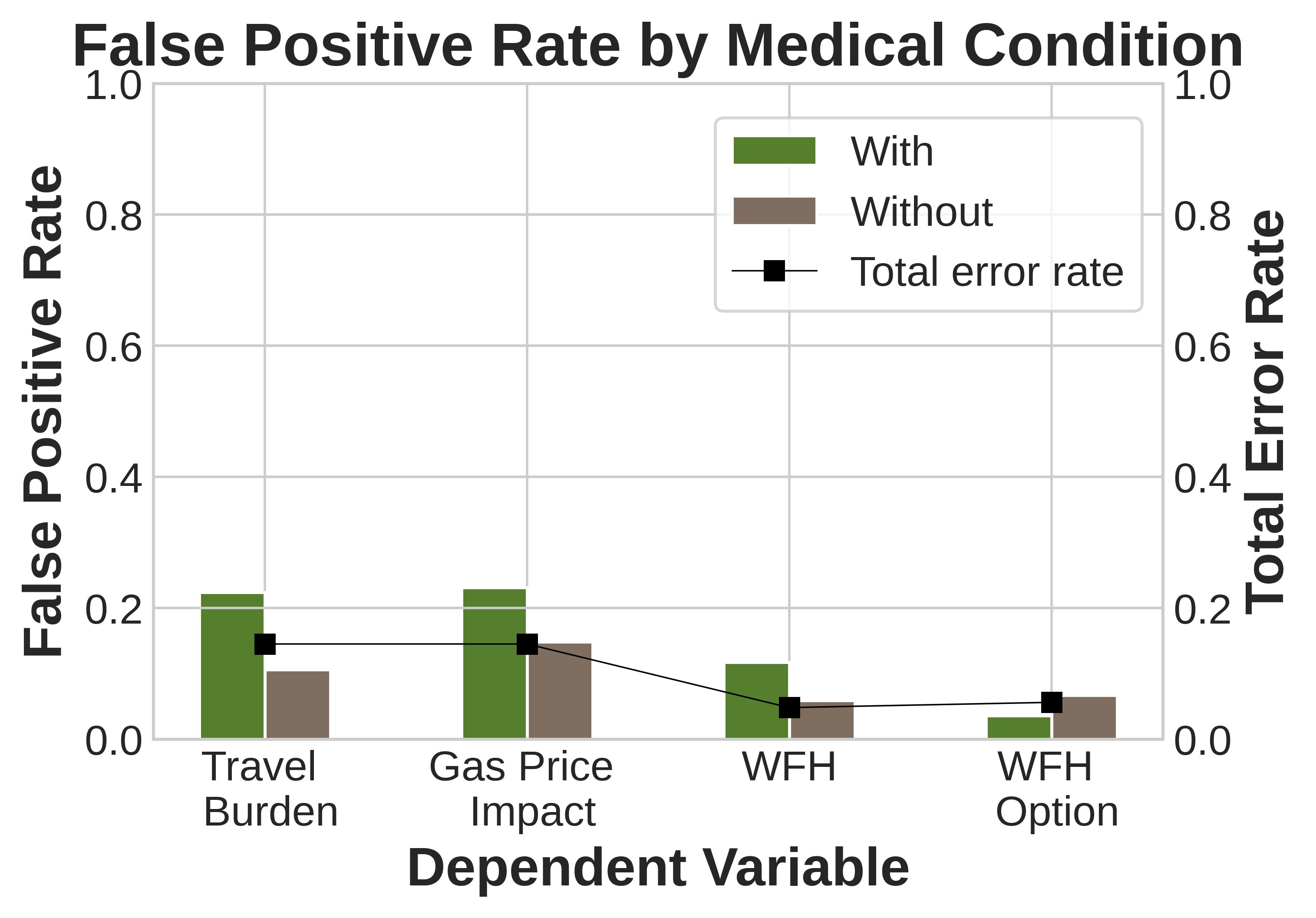}
\caption{}\label{hist_wfh_dnn_med}
\end{subfigure}\hfill
\begin{subfigure}[b]{.32\linewidth}
\centering
\includegraphics[width=\linewidth]{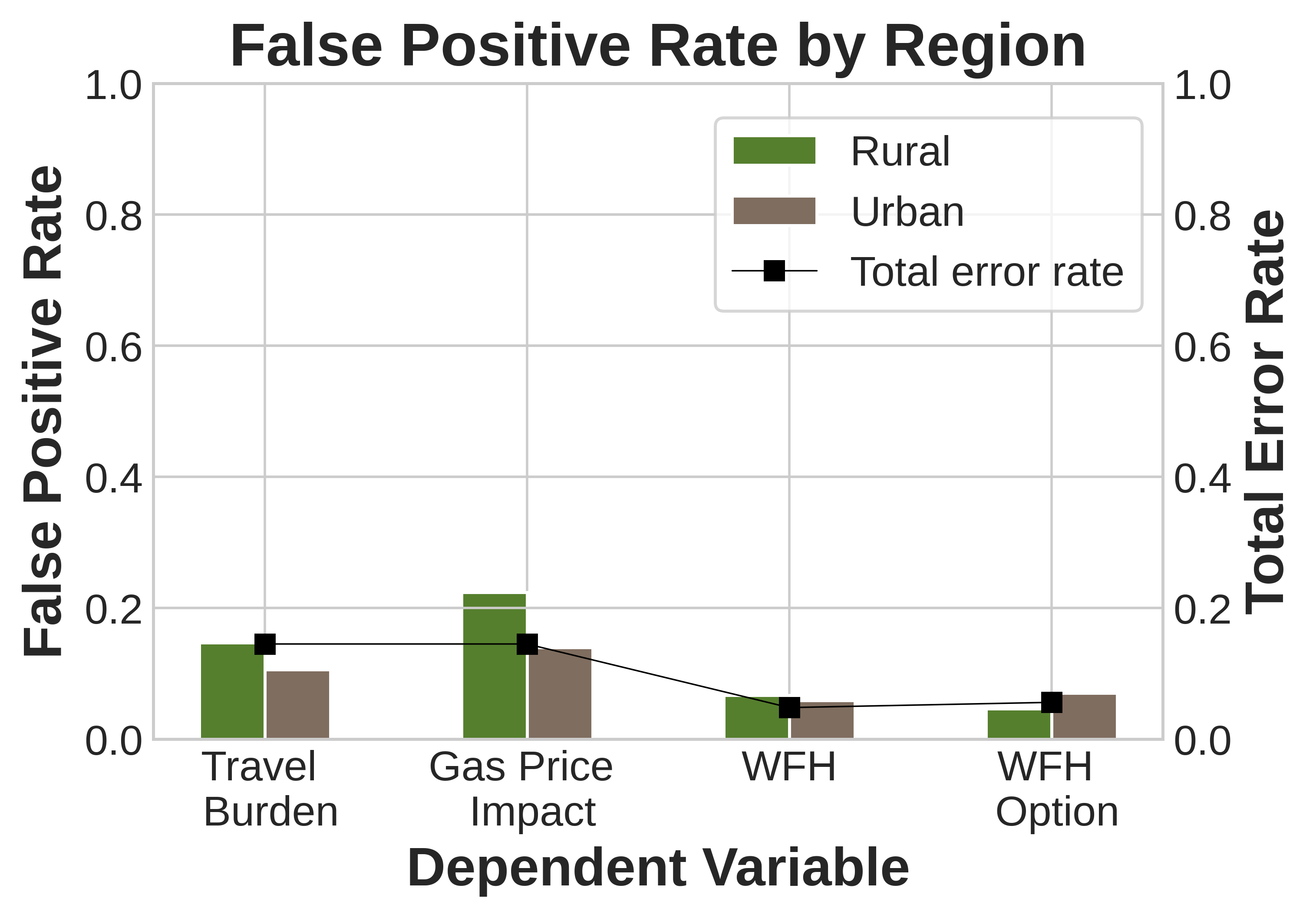}
\caption{}\label{hist_wfh_dnn_rural}
\end{subfigure}\hfill
\begin{subfigure}[b]{.32\linewidth}
\centering
\end{subfigure}\hfill

\caption{Disparity of prediction accuracy (DNN): work from home, work from home option, travel burden, gas price impact}
\label{figure_hist_dnn_wfh}
\end{figure}

The comparison of prediction accuracy with respect to various protected variables are presented by the bar charts in Figure \ref{figure_hist_logit_mode} and \ref{figure_hist_logit_wfh} for BLR and in Figure \ref{figure_hist_dnn_mode} and \ref{figure_hist_dnn_wfh} for DNN. Each bar chart depicts the prediction accuracy of two populations grouped by a specific protected variable (race, gender, income, medical condition, region) side by side. The height gap of two adjacent bars shows the prediction disparity for that protected variable. Figure \ref{figure_hist_logit_mode} and \ref{figure_hist_dnn_mode} illustrate the prediction results of the dependent variables regarding travel mode usage by BLR and DNN. Figure \ref{figure_hist_logit_wfh} and \ref{figure_hist_dnn_wfh} plot the prediction results of the dependent variables ``work from  home'',  ``work from home option'', ``travel burden'' and ``gas price impact'' by BLR and DNN. The dependent variables are specified on the x-axis of the bar charts.\\

\noindent The y-axis of the bar charts represents one of the error rates: FPR or FNR. We examine FNR for the first group of dependent variables since we are concerned about cases where active users of a certain travel mode are not identified (Figure \ref{figure_hist_logit_mode} and \ref{figure_hist_dnn_mode}), and FPR for the second group of dependent variables (Figure \ref{figure_hist_logit_wfh} and \ref{figure_hist_dnn_wfh}) since we want to focus on instances which have negative outcomes but are wrongly identified as positive (e.g. people who do not have the option of working from home but are mistakenly identified as having the option). In each bar chart, the height of a bar represents the magnitude of the class-specific FNR or FPR. The total error rate is also presented which refers to one minus the weighted accuracy for all samples.\\

\noindent First, we focus on the BLR results. Figure \ref{figure_hist_logit_mode} and \ref{figure_hist_logit_wfh} show that prediction disparities are common with the implementation of BLR. Figure \ref{figure_hist_logit_mode} presents the false negative bias across different populations regarding the frequent usage of different travel modes. The plots show that except for the fairly consistent prediction accuracy between male and female, considerable disparity in prediction accuracy exists for all other protected variables. Racial bias is large in the prediction of frequent usage of car and bus; income bias and bias towards medical condition are considerable in the predictions of car, bus and rideshare. Among all the protected attributes, the regional disparity is the largest with respect to predicting the frequent usage of bike, bus and rideshare. As a result, the proportion of rural residents that use bike, bus and rideshare frequently is underestimated compared with that of the urban residents. If policy makers use the modeling results to inform transportation resource allocations such as the planning of bike lanes and bus routes without considering the prediction bias, the rural area will very likely be under-served.\\

\noindent In terms of the dependent variables, when estimating high frequent usage of rideshare, we find that for all protected variables, the disadvantaged group always has higher FNR than the other group with either BLR or DNN. This finding indicates that the communities where these disadvantaged social groups (ethnic minorities, women, low-income groups, people who have medical conditions and rural residents) dominate could experience less ride sharing service if decision makers use these data for rideshare demand estimation and do not account for the prediction disparity.\\

\noindent Figure \ref{figure_hist_logit_wfh} illustrates significant racial bias, income bias and health-related bias for the predictions of ``travel burden'' and ``gas price impact''. In other words, the ethnic minority population, the low-income population and people with health conditions are more likely to be predicted as ``regarding travel as a financial burden'' and ``agreeing that price of gasoline affects travel'' when the true outcome is actually negative. These biases could lead to disadvantageous consequences for vulnerable populations. For example, given that banks are less likely to loan to someone if they perceive that person to be under financial stress, if the ethnic minorities suffer from higher FPR in travel burden prediction, they are more likely to be rejected when applying for loans.\\

\noindent Next, we compare the results between BLR and DNN. The prediction disparity and the total error rate of BLR are considerably larger than those of DNN in all scenarios, showing that the prediction disparity of BLR largely comes from model misspecifications. By reducing model misspecification, DNN can not only increase prediction accuracy, but also improve computational fairness. These findings are consistent with our simulation results as illustrated in Figure \ref{figure_syn2}, which show the positive relationship between the magnitude of the fairness gap and the degree of model misspecifications. \\

\noindent However, the prediction results of DNN (Figure \ref{figure_hist_dnn_mode} and Figure \ref{figure_hist_dnn_wfh}) also show that even if the prediction achieves very high accuracy (>94\% for all dependent variables in Figure \ref{figure_hist_dnn_mode} and >85\% for all dependent variables in Figure \ref{figure_hist_dnn_wfh}), the fairness gap still exists, and for some sensitive variables the prediction disparity can be higher than 15\% (e.g. the FNR gap for the frequent usage of rideshare prediction between rural and urban residents). This finding tells us that deploying complex models aimed at improving prediction accuracy cannot guarantee to eliminate prediction biases. Therefore, a dedicated method should be deployed to mitigate unintended bias rather than merely attempting to improve the model prediction power. \\

\noindent Besides, with the exception of the racial difference in bike frequency prediction, the signs of the fairness gaps for other protected variable and dependent variable combinations are the same between BLR and DNN. The consistency between the two models suggest that the prediction disparity may reveal bias inherent in the existing inequality in society which is baked into the data. For example, in the predictions of mode usage, it is found that the socially disadvantaged groups such as the ethnic minority group and people having medical conditions tend to suffer from higher FNR in predictions of car and rideshare usage, probably because they are negatively associated with certain factors (e.g. income) that positively contribute to the usage of these modes which are more expensive compared with other modes.\\

\noindent We also conduct the sensitivity analysis by changing the number of layers in DNN, the batch size and the weight initialization. The results are presented in Appendix~\ref{sec:sensitivity}, which show that the fairness gaps generally still exist with these variants, and in most cases, changing these hyperparameters does not change the signs of the fairness gaps.
\subsubsection{Bias mitigation results}
\begin{figure}
\begin{subfigure}[b]{.32\linewidth}
\centering
\end{subfigure}\hfill
\begin{subfigure}[b]{.4\linewidth}
\centering
\includegraphics[width=\linewidth]{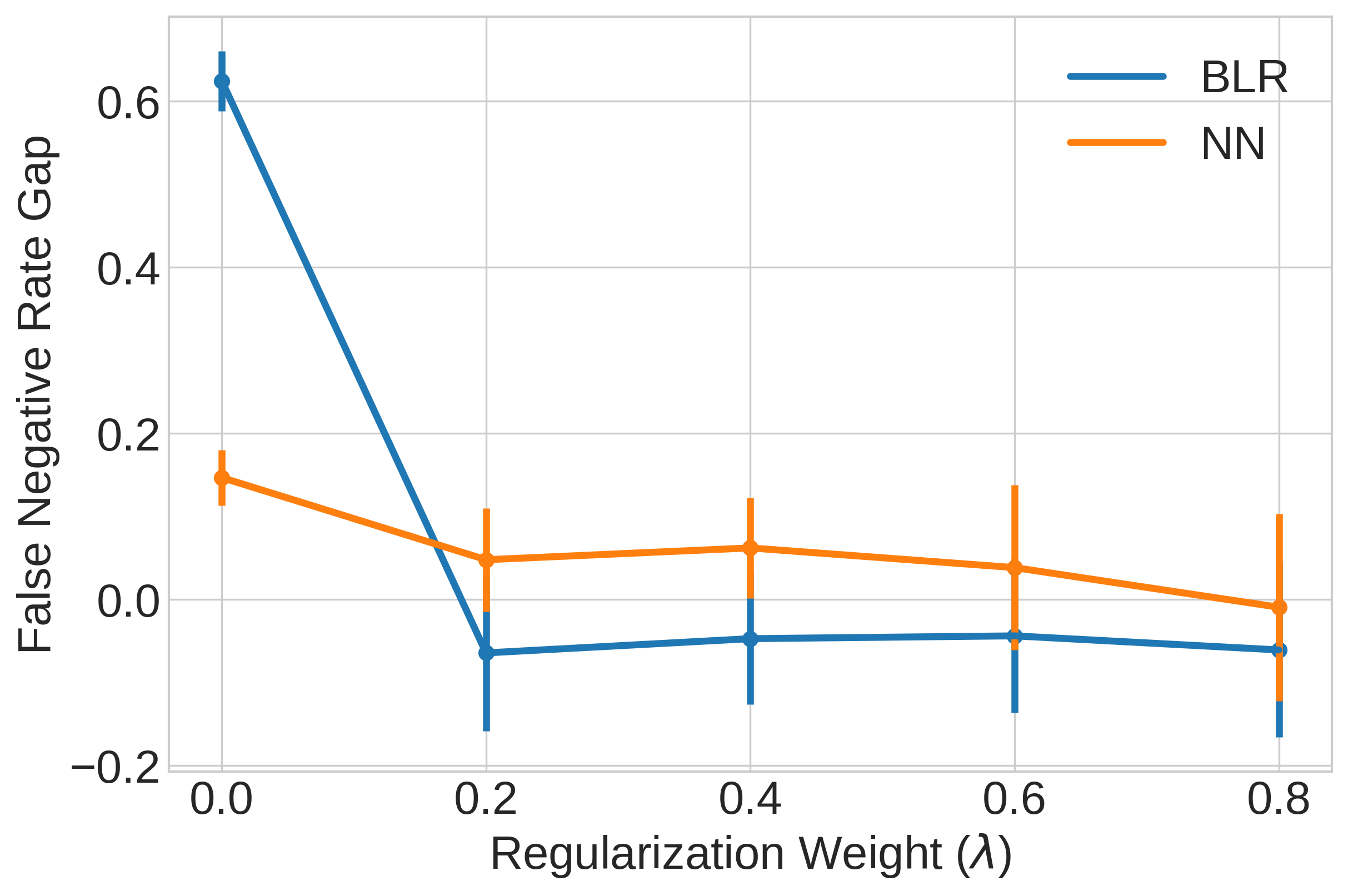}
\caption{FNR gap vs. Regularization Weight}\label{NHTS_mitigate_mode_fairness}
\end{subfigure}\hfill
\begin{subfigure}[b]{.4\linewidth}
\centering
\includegraphics[width=\linewidth]{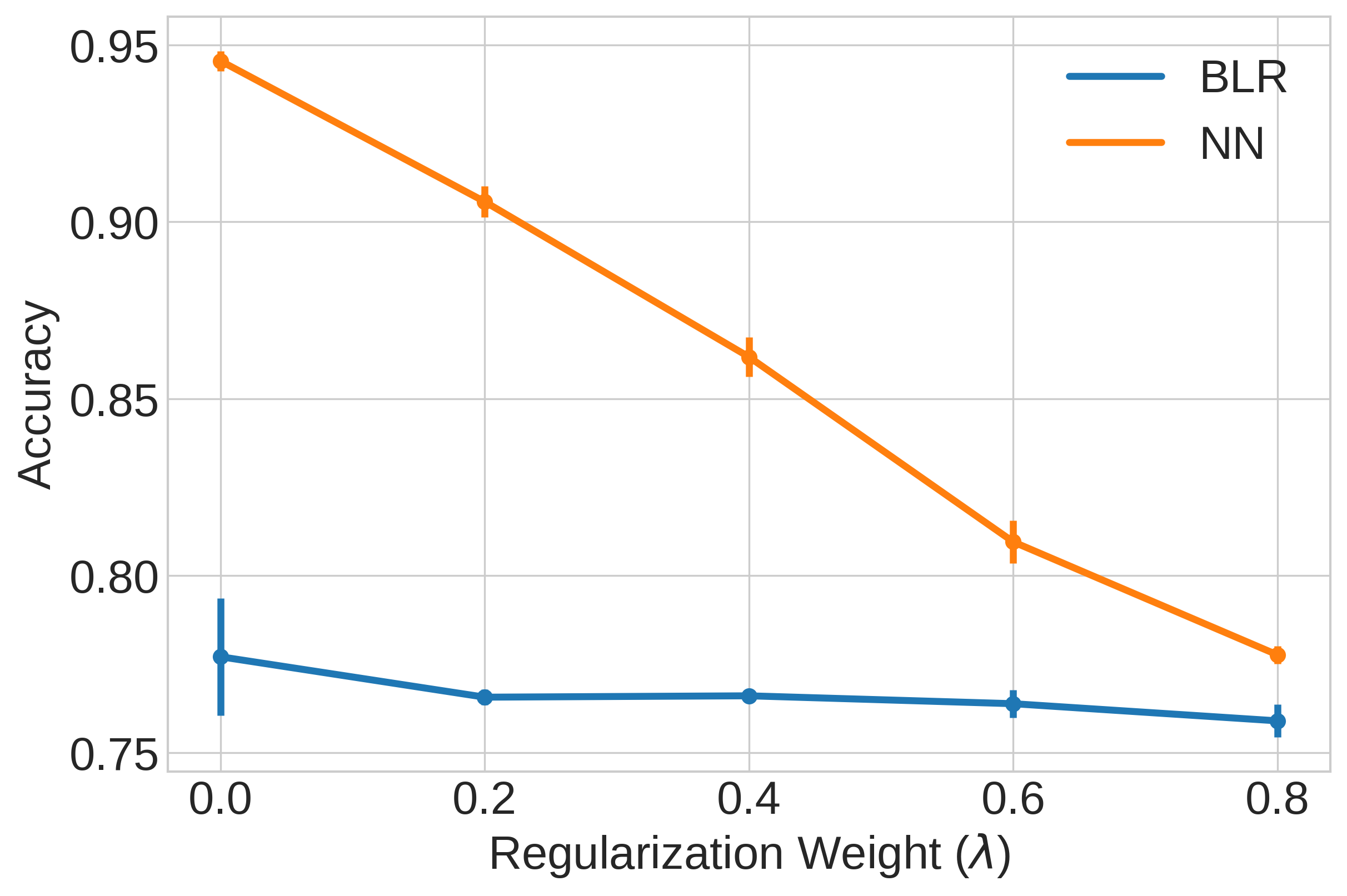}
\caption{Accuracy vs. Regularization Weight}\label{NHTS_mitigate_mode_accuracy}
\end{subfigure}\hfill
\begin{subfigure}[b]{.32\linewidth}
\centering
\end{subfigure}\hfill

\caption{Fairness and accuracy by bias mitigation weight ($\lambda$): regional bias in the prediction of frequent rideshare usage}
\label{NHTS_mitigate_mode}
\end{figure}

\begin{figure}
\begin{subfigure}[b]{.32\linewidth}
\centering
\end{subfigure}\hfill
\begin{subfigure}[b]{.4\linewidth}
\centering
\includegraphics[width=\linewidth]{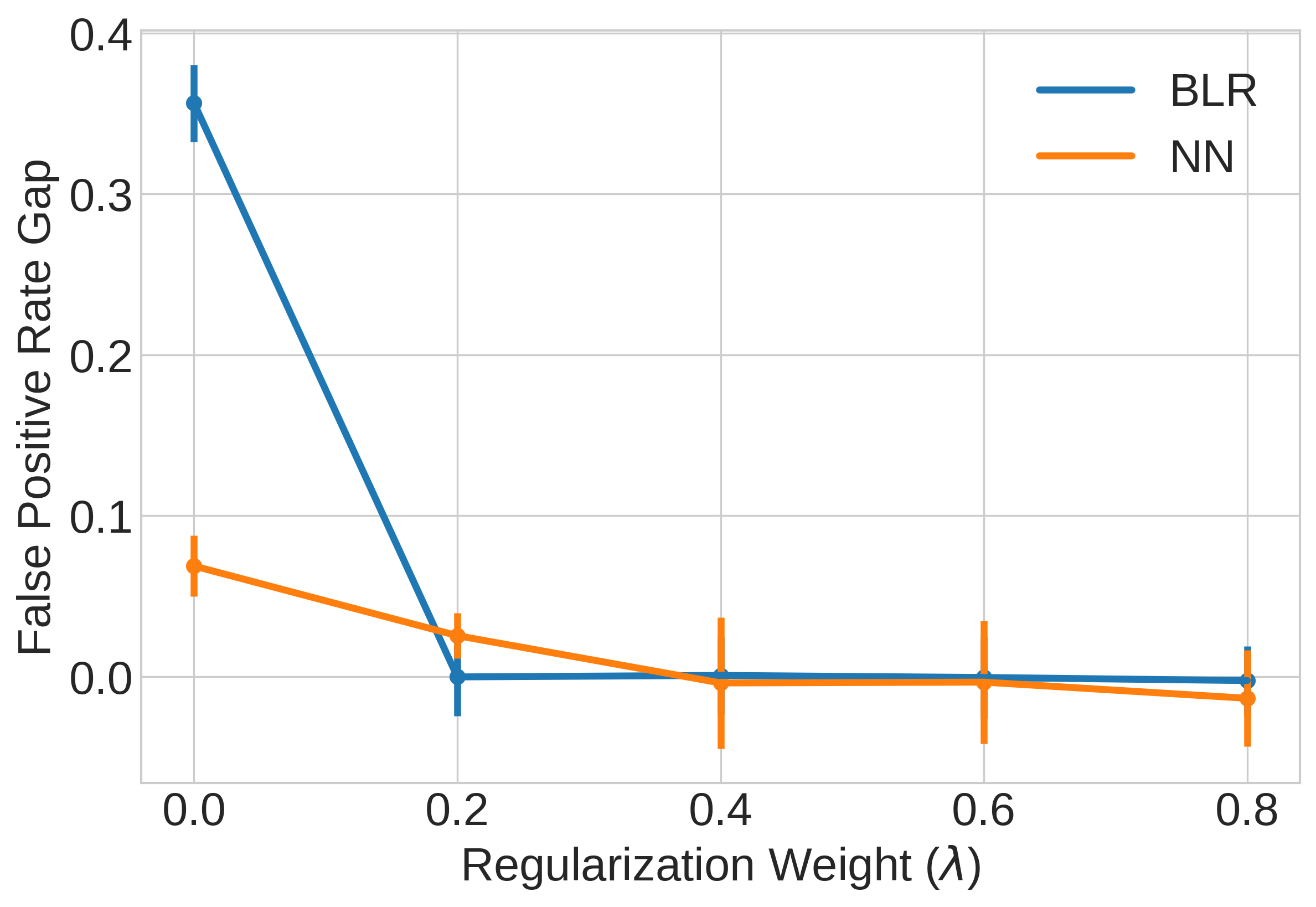}
\caption{FPR gap vs. Regularization Weight}\label{NHTS_mitigate_WFH_fairness}
\end{subfigure}\hfill
\begin{subfigure}[b]{.4\linewidth}
\centering
\includegraphics[width=\linewidth]{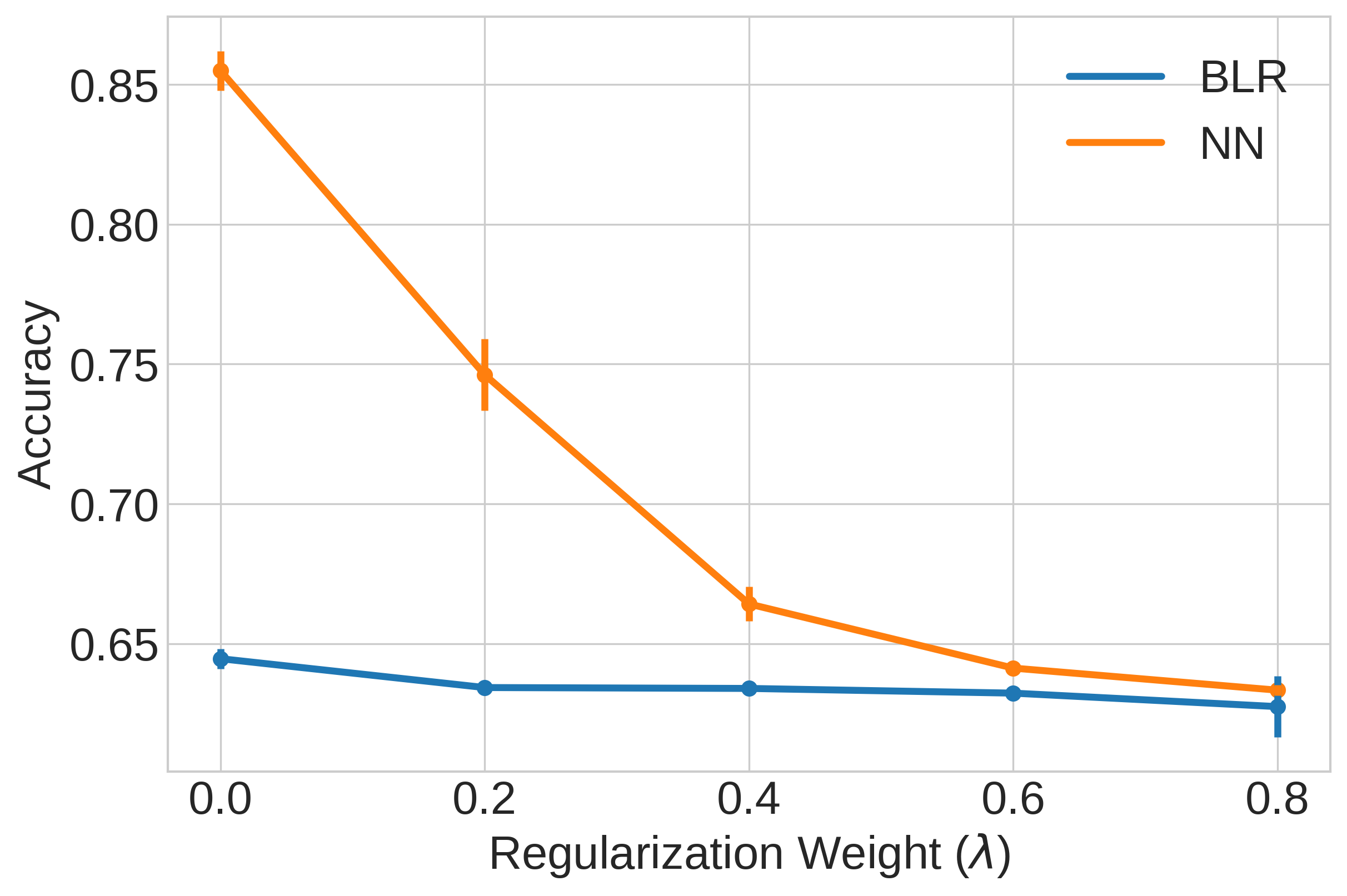}
\caption{Accuracy vs. Regularization Weight}\label{NHTS_mitigate_WFH_accuracy}
\end{subfigure}\hfill
\begin{subfigure}[b]{.32\linewidth}
\centering
\end{subfigure}\hfill

\caption{Fairness and accuracy by bias mitigation weight ($\lambda$): racial bias in the prediction of travel burden}
\label{NHTS_mitigate_WFH}
\end{figure}

We adopt the absolute correlation regularization method to mitigate the unintended bias for both BLR and DNN models. The method is applied to mitigate two types of prediction disparities: the FNR gap for estimating the frequent rideshare usage between rural and urban residents, and the FPR gap for the prediction of ``travel burden'' between the ethnic minority group and the majority group, as the initial prediction disparity in these two prediction tasks is substantial compared with those in other prediction tasks. In Figure \ref{NHTS_mitigate_mode} and \ref{NHTS_mitigate_WFH}, plot (a) reports the average prediction disparity and plot (b) reports the average prediction accuracy of 5-fold cross-validation with varied bias mitigation weight ($\lambda$). In both plots, higher weight indicates larger penalty for prediction bias. \\

\noindent Both Figure \ref{NHTS_mitigate_mode} and \ref{NHTS_mitigate_WFH} demonstrate the effectiveness of our bias mitigation method. The prediction disparity decreases as we increase $\lambda$, and this effect is particularly substantial for BLR. The blue curves in these two figures show that the fairness gap diminishes sharply even if only a small degree of bias mitigation weight is applied. \\

\noindent In Figure \ref{NHTS_mitigate_mode}, we find that increasing $\lambda$ from 0 to 0.2 can reduce the FPR gap from 62.4\% to -6.4\% for BLR and from 14.5\% to 4.8\% for DNN, only at the expense of reducing the overall accuracy from 77.7\% to 76.6\% for BLR and from 94.5\% to 90.6\% for DNN, and a similar effect is found in Figure \ref{NHTS_mitigate_WFH}. These results indicate that with the adoption of our bias mitigation method, we can reduce the fairness gap in BLR to a similar level of that in DNN. In Appendix~\ref{sec:training}, we also show the convergence of the loss functions in training. \\

\noindent Appendix~\ref{sec:other_metric} reports the change of the F1-score gap and the other fairness gap (the FPR gap for the frequent rideshare usage prediction and the FNR gap for the travel burden prediction) after these bias mitigations. The results indicate that the absolute values of both the FNR gap and the FPR gap decrease in these two prediction tasks after bias mitigation. The absolute F1-score gap also drops when BLR is applied. When DNN is applied, the absolute F1-score gap drops only for the travel burden prediction, but not for the frequent rideshare usage prediction.  \\

\noindent We also report the change of prediction disparity for other protected variables in Appendix~\ref{other_var}. The results show that mitigating the bias for one variable has the possibility of increasing prediction disparity for other variables. Therefore, future research is needed to explore ways to mitigate bias for multiple protected variables. One natural extension is to add the correlation regularization terms for all protected variables of interest to the objective function during training, thus protecting fairness for various protected variables.

\subsection{The Chicago Daily Travel Survey}
\subsubsection{Data and method}
To test the validity of our research findings, we further apply the fairness evaluation and mitigation methods to the Chicago travel survey dataset. The 2018-19 My Daily Travel Survey in Chicago\footnote{The data can be found on the Chicago Metropolitan Agency for Planning Data Hub: https://datahub.cmap.illinois.gov/dataset/mydailytravel-2018-2019-public} was produced by the Chicago Metropolitan Agency for Planning (CMAP), and the data was collected between August 2018 and April 2019. The dataset captures and represents key information regarding demographic, land-use, and travel behavior in the Chicago region. \\

\noindent We use this data to predict people's commuting mode. Among the 13,730 total observations, the weighted percentage of using auto as the travel mode to work is 73.1\%.

Therefore, we examine whether an individual uses an auto or non-auto travel mode to work as the dependent variable. The protected variables examined are race, income and disability. In terms of ``race'', we define the ethnic minority group as the non-white population. In terms of the variable ``income'', we identify low-income households based on the combination of household size and last year's household income following the 2019 U.S. Federal Poverty Guidelines \cite{usoffice}. For ``disability'', we compare the modeling results between disabled individuals and nondisabled individuals. Descriptive statistics of the dependent variable, the explanatory variables and the protected variables are presented in Appendix~\ref{sec:descriptive_chicago}. \\

\subsubsection{Experiments and results}
We adopt the same models that have been applied to the NHTS data to measure fairness and mitigate bias for work travel mode prediction using the Chicago travel survey data. BLR is implemented through the scikit-learn library. DNN with 3 hidden layers is implemented in TensorFlow using the mini-batch gradient descent method with step size 0.001 during training. The best model among the 1000 epochs is chosen and later performs prediction over the test data. The 5-fold validation is conducted for each experiment.\\

\noindent Table \ref{chicago_existence} shows that the disadvantaged groups (ethnic minority, low-income and disable populations) all have higher FNR than the privileged groups when predicting whether the work travel mode is auto, as the FNR gaps are all positive with both BLR and DNN. This is consistent with the findings in NHTS which show that these disadvantaged groups all have higher FNR when predicting frequent car usage.\\

\noindent Then, we apply the bias mitigation for ``race'', ``income'' and ``disability'' separately, and report the bias mitigation results for each of the protected variables in Table \ref{chicago_mitigation}. The mitigation weights of 0.6, 0.8 and 0.9 are applied to reduce bias for each of the three protected variables. The results indicate that in general, increasing the bias mitigation weight can reduce the absolute value of the FNR gap with a drop in accuracy, showing that the bias mitigation method is effective in reducing the prediction disparity.

\begin{table}[!htbp]
\centering
\begin{threeparttable}
\begin{tabular}{c|c|ccc}
\toprule
Model &       Accuracy & \multicolumn{3}{c}{FNR gap} \\
\cline{3-5}
{} &     {} &       Race &     Income &     Disability \\
\midrule
BLR &  0.813 (0.012) &  0.028 (0.009) &   0.145 (0.05) &  0.085 (0.079) \\
DNN &  0.917 (0.001) &  0.087 (0.015) &  0.135 (0.116) &  0.038 (0.036) \\
\bottomrule
\end{tabular}
\begin{tablenotes}
      \small
      \item Note: This table reports the means and standard deviations (in parentheses) of the accuracy and FNR gap of 5-fold cross-validation for BLR and DNN (with 3 hidden layers). 
    \end{tablenotes}
\caption{Fairness and accuracy results for the work travel mode prediction on the test set of the Chicago travel survey using BLR and DNN}
\label{chicago_existence}
\end{threeparttable}
\end{table}

\begin{table}[!htbp]
\centering
\resizebox{\textwidth}{!}{%
\begin{threeparttable}
\begin{tabular}{c|cc|cc|cc}
\toprule
$\lambda$ & \multicolumn{2}{c|}{Race} & \multicolumn{2}{c|}{Income} & \multicolumn{2}{c}{Disability} \\
\cline{2-7}
{} &        FNR gap &       Accuracy &        FNR gap &       Accuracy &         FNR gap &       Accuracy \\
\midrule
0.0 &  0.087 (0.015) &  0.917 (0.001) &  0.135 (0.116) &  0.917 (0.001) &   0.038 (0.036) &  0.917 (0.001) \\
0.6 &  0.033 (0.029) &  0.867 (0.005) &  0.091 (0.078) &  0.867 (0.004) &    0.046 (0.09) &  0.889 (0.004) \\
0.8 &  0.014 (0.025) &    0.838 (0.0) &  0.022 (0.032) &  0.839 (0.003) &  -0.023 (0.063) &  0.856 (0.004) \\
0.9 &  0.001 (0.015) &  0.827 (0.002) &  0.032 (0.026) &  0.824 (0.001) &   0.019 (0.104) &  0.836 (0.003) \\
\bottomrule
\end{tabular}
\begin{tablenotes}
      \small
      \item Note: the prediction model is DNN with 3 hidden layers. $\lambda$ represents the bias mitigation weight. The means and standard deviations (in parentheses) are calculated based on the 5-fold cross-validation.
    \end{tablenotes}
\caption{The results of bias mitigation for different protected variables (the Chicago travel survey)}
\label{chicago_mitigation}
\end{threeparttable}%
}
\end{table}

\section{Conclusions}
This study investigates equality of opportunity as a measurement of computational fairness in travel behavior modeling with BLR and DNN. Fairness has long been a critical concern in transportation studies. However, past transportation equity studies generally evaluated equity based on the cost and benefit analysis for different populations and neglected the potential fairness problem in machine-learned models. This research aims to enrich the transportation equity research by emphasizing the integration of fairness metrics into the modeling process to improve fair transportation decision making and resource allocation.\\


\noindent First, we use the concept of equality of opportunity to define computational fairness in travel behavior modeling, which is measured by the gap in true positive rates between two groups of populations. This definition is connected to the equality of opportunity in traditional transportation equity literature, since choice probabilities in discrete choice models are a natural metric for opportunities. The unfair prediction of choice probabilities in modeling can lead to the allocation of inadequate transportation resources to disadvantaged neighborhoods.\\


\noindent Then, we conduct simulated experiments to show two sources of prediction disparities: the bias inherent in the data structure and the bias in the modeling process. Prediction disparity increases as the correlation between the protected variables and the explanatory variable(s) increases, exhibiting the bias inherent in the data structure but not caused by modelers. The inherent bias illustrates how prediction disparities may still exist without the presence of human errors. When the true model specification cannot be captured by BLR but can be captured by DNN, DNN can produce lower prediction bias than BLR, owing to its ability to capture the complex relationships among variables. This type of bias is the algorithmic bias that can be mitigated by choosing a more fitting model, thus is a sort of human bias. \\


\noindent Next, we conduct computational fairness analysis on the NHTS and the Chicago travel survey datasets using BLR and DNN. The results reveal the prevalence of prediction disparities in travel behavior modeling particularly when BLR is adopted, and these prediction disparities are consistent with the change of the hyperparameters including the number of DNN layers, the batch size and the weight initialization. Though the magnitudes of unfairness are different, the signs of the fairness gap (e.g. which group has higher FNR/FPR) are the same between BLR and DNN in the vast majority of our prediction cases, probably reflecting the bias inherent in the existing population disparity. For the NHTS analysis, we find that the regional disparity in predictions of frequent usage of bike, bus and rideshare is the largest among all the protected attributes, indicating that bike, bus and rideshare services allocated by the predictions will be withheld from rural residents compared with urban residents. When estimating the high frequent usage of rideshare, the disadvantaged group has higher FNR than the other group for all protected variables, indicating that these disadvantaged social groups might receive insufficient ride sharing service if the TNC companies use these data for rideshare demand estimation but do not consider prediction disparity. For the Chicago travel survey analysis, the results show that the disadvantaged groups have higher FNR when predicting whether the travel mode is auto, which is consistent with the findings in the NHTS analysis.\\

\noindent Finally, we adopt the absolute correlation regularization method to mitigate the bias in the BLR and DNN prediction using the synthetic, NHTS and the Chicago travel survey datasets. The method proves to be effective for both BLR and DNN. When the initial fairness gap is large, applying our bias mitigation method with only a small weight of regularization can considerably improve the fairness result. Though there is an accuracy-fairness trade-off in most predictions, we can achieve substantial improvement in prediction fairness with only a small reduction in accuracy by careful selection of the bias mitigation weight. The synthetic experiments show that the bias mitigation works whether the bias is caused by inherent or modeling bias, and whether the dataset is imbalanced or not. The NHTS experiments also show that through mitigating the bias for one fairness metric (e.g. the FNR gap), generally other fairness metrics (e.g. the FPR gap and the F1-score gap) can also be improved. However, mitigating bias for one protected variable may increase the bias for other protected variables, which necessitates future research into solutions to improving fairness for multiple protected variables simultaneously.\\

\noindent All in all, we argue that researchers and policy makers should be aware of the normative aspect in seemingly value-neural machine learning predictions, since the prediction disparities can lead to severely unfair treatment of already marginalized social groups. Only after acknowledging the existence of the biases can policy makers start to adopt effective remedies. 


\section{Discussions and Future Work}
\noindent This study reveals the need to incorporate the equality of opportunity metric into the travel demand modeling framework, which is especially important when forecasting mode split under various transportation service deployment scenarios. It is also important when predicting the market share of new travel modes, such as automated vehicles, electric mobility and mobility-as-a-service packages. Due to the discriminatory behaviors exhibited by the prediction algorithms, socially disadvantaged groups may be under-served by the providers of these services. In the context of personalized on-demand mobility service, the preferences of the passengers towards various travel mode alternatives are captured by the choice modeling results, and the operators incorporate these predicted choice probabilities into their optimization framework for vehicle fleet allocation in order to maximize their profits \cite{atasoy2015concept,danaf2019context}. Given the potential existence of computational bias, the passengers from the disadvantaged group may not get access to their desired travel mode, which consequently exacerbates the social inequity problem.\\

\noindent Inherent and modeling biases are only two sources of prediction disparities. However, future studies should investigate other factors that also lead to prediction disparity and pinpoint more granular sources of bias in travel predictions. Previous literature has shown that prediction disparities can stem from various sources such as sampling procedure, feature quality and algorithms, but no study has examined these factors in the context of travel behavior modeling \cite{barocas2016big,suresh2019framework,zliobaite2017fairness}. While we use equality of opportunity as the fairness metric in this research, scholars can also consider other metrics to measure computational fairness. Another future research direction is to study fairness issues with machine learning models other than DNN and BLR. In addition, while we try to address computational fairness through an algorithmic approach, future studies can explore how to combine our computational perspective with other remedies such as taking a participatory effort to evaluate fairness in the modeling results for various transportation programs. \\ 

\noindent This study demonstrates that adjusting BLR and DNN models to reduce prediction disparity may lead to lower prediction accuracy, but how to trade off accuracy and fairness essentially involves a value judgement, on which the authors do not take a stance. In fact, this needs to be analyzed on the basis of different normative principles, and may involve some ethical debates about what should be considered ``fair'' in a given context. In addition, mitigating unintended prediction biases can involve more costs. Policy makers should determine whether the gain in computational fairness can outweigh the efforts of improving computational fairness, such as employing bias mitigation algorithms or collecting data that represent socially marginalized groups. Further cost and benefit analysis is necessary to bridge the gap between algorithmic fairness tools and social justice needs in the real world.

\section{Acknowledgements}
We thank Singapore-MIT Alliance for Research and Technology (SMART) for partially funding this research. 

\section{Auther Contribution Statement}
The authors confirm contribution to the paper as follows: study conception and design: Y.Z., S.W, J.Z.; data collection: Y.Z., S.W.; analysis and interpretation of results: Y.Z.; draft manuscript preparation: Y.Z., S.W.; research supervision: J.Z. All authors reviewed the results and approved the final version of the manuscript.

\section{Conflict of Interests}
The authors do not have any conflicts of interest to declare.

\bibliographystyle{vancouver}
\bibliography{bibliography}

\newpage
\section*{Appendix}
\appendix
\label{appendix}
\section{The Synthetic Data Generation Process}
\label{sec:DGP}
\noindent The label $y$ is generated based on the following equations:
\begin{linenomath}
\begin{gather}
U_{i}=V_{i}+\varepsilon_{i}=f(x_{i},\boldsymbol{k_{i}})+\varepsilon_{i} \label{eqn_one}\\
Pr(y_{i}=1)=\frac{1}{1+\exp{(-V_{i})}}\label{eqn_2}
\end{gather}
\end{linenomath}
Equation \ref{eqn_one} denotes the true utility function specification, where $\varepsilon$ is the extreme value distributed random term. Equation \ref{eqn_2} is the Sigmoid function calculating the probability of the binary outcomes. In simulations, for each individual $i$ we draw $y_{i}$ from a binomial distribution which takes value 1 with $Pr(y_{i}=1)$. Note that this is where sampling errors may occur.
\begin{linenomath}
\begin{gather}
    (a_{i},x_{i}) \sim N\left(0,\begin{pmatrix}
    1 & Cov_{ax} \label{eqn_3}\\
Cov_{ax} & 1
    \end{pmatrix}\right)\\
    z_{i}=
    \begin{cases}
      1, & \text{if } a_{i} \geq0 \\
      0, & \text{if } a_{i}<0
    \end{cases}\label{eqn_4}\\
    \boldsymbol{k_{i}} \sim N(0,I) \label{eqn_5}
\end{gather}
\end{linenomath}
\noindent $z$, $x$ and $\boldsymbol{k}$ are derived based on Equation \ref{eqn_3}, \ref{eqn_4} and \ref{eqn_5}. $a$ is an intermediate variable for the creation of $z$. $a$, $x$ and all of the elements in $\boldsymbol{k}$ are drawn from a multivariate Gaussian distribution with zero-mean and unit-variance. The above variables drawn from this distribution are all independent, except that $a$, $x$ are correlated with the covariance being $Cov_{ax}$. $z$ takes value 1 if $a \geq 0$ and takes value 0 otherwise. Therefore, the mean value of $z$ is 0.5, indicating that the number of positive values and negative values of $z$ is approximately the same. In our simulations, the value of $Cov_{ax}$ varies across 0, 0.25, 0.5, 0.75, 1. The corresponding values of $Cov(z, x)$ are 0, 0.1, 0.2, 0.3, 0.4, each of which is empirically calculated as the value of $Cov(z, x)$ averaged across the 3 independently generated datasets used in the simulations for each $Cov_{ax}$.\\

\noindent $f(x_{i},\boldsymbol{k_{i}})$ in Equation \ref{eqn_one} indicates the true systematic utility function. We consider two scenarios: 
\begin{linenomath}
\begin{gather}
f(x_{i},\boldsymbol{k_{i}})=\alpha+x_{i}\beta_{x_{1}}+\boldsymbol{k_{i}}^\top \boldsymbol{\beta_{k_{1}}} \label{f1}\\
f(x_{i},\boldsymbol{k_{i}})=\alpha+x_{i}\beta_{x_{1}}+x_{i}^2\beta_{x_{2}}+\boldsymbol{k_{i}}^\top \boldsymbol{\beta_{k_{1}}}+(\boldsymbol{k_{i}}\odot \boldsymbol{k_{i}})^\top \boldsymbol{\beta_{k_{2}}} \label{f2}
\end{gather}
\end{linenomath}
In Scenario 1, the utility specification takes a linear form (Equation \ref{f1}). We set $\alpha$ to 0 and $\beta_{x_{1}}$ to 1. Each entry of $\boldsymbol{\beta_{k_{1}}}$ takes \{-0.5,0.5\} values with equal probabilities. As such, $x$ will have the strongest positive influence on $V$ compared with other explanatory variables.\\

\noindent In Scenario 2, the utility specification takes a quadratic form (Equation \ref{f2}). We set $\alpha$ to -0.5, $\beta_{x_{1}}$ to 1 and $\beta_{x_{2}}$ to 0.5. Each entry of $\boldsymbol{\beta_{k_{1}}}$ and $\boldsymbol{\beta_{k_{2}}}$ takes \{-0.5,0.5\} values with equal probabilities.\\

\noindent This setup for both scenarios makes sure that $E(V)$ is equal to 0, so the average numbers of outcomes $y$ taking the value 1 and 0 are approximately the same.

\section{Descriptive Statistics (NHTS Survey)}
\label{sec:descriptive_nhts}
\begin{enumerate}
   \item \textit{Dependent variables: travel behavior and attitude indicators}
\begin{itemize}
    \item Work from home: "Do you usually work from home?" (Yes=1)
    \item Work from home option: "Do you have the option of working from home?" (Yes=1)
    \item Travel burden: "Do you agree that travel is a financial burden?" (``Strongly agree'' or ``Agree''= 1)
    \item Gas price impact: "Do your agree that gas price affects travel?" (``Strongly agree'' or ``Agree''= 1)
\end{itemize}
\item \textit{Dependent variables: variables regarding travel mode usage}
\begin{itemize}
    \item Frequent usage of bike: "Do you use bike for travel daily or a few times a week?" (Yes=1)
    \item Frequent usage of car: "Do you use car for travel daily or a few times a week?" (Yes=1)
    \item Frequent usage of bus: "Do you use bus for travel daily or a few times a week?" (Yes=1)
    \item Frequent usage of rideshare: "Do you use rideshare at least once in the past 30 days?" (Yes=1)
\end{itemize}
\end{enumerate}

\setcounter{table}{0}
\renewcommand{\thetable}{B\arabic{table}}
\begin{table}[hbt!]
    \centering
    \begin{threeparttable}
\begin{tabular}{llllll}
\toprule
{} Variable&    Mean &    Std. &  Min &  Median &    Max \\
\midrule
\textbf{\textit{Socio-demographics:}}                                              &   &  &  &   &    \\
Age                                                &  45.834 &  17.372 &  6.0 &    45.0 &   92.0 \\
*Gender (Male=1)                                    &   0.479 &   0.500 &  0.0 &     0.0 &    1.0 \\
*Ethnic minority (Yes=1)                            &   0.240 &   0.427 &  0.0 &     0.0 &    1.0 \\
*Low income household (Yes=1)                       &   0.130 &   0.337 &  0.0 &     0.0 &    1.0 \\
*Medical condition (Yes=1)                          &   0.065 &   0.246 &  0.0 &     0.0 &    1.0 \\
Driver status (Yes=1)                                     &   0.915 &   0.280 &  0.0 &     1.0 &    1.0 \\
Education: Less than high school (Yes=1)           &   0.068 &   0.252 &  0.0 &     0.0 &    1.0 \\
Education: High school graduate (Yes=1)            &   0.190 &   0.392 &  0.0 &     0.0 &    1.0 \\
Education: College degree (Yes=1)                  &   0.293 &   0.455 &  0.0 &     0.0 &    1.0 \\
Education: Bachelor's degree (Yes=1)               &   0.245 &   0.430 &  0.0 &     0.0 &    1.0 \\
Home ownership (Yes=1)                             &   0.661 &   0.473 &  0.0 &     1.0 &    1.0 \\
Primary activity: absent (Yes=1)                   &   0.027 &   0.163 &  0.0 &     0.0 &    1.0 \\
Primary activity: homemaker (Yes=1)                &   0.073 &   0.260 &  0.0 &     0.0 &    1.0 \\
Primary activity: unemployed (Yes=1)               &   0.035 &   0.184 &  0.0 &     0.0 &    1.0 \\
Primary activity: retired (Yes=1)                  &   0.158 &   0.365 &  0.0 &     0.0 &    1.0 \\
Primary activity: going to school (Yes=1)          &   0.067 &   0.250 &  0.0 &     0.0 &    1.0 \\
Born in the U.S. (Yes=1)                           &   0.861 &   0.346 &  0.0 &     1.0 &    1.0 \\
More than one job (Yes=1)                          &   0.075 &   0.263 &  0.0 &     0.0 &    1.0 \\
Health level (Excellent=1, Poor=5)                 &   2.142 &   0.960 &  1.0 &     2.0 &    5.0 \\
Job: Sales or service (Yes=1)                      &   0.168 &   0.373 &  0.0 &     0.0 &    1.0 \\
Job: Clerical or administrative support (Yes=1)    &   0.070 &   0.256 &  0.0 &     0.0 &    1.0 \\
Job: Manufacturing type (Yes=1)                    &   0.087 &   0.282 &  0.0 &     0.0 &    1.0 \\
Work for pay (Yes=1)                               &   0.069 &   0.253 &  0.0 &     0.0 &    1.0 \\
Level of physical activity &   \multirow{2}{*}{2.173} &    \multirow{2}{*}{0.594} &   \multirow{2}{*}{1.0} &      \multirow{2}{*}{2.0} &     \multirow{2}{*}{3.0} \\
(Never/rarely=1, Vigorous=3) &    &  &  &  &    \\
In public or private school (Yes=1)                &   0.024 &   0.154 &  0.0 &     0.0 &    1.0 \\
Full-time worker (Yes=1)                           &   0.504 &   0.500 &  0.0 &     1.0 &    1.0 \\
\textbf{\textit{Household Variables:}}                                              &   &  &  &   &    \\
Number of drivers in the HH                        &   2.038 &   0.945 &  0.0 &     2.0 &    9.0 \\

\bottomrule
\end{tabular}
    \begin{tablenotes}
      \small
      \item Note: the sample weights are used to compute the summary statistics; (*) denotes the independent variables that are also treated as the protected variables based on which we examine the prediction disparities.
    \end{tablenotes}

  \end{threeparttable}
    \caption{Summary statistics of the explanatory and dependent variables}
    \label{tab:summary1}
\end{table}

\renewcommand{\thetable}{B\arabic{table}}
\begin{table}[hbt!]
    \centering
    \begin{threeparttable}
\begin{tabular}{llllll}
\toprule
{} Variable&    Mean &    Std. &  Min &  Median &    Max \\
\midrule
Last year's household income (K \$)                      &  84.009 &  64.924 &  5.0 &    62.5 &  250.0 \\
Number of household members                        &   2.915 &   1.435 &  1.0 &     3.0 &   13.0 \\
Number of household vehicles                       &   2.187 &   1.305 &  0.0 &     2.0 &   12.0 \\
No child in the HH (Yes=1)                         &   0.352 &   0.478 &  0.0 &     0.0 &    1.0 \\
Youngest child's age < 15 in the HH (Yes=1)        &   0.343 &   0.475 &  0.0 &     0.0 &    1.0 \\
Youngest child's age < 21 in the HH (Yes=1)        &   0.102 &   0.303 &  0.0 &     0.0 &    1.0 \\
Number of workers in household                     &   1.495 &   1.004 &  0.0 &     1.0 &    7.0 \\
\textbf{\textit{Built Environment:}}                                              &   &  &  &   &    \\
*Household in urban area (Yes=1)                    &   0.836 &   0.371 &  0.0 &     1.0 &    1.0 \\
Northeast Region (Yes=1)                           &   0.178 &   0.382 &  0.0 &     0.0 &    1.0 \\
Midwest Region (Yes=1)                             &   0.217 &   0.412 &  0.0 &     0.0 &    1.0 \\
Gas Price (\$)                                      &   2.395 &   0.208 &  2.0 &     2.4 &    3.0 \\
\% of renter-occupied housing in the block group    &  31.226 &  22.877 &  2.0 &    20.0 &   97.5 \\
Housing units/sq.m in the block group  &   \multirow{2}{*}{3.004} &  \multirow{2}{*}{5.409} &  \multirow{2}{*}{0.0} &     \multirow{2}{*}{1.5} &   \multirow{2}{*}{30.0} \\
(in thousands) &    &  &   &    &   \\
Second City (Yes=1)                                &   0.201 &   0.401 &  0.0 &     0.0 &    1.0 \\
Suburban (Yes=1)                                   &   0.235 &   0.424 &  0.0 &     0.0 &    1.0 \\
Small Town (Yes=1)                                 &   0.195 &   0.396 &  0.0 &     0.0 &    1.0 \\
Number of workers/sq.m in the census tract&   \multirow{2}{*}{1.751} &   \multirow{2}{*}{1.703} &  \multirow{2}{*}{0.0} &     \multirow{2}{*}{1.5} &    \multirow{2}{*}{5.0} \\
(in thousands) &    &   &  &    &    \\
MSA population> 1 million, without rail (Yes=1)    &   0.282 &   0.450 &  0.0 &     0.0 &    1.0 \\
MSA population < 1 million (Yes=1)                 &   0.299 &   0.458 &  0.0 &     0.0 &    1.0 \\
Population size of the MSA (in millions)           &   1.959 &   1.643 &  0.0 &     2.0 &    4.0 \\
In an urban area (Yes=1)                           &   0.741 &   0.438 &  0.0 &     1.0 &    1.0 \\
In an urban cluster (Yes=1)                        &   0.095 &   0.293 &  0.0 &     0.0 &    1.0 \\
Area surrounded by urban areas (Yes=1)             &   0.000 &   0.022 &  0.0 &     0.0 &    1.0 \\
Urban area size (in millions)                      &   1.018 &   0.910 &  0.0 &     0.8 &    2.0 \\

\textbf{\textit{Travel Pattern and Internet Usage:}}                                              &   &  &  &   &    \\
Flexible work time (Yes=1)                         &   0.307 &   0.461 &  0.0 &     0.0 &    1.0 \\
Travel day began at home location (Yes=1)          &   0.938 &   0.241 &  0.0 &     1.0 &    1.0 \\
Frequent internet use (Yes=1)                      &   0.955 &   0.208 &  0.0 &     1.0 &    1.0 \\
\midrule
\textbf{\textit{Dependent Variables:}}                             &   &  &  &   &    \\
Work from home option (Yes=1)                      &   0.184 &   0.388 &  0.0 &     0.0 &    1.0 \\
Work from home (Yes=1)                             &   0.116 &   0.321 &  0.0 &     0.0 &    1.0 \\
Travel is a financial burden (Yes=1)               &   0.396 &   0.489 &  0.0 &     0.0 &    1.0 \\
Gas price affects travel (Yes=1)                   &   0.478 &   0.500 &  0.0 &     0.0 &    1.0 \\
Frequent usage of bike (Yes=1)                     &   0.066 &   0.248 &  0.0 &     0.0 &    1.0 \\
Frequent usage of bus (Yes=1)                      &   0.063 &   0.243 &  0.0 &     0.0 &    1.0 \\
Frequent usage of car (Yes=1)                      &   0.926 &   0.262 &  0.0 &     1.0 &    1.0 \\
Frequent usage of rideshare (Yes=1)                &   0.114 &   0.317 &  0.0 &     0.0 &    1.0 \\
\bottomrule
\end{tabular}
\begin{tablenotes}
      \small
      \item Note: the sample weights are used to compute the summary statistics; (*) denotes the independent variables that are also treated 
as the protected variables based on which we examine the prediction disparities.
    \end{tablenotes}
        \caption{(Cont.) Summary statistics of the explanatory and dependent variables}
    \label{tab:summary2}
  \end{threeparttable}
\end{table}
\FloatBarrier

\begin{table}[!htbp]
\centering
\begin{threeparttable}
\begin{tabular}{cc|cccc}
\toprule
  \multicolumn{2}{c|}{Protected Variable} & \multicolumn{4}{c}{Dependent Variable} \\ 
\hline
                Type &           Value &                WFH &    WFHO &      TB &     GPI \\
\midrule
                \multirow{2}{*}{Race} &             Minority &             29.73\% &  31.25\% &   46.6\% &  56.59\% \\
                   &             Majority &             34.02\% &  33.79\% &  33.02\% &  42.98\% \\
 \hline             \multirow{2}{*}{Gender} &                 Male &             33.52\% &  37.67\% &  34.21\% &  43.66\% \\
                   &               Female &             33.19\% &  28.92\% &  35.95\% &  46.37\% \\
 \hline             \multirow{2}{*}{Income} &                  Low &             37.62\% &  15.53\% &  56.81\% &  67.23\% \\
                   &       Middle or High &              33.1\% &  34.29\% &  33.14\% &  43.07\% \\
 \hline  \multirow{2}{*}{Medical Condition} &                 With &             49.72\% &  31.54\% &   49.2\% &  57.53\% \\
                   &              Without &             32.99\% &  33.41\% &   34.1\% &  44.19\% \\
 \hline         \multirow{2}{*}{Region} &                Urban &             32.71\% &   34.9\% &  33.63\% &  42.71\% \\
                   &                Rural &              35.9\% &  26.76\% &  40.67\% &  53.85\% \\ \hline 
                   &          Sample Size &             160110 &  204603 &  703647 &  703647 \\
                   &  Number of Positives &              53400 &   68296 &  247300 &  317432 \\
\bottomrule
\end{tabular}

\begin{tablenotes}
      \small
      \item Note: ``WFH'' stands for ``work from home'', ``WFHO'' stands for ``work from home option'', ``TB'' stands for ``travel burden'', ``GPI'' stands for ``gas price impact''; each percentage number indicates the proportion of positives to the total number of the corresponding dependent variable in the subset of the corresponding protected variable; since the outcome distributions of ``WFH'' and ``WFHO'' are highly skewed, for these two variables the data is balanced so that the ratio of the major outcome class to the minor outcome class instances is 2:1.
    \end{tablenotes}
\caption{Summary statistics of dependent variables by protected variables in the data set}
\label{descriptive_1}
\end{threeparttable}
\end{table}
\clearpage
\begin{table}[!htbp]
\centering
\begin{threeparttable}
\begin{tabular}{cc|cccc}
\toprule
          \multicolumn{2}{c|}{Protected Variable} & \multicolumn{4}{c}{Frequent Usage of} \\ 
\hline
                        Type &                 Value &               Bus &    Bike &     Car & Rideshare \\
\midrule
                        \multirow{2}{*}{Race} &              Minority &            55.03\% &  30.91\% &  48.35\% &    37.28\% \\
                             &              Majority &             27.1\% &  33.83\% &  71.28\% &    32.56\% \\
 \hline                     \multirow{2}{*}{Gender} &                  Male &            33.03\% &  35.99\% &  66.65\% &    35.09\% \\
                             &                Female &            33.67\% &  30.96\% &   66.6\% &    31.75\% \\
 \hline                       \multirow{2}{*}{Income} &            Low &            62.76\% &  39.06\% &  29.27\% &    20.76\% \\
                             &        Middle or High &            28.54\% &  32.82\% &  73.98\% &    34.34\% \\
 \hline 
           \multirow{2}{*}{Medical Condition} &      With &            46.37\% &  26.01\% &  38.92\% &    13.31\% \\
                             &  Without &            32.21\% &  33.88\% &  69.98\% &    34.53\% \\
 \hline                       \multirow{2}{*}{Region} &                 Urban &            37.81\% &   35.8\% &  63.82\% &    37.96\% \\
                             &                 Rural &            10.79\% &  23.09\% &  78.85\% &    11.43\% \\ \hline  
                 & Sample Size&             72235 &  130939 &   95323 &    181132 \\
 &     Number of Positives                  &             24106 &   43712 &   63504 &     60378 \\
\bottomrule
\end{tabular}
\begin{tablenotes}
      \small
      \item Note: since the distributions of the travel mode usage are highly skewed, for each dependent variable the data is balanced so that the ratio of the major outcome class to the minor outcome class instances is 2:1.
    \end{tablenotes}
\caption{(Cont.) Summary statistics of dependent variables by protected variables in the data set}
\label{descriptive_2}
\end{threeparttable}
\end{table}

\section{Convergence of Loss Values in the Training Process}\label{sec:training}
\setcounter{figure}{0}
\renewcommand{\thefigure}{C\arabic{figure}}
\begin{figure}
\begin{subfigure}[b]{.32\linewidth}
\centering
\end{subfigure}\hfill
\begin{subfigure}[b]{.4\linewidth}
\centering
\includegraphics[width=\linewidth]{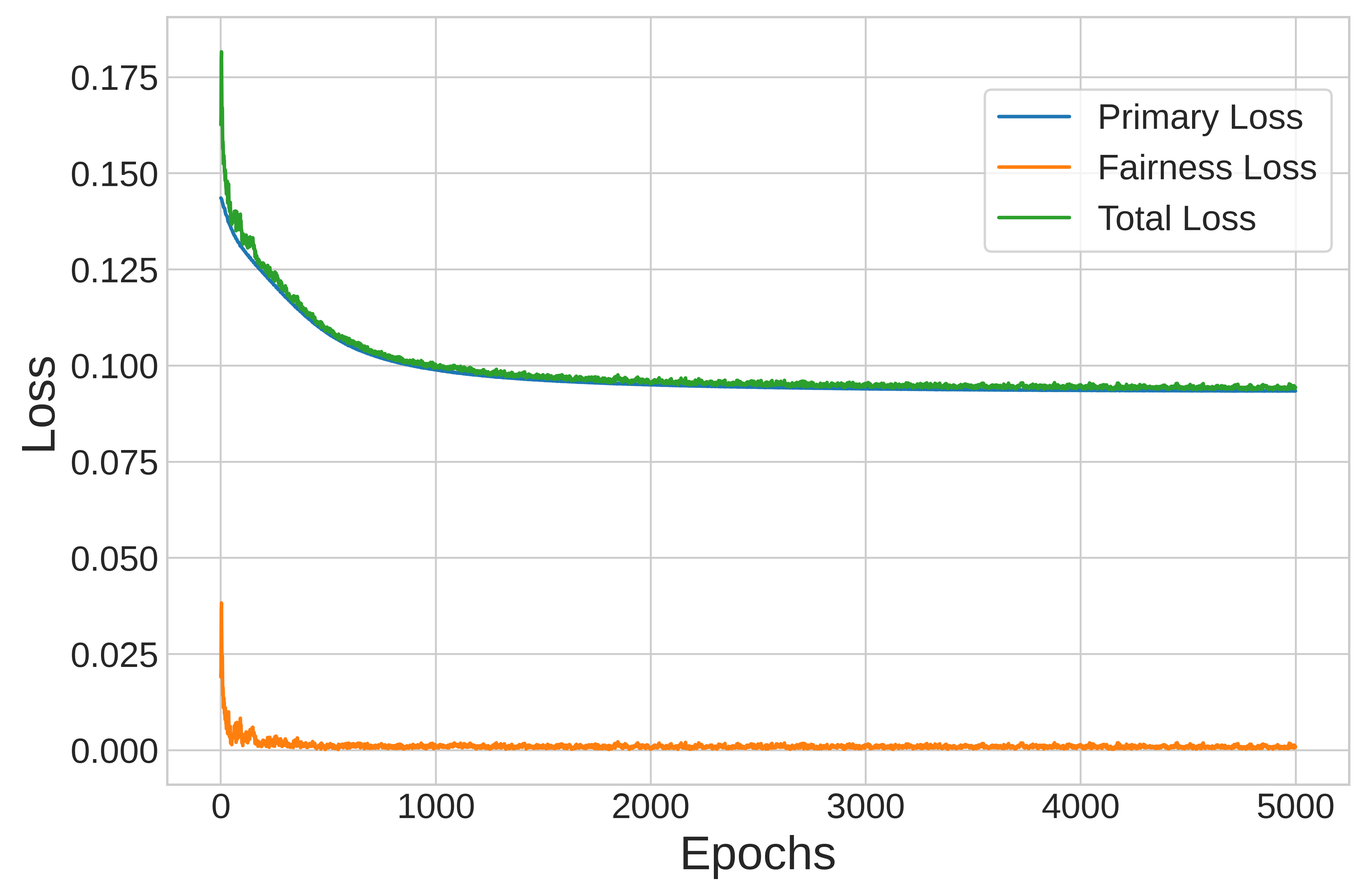}
\caption{Training data}\label{process_train}
\end{subfigure}\hfill
\begin{subfigure}[b]{.4\linewidth}
\centering
\includegraphics[width=\linewidth]{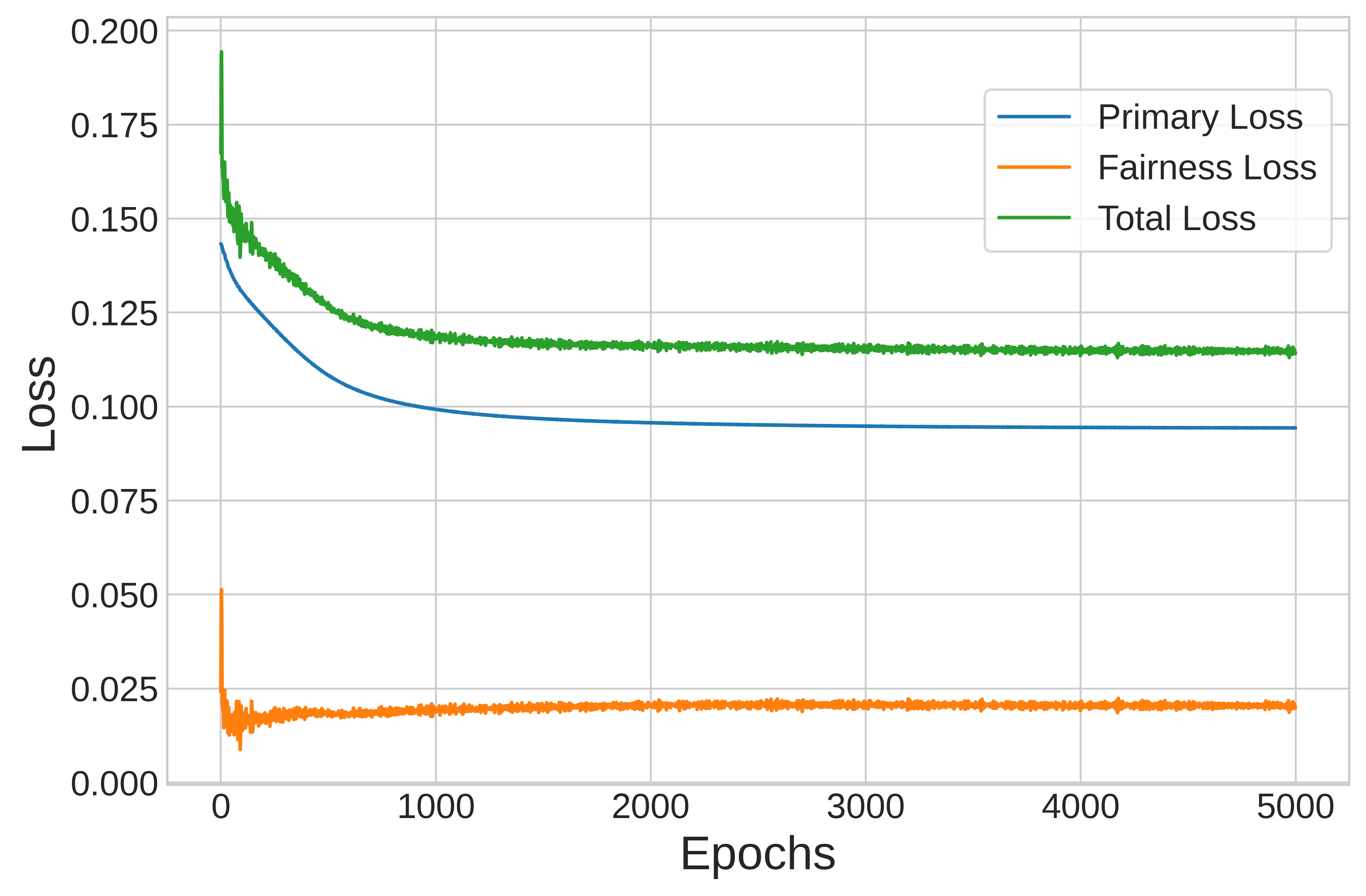}
\caption{Testing data}\label{process_test}
\end{subfigure}\hfill
\begin{subfigure}[b]{.32\linewidth}
\centering
\end{subfigure}\hfill

\caption{Change of loss values in the training process. Protected variable: urban-rural divide; dependent variable: the frequent usage of rideshare; mitigation weight ($\lambda$): 0.8.}
\label{training_process}
\end{figure}

\noindent The convergence of loss values during the training process is shown in Figure \ref{training_process}, where we try to mitigate the prediction disparity between rural and urban residents when predicting frequent usage of rideshare using DNN. The bias mitigation weight is 0.8. Primary loss refers to $(1-\lambda)$ times the cross-entropy loss which is used to increase the prediction accuracy, whereas fairness loss refers to $\lambda$ times the correlation loss which is applied to mitigate the prediction bias. The total loss is computed as the sum of the primary and the fairness loss.\\

\noindent From Figure \ref{process_train}, we can see that in the first few epochs, the reduction of total loss is mainly driven by the reduction of fairness loss as the values of both losses drop sharply. As the fairness loss drops to nearly zero, the algorithm then primarily tries to mitigate the prediction loss, while the fairness loss remains very small in the following training steps. The testing data shows similar trends of loss values during the training process (Figure \ref{process_test}), except that the fairness loss never drops to the near-zero level. All in all, Figure \ref{training_process} illustrates how our bias mitigation method works, and how the algorithm manages to substantially reduce fairness loss at early stage of training.
\newpage
\newgeometry{hmargin=2cm,vmargin=3cm}
\begin{landscape}
\section{Sensitivity Analysis}\label{sec:sensitivity}
\setcounter{table}{0}
\renewcommand{\thetable}{D\arabic{table}}
\begin{table}[hbt!]
    \centering
    \begin{threeparttable}
\begin{tabular}{c|cccc|cccc}
\toprule
Model & \multicolumn{4}{c|}{FNR gap} & \multicolumn{4}{c}{FPR gap} \\
\cline{2-9}
{} &            Bike &             Bus &             Car &       Rideshare &              TB &            GPI &             WFH &            WFHO \\
\midrule
 \multicolumn{9}{l}{\textbf{(Sensitive Attribute = Race)}}  
\\
Main: logit       &   0.121 (0.018) &   -0.228 (0.02) &    0.14 (0.016) &   0.047 (0.017) &   0.357 (0.012) &  0.303 (0.015) &   -0.004 (0.01) &  -0.012 (0.008) \\
Main: DNN         &  -0.027 (0.007) &   -0.02 (0.007) &   0.058 (0.013) &   0.002 (0.009) &   0.071 (0.007) &   0.056 (0.01) &  -0.011 (0.005) &   -0.01 (0.005) \\
2 layers DNN      &  -0.024 (0.013) &  -0.032 (0.007) &   0.053 (0.017) &  -0.003 (0.008) &    0.108 (0.01) &  0.091 (0.015) &  -0.019 (0.005) &   -0.02 (0.005) \\
4 layers DNN      &  -0.022 (0.008) &   -0.013 (0.01) &   0.053 (0.009) &   -0.002 (0.01) &   0.046 (0.008) &  0.036 (0.011) &  -0.007 (0.006) &  -0.012 (0.005) \\
Batch size=5000 &  -0.033 (0.016) &  -0.046 (0.009) &   0.041 (0.005) &   0.011 (0.019) &   0.041 (0.009) &  0.048 (0.006) &  -0.009 (0.004) &  -0.035 (0.006) \\
He initialization &   -0.025 (0.01) &  -0.024 (0.008) &   0.053 (0.022) &  -0.001 (0.008) &   0.066 (0.008) &  0.057 (0.004) &  -0.011 (0.004) &  -0.014 (0.007) \\
 \multicolumn{9}{l}{\textbf{(Sensitive Attribute = Gender)}}  
\\
Main: logit       &   0.127 (0.026) &  -0.014 (0.008) &  -0.004 (0.011) &   0.022 (0.032) &     0.02 (0.01) &  0.047 (0.006) &   0.029 (0.006) &  -0.074 (0.008) \\
Main: DNN         &    0.017 (0.01) &  -0.002 (0.006) &  -0.001 (0.011) &    0.01 (0.009) &   0.008 (0.004) &  0.015 (0.007) &  -0.001 (0.006) &  -0.022 (0.005) \\
2 layers DNN      &   0.033 (0.015) &   -0.003 (0.01) &  -0.004 (0.003) &    0.011 (0.01) &   0.012 (0.005) &  0.018 (0.007) &   0.003 (0.007) &  -0.029 (0.007) \\
4 layers DNN      &   0.009 (0.013) &  -0.002 (0.006) &  -0.003 (0.005) &   0.008 (0.009) &   0.007 (0.004) &  0.011 (0.004) &  -0.003 (0.008) &  -0.016 (0.004) \\
Batch size=5000 &   0.031 (0.013) &  -0.007 (0.007) &   0.002 (0.008) &   0.016 (0.015) &  -0.001 (0.003) &  0.006 (0.004) &  -0.004 (0.004) &  -0.063 (0.008) \\
He initialization &   0.015 (0.008) &  -0.004 (0.008) &   -0.003 (0.01) &   0.011 (0.008) &   0.012 (0.003) &  0.016 (0.003) &  -0.002 (0.006) &  -0.021 (0.004) \\
 \multicolumn{9}{l}{\textbf{(Sensitive Attribute = Income)}}  \\
Main: logit       &   -0.09 (0.039) &  -0.216 (0.009) &   0.353 (0.022) &    0.25 (0.023) &   0.605 (0.013) &  0.564 (0.003) &   0.053 (0.014) &  -0.139 (0.011) \\
Main: DNN         &  -0.051 (0.009) &  -0.019 (0.009) &   0.133 (0.018) &   0.003 (0.011) &   0.101 (0.011) &  0.084 (0.016) &  -0.006 (0.011) &  -0.049 (0.007) \\
2 layers DNN      &  -0.076 (0.009) &   -0.03 (0.006) &    0.13 (0.012) &   0.018 (0.019) &   0.165 (0.021) &    0.15 (0.02) &  -0.012 (0.011) &  -0.068 (0.009) \\
4 layers DNN      &  -0.031 (0.004) &   -0.014 (0.01) &   0.127 (0.017) &   -0.007 (0.01) &   0.072 (0.005) &  0.054 (0.007) &  -0.002 (0.014) &  -0.036 (0.007) \\
Batch size=5000 &  -0.087 (0.012) &  -0.044 (0.014) &   0.092 (0.018) &   0.037 (0.034) &    0.042 (0.01) &  0.046 (0.011) &  -0.001 (0.012) &  -0.119 (0.016) \\
He initialization &  -0.041 (0.005) &   -0.02 (0.008) &   0.114 (0.028) &   0.002 (0.006) &   0.089 (0.007) &  0.076 (0.009) &  -0.006 (0.007) &  -0.046 (0.005) \\

\bottomrule
\end{tabular}
\begin{tablenotes}
      \small
      \item Note: \\1. means and standard deviations (in parentheses) of the FNR gap/FPR gap and accuracy of 5-fold cross-validation trained with different models.\\ 2. "Main: logit" is the BLR model, "Main: DNN" is the default DNN model tested in Section \ref{nhts_exp}. For this default DNN model, the number of hidden layers is 3, and it is trained with a batch size of 500,000 and the Xavier uniform weight initialization. The other four models are adapted from the default DNN model: "2 layers DNN" and "4 layers DNN" change the number of DNN hidden layers to 2 and 4 while keeping other hyperparameters fixed. "Batch size=5000" changes the batch size to 5000. "He initialization" change the weight initializer to He normal initializer.
    \end{tablenotes}
\caption{Fairness and accuracy results of different models}
\label{sensitivity1}
\end{threeparttable}
\end{table}

\begin{table}[hbt!]
    \centering
    \begin{threeparttable}
\begin{tabular}{c|cccc|cccc}
\toprule
Model & \multicolumn{4}{c|}{FNR gap} & \multicolumn{4}{c}{FPR gap} \\
\cline{2-9}
{} &            Bike &             Bus &             Car &       Rideshare &              TB &            GPI &             WFH &            WFHO \\
\midrule
 \multicolumn{9}{l}{\textbf{(Sensitive Attribute = Medical Condition)}}  \\
Main: logit       &  -0.006 (0.054) &   -0.14 (0.023) &   0.253 (0.044) &   0.433 (0.055) &    0.425 (0.02) &  0.321 (0.025) &   0.137 (0.057) &  -0.106 (0.009) \\
Main: DNN         &   -0.02 (0.026) &  -0.016 (0.006) &   0.091 (0.027) &   0.043 (0.025) &   0.117 (0.014) &  0.083 (0.012) &   0.065 (0.042) &  -0.028 (0.014) \\
2 layers DNN      &  -0.026 (0.024) &  -0.029 (0.005) &   0.089 (0.033) &   0.061 (0.026) &    0.154 (0.02) &   0.106 (0.01) &   0.053 (0.023) &   -0.023 (0.01) \\
4 layers DNN      &  -0.011 (0.023) &  -0.018 (0.005) &    0.075 (0.01) &    0.022 (0.02) &    0.086 (0.01) &   0.064 (0.01) &    0.065 (0.04) &   -0.017 (0.02) \\
Batch size=5000 &   -0.01 (0.034) &  -0.037 (0.006) &   0.057 (0.019) &   0.116 (0.069) &   0.055 (0.012) &   0.05 (0.008) &    0.05 (0.031) &  -0.076 (0.018) \\
He initialization &  -0.016 (0.031) &   -0.017 (0.01) &   0.082 (0.027) &   0.042 (0.026) &   0.115 (0.022) &  0.075 (0.013) &    0.057 (0.03) &  -0.025 (0.009) \\
 \multicolumn{9}{l}{\textbf{(Sensitive Attribute = Region)}}  \\
Main: logit       &   0.356 (0.016) &   0.695 (0.046) &  -0.068 (0.012) &   0.624 (0.018) &    0.073 (0.01) &  0.244 (0.008) &   0.068 (0.011) &  -0.073 (0.017) \\
Main: DNN         &   0.076 (0.023) &   0.116 (0.027) &  -0.026 (0.014) &   0.146 (0.028) &   0.041 (0.005) &   0.082 (0.01) &   0.009 (0.008) &   -0.021 (0.01) \\
2 layers DNN      &   0.088 (0.027) &   0.168 (0.026) &  -0.018 (0.009) &    0.222 (0.04) &   0.053 (0.004) &  0.112 (0.013) &   0.018 (0.009) &   -0.03 (0.013) \\
4 layers DNN      &   0.053 (0.012) &   0.079 (0.033) &  -0.027 (0.007) &    0.089 (0.02) &   0.037 (0.008) &  0.063 (0.004) &    0.01 (0.005) &  -0.018 (0.005) \\
Batch size=5000 &   0.112 (0.047) &   0.173 (0.027) &  -0.022 (0.006) &   0.391 (0.056) &   0.026 (0.015) &   0.05 (0.017) &   0.009 (0.005) &  -0.039 (0.013) \\
He initialization &    0.073 (0.01) &    0.13 (0.039) &  -0.023 (0.012) &    0.136 (0.03) &   0.038 (0.007) &  0.079 (0.008) &   0.017 (0.006) &  -0.023 (0.003) \\
\toprule
Model & \multicolumn{8}{c}{Accuracy} \\
\cline{2-9}
{} &            Bike &             Bus &             Car &       Rideshare &              TB &            GPI &             WFH &            WFHO \\
\midrule

Main: logit       &   0.697 (0.006) &   0.778 (0.007) &    0.84 (0.004) &   0.777 (0.008) &   0.645 (0.002) &  0.652 (0.003) &   0.763 (0.003) &   0.791 (0.003) \\
Main: DNN         &   0.955 (0.002) &   0.977 (0.001) &    0.98 (0.001) &   0.946 (0.001) &   0.854 (0.004) &  0.856 (0.002) &   0.952 (0.001) &   0.944 (0.001) \\
2 layers DNN      &   0.928 (0.001) &   0.964 (0.001) &   0.968 (0.001) &   0.923 (0.001) &   0.795 (0.003) &  0.796 (0.002) &   0.928 (0.002) &   0.919 (0.002) \\
4 layers DNN      &   0.971 (0.001) &  0.984 (0.0005) &   0.986 (0.001) &   0.966 (0.001) &   0.893 (0.003) &  0.893 (0.003) &   0.971 (0.001) &   0.966 (0.001) \\
Batch size=5000 &   0.949 (0.004) &   0.965 (0.002) &   0.969 (0.001) &   0.866 (0.007) &   0.896 (0.004) &  0.897 (0.004) &    0.94 (0.007) &   0.867 (0.007) \\
He initialization &   0.959 (0.001) &   0.978 (0.001) &  0.979 (0.0004) &   0.952 (0.001) &   0.871 (0.004) &  0.867 (0.005) &   0.959 (0.002) &   0.951 (0.001) \\
\bottomrule
\end{tabular}

\caption{(Cont.) Fairness and accuracy results of different models}
\label{sensitivity2}
\end{threeparttable}
\end{table}

\end{landscape}
\restoregeometry
\newpage
\section{Other Fairness Measures after Bias Mitigation}
\label{sec:other_metric}

Figure \ref{NHTS_mitigate_mode_FPR} indicates that when mitigating the absolute value of the FNR gap in terms of ``region'' for the frequent rideshare usage prediction, the absolute value of the FPR gap also decreases (i.e., the value becomes closer to zero). Similarly, Figure \ref{NHTS_mitigate_WFH_FNR} shows that when mitigating the absolute value of the FPR gap in terms of "race" for the travel burden prediction, the absolute value of the FNR gap also decreases.\\

\noindent In terms of the F1-score\footnote{The F1-score conveys the balance between recall and precision. The F1-score gap is calculated as: F1-score gap=$    \frac{TP_{z=0}}{TP_{z=0}+0.5*(FP_{z=0}+FN_{z=0})}-\frac{TP_{z=1}}{TP_{z=1}+0.5*(FP_{z=1}+FN_{z=1})}$.}, the results show that the bias mitigation also leads to drops in the absolute F1-score gaps in Figure \ref{NHTS_mitigate_WFH_F1}. However, in Figure \ref{NHTS_mitigate_mode_F1}, the absolute F1-score gap only drops when BLR is applied, but not for DNN.

\setcounter{figure}{0}
\renewcommand{\thefigure}{E\arabic{figure}}

\begin{figure}[h!]
\begin{subfigure}[b]{.32\linewidth}
\centering
\end{subfigure}\hfill
\begin{subfigure}[b]{.4\linewidth}
\centering
\includegraphics[width=\linewidth]{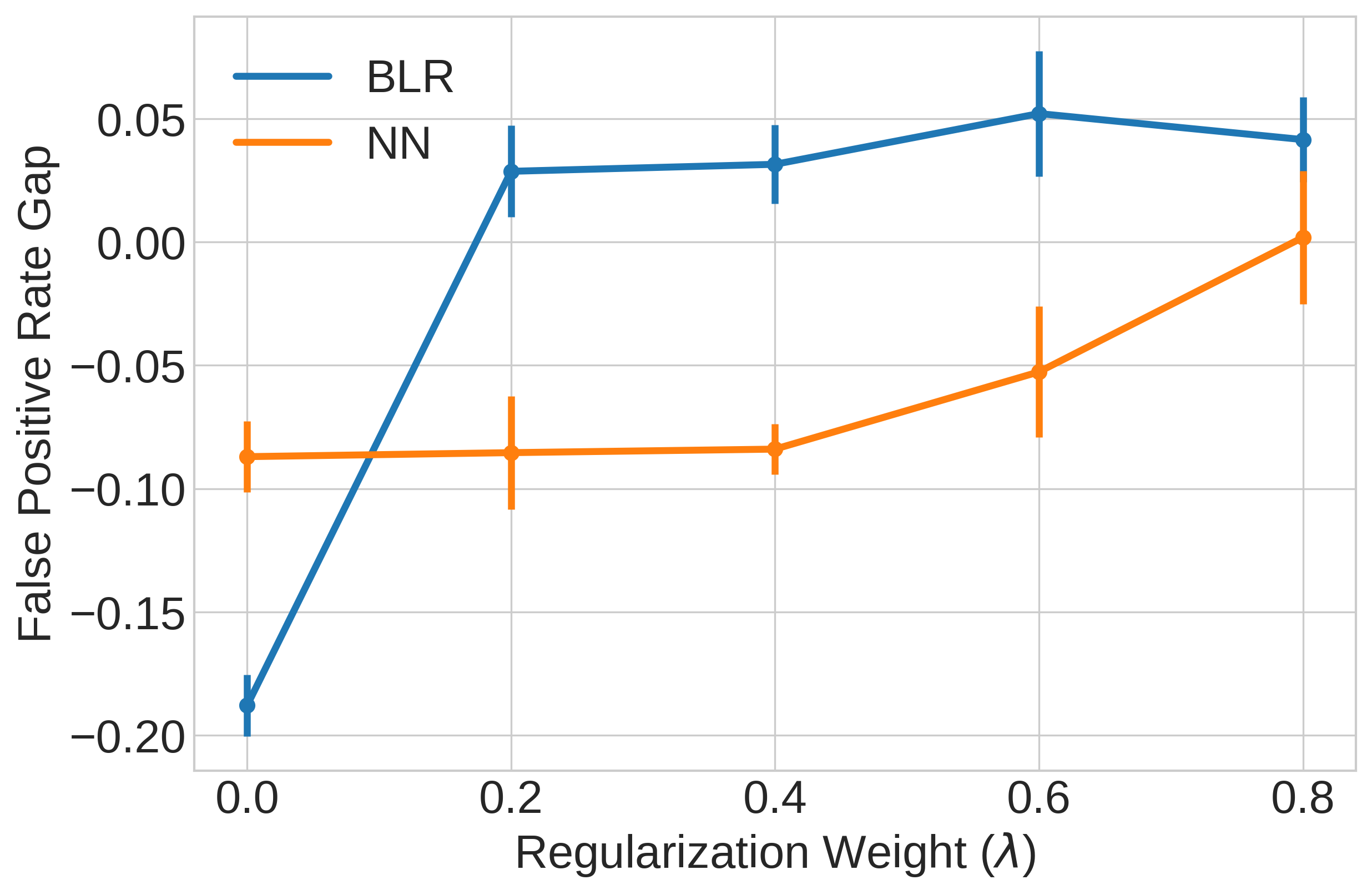}
\caption{FPR Gap vs. Regularization Weight for FNR}\label{NHTS_mitigate_mode_FPR}
\end{subfigure}\hfill
\begin{subfigure}[b]{.4\linewidth}
\centering
\includegraphics[width=\linewidth]{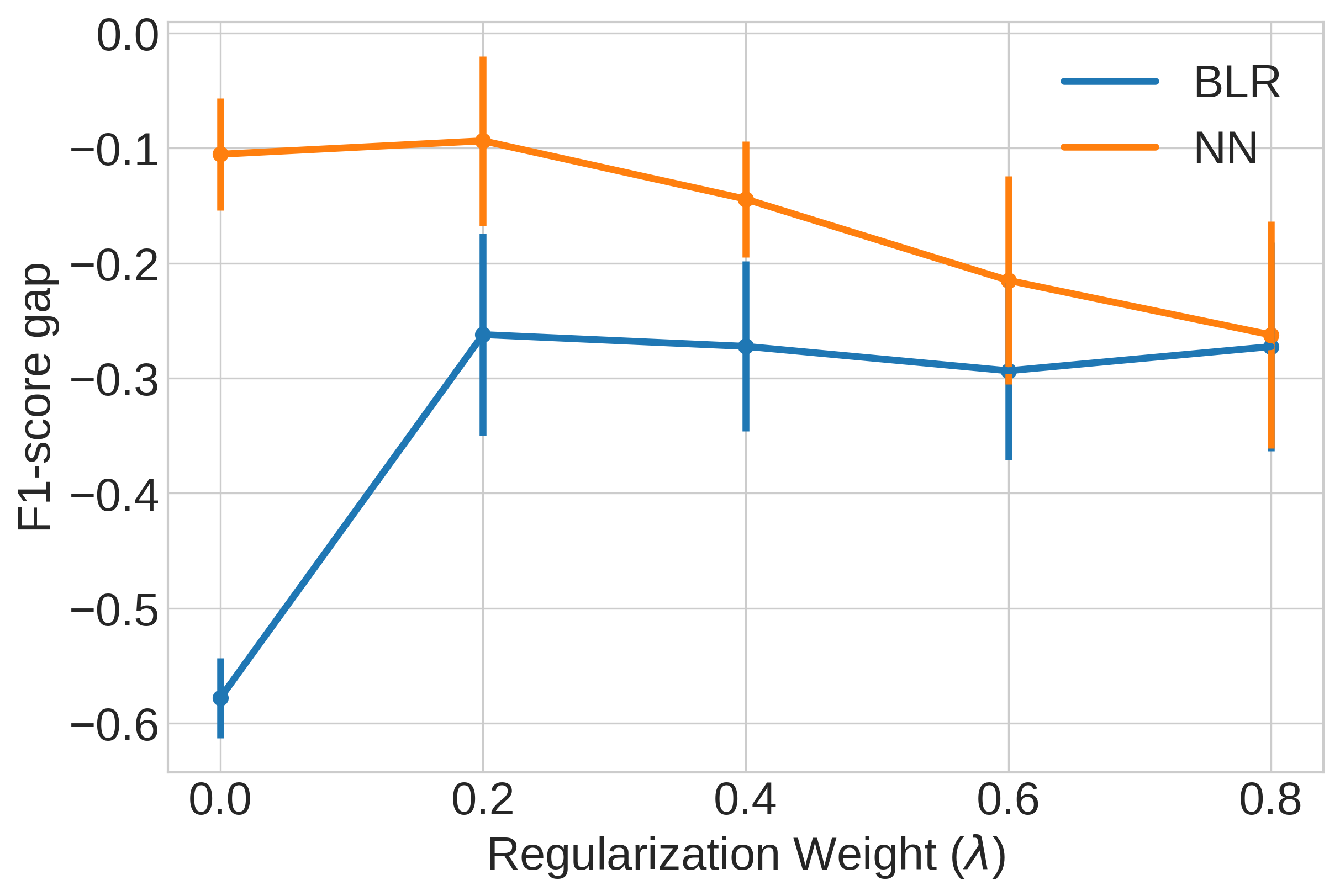}
\caption{F1-score Gap vs. Regularization Weight for FNR}\label{NHTS_mitigate_mode_F1}
\end{subfigure}\hfill
\begin{subfigure}[b]{.32\linewidth}
\centering
\end{subfigure}\hfill

\caption{FPR gap and F1-score gap by bias mitigation weight ($\lambda$) for FNR: regional gap in the prediction of frequent rideshare usage}
\label{NHTS_mitigate_mode_appendix}
\end{figure}

\begin{figure}[h!]
\begin{subfigure}[b]{.32\linewidth}
\centering
\end{subfigure}\hfill
\begin{subfigure}[b]{.4\linewidth}
\centering
\includegraphics[width=\linewidth]{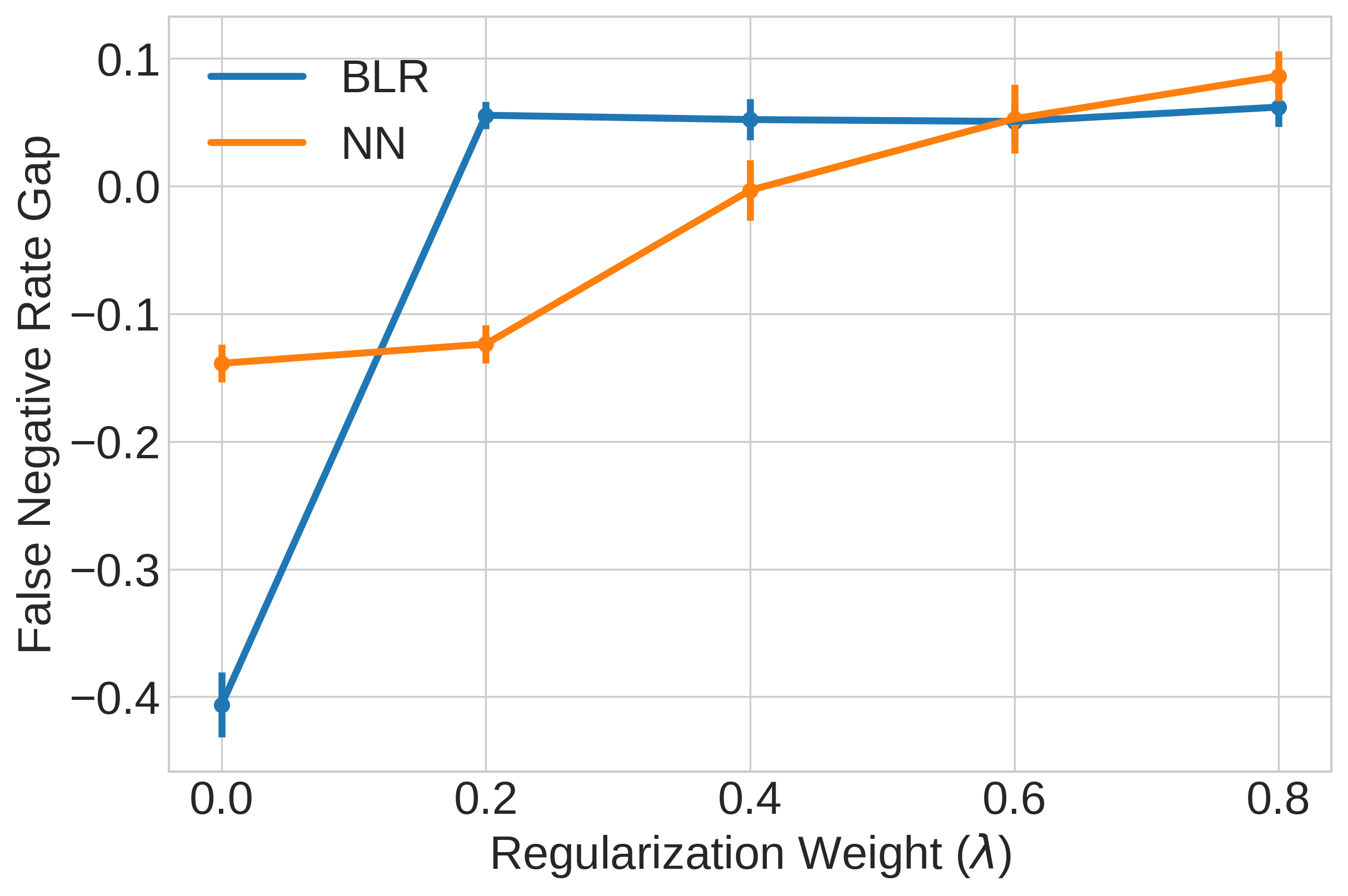}
\caption{FNR gap vs. Regularization Weight for FPR}\label{NHTS_mitigate_WFH_FNR}
\end{subfigure}\hfill
\begin{subfigure}[b]{.4\linewidth}
\centering
\includegraphics[width=\linewidth]{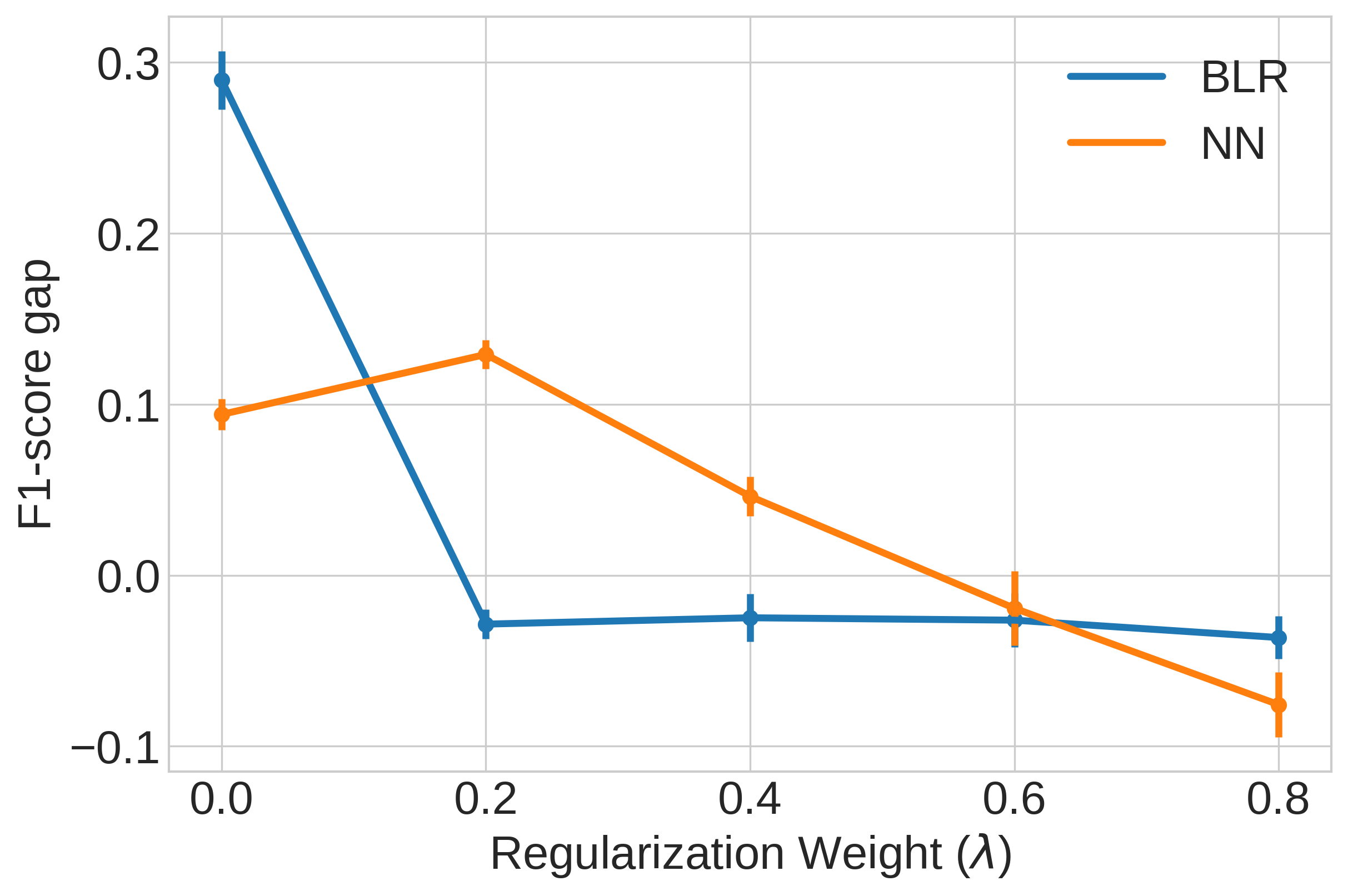}
\caption{F1-score gap vs. Regularization Weight for FPR}\label{NHTS_mitigate_WFH_F1}
\end{subfigure}\hfill
\begin{subfigure}[b]{.32\linewidth}
\centering
\end{subfigure}\hfill

\caption{FNR gap and F1-score gap by bias mitigation weight ($\lambda$) for FPR: racial gap in the prediction of travel burden}

\label{NHTS_mitigate_WFH_add}
\end{figure}

\newpage
\section{Effects of Bias Mitigation on Other Protected Variables}
\label{other_var}
\setcounter{table}{0}
\renewcommand{\thetable}{F\arabic{table}}

\begin{table}[hbt!]
    \centering
    \begin{threeparttable}
\resizebox{\textwidth}{!}{\begin{tabular}{c|ccccc|c}
\toprule
$\lambda$ & \multicolumn{5}{c|}{FNR gap} &       Accuracy \\
\cline{2-6}
{} &           Race &         Gender &        Income &        Medical Condition & \textbf{Region}& \\
\midrule
0.0 &  0.047 (0.017) &  0.022 (0.032) &   0.25 (0.023) &  0.433 (0.055) &   \textbf{0.624 (0.018)} &  0.777 (0.008) \\
0.2 &  0.069 (0.025) &  0.049 (0.035) &  0.295 (0.024) &   0.47 (0.055) &  \textbf{-0.064 (0.048)} &  0.766 (0.001) \\
0.4 &   0.07 (0.025) &  0.047 (0.035) &  0.285 (0.037) &  0.471 (0.056) &   \textbf{-0.047 (0.04)} &  0.766 (0.001) \\
0.6 &   0.077 (0.02) &  0.055 (0.036) &  0.325 (0.027) &  0.488 (0.045) &  \textbf{-0.044 (0.047)} &  0.764 (0.002) \\
0.8 &    0.11 (0.02) &  0.066 (0.033) &  0.373 (0.028) &   0.48 (0.047) &  \textbf{-0.061 (0.054)} &  0.759 (0.002)  \\
\bottomrule
\end{tabular}}
\caption{Change of FNR gaps for various protected variables with the increase of bias mitigation weight ($\lambda$) regarding "region" when predicting the frequent rideshare usage BLR.}
\label{effect_mitigate_mode_BLR}
\end{threeparttable}
\end{table}

\begin{table}[hbt!]
    \centering
    \begin{threeparttable}
\resizebox{\textwidth}{!}{\begin{tabular}{c|ccccc|c}
\toprule
$\lambda$ & \multicolumn{5}{c|}{FPR gap} &       Accuracy \\
\cline{2-6}
{} &           \textbf{Race} &         Gender &        Income &        Medical Condition & Region& \\

\midrule
0.0 &    \textbf{0.357 (0.012)} &     0.02 (0.01) &  0.605 (0.013) &   0.425 (0.02) &   0.073 (0.01) &  0.645 (0.002) \\
0.2 &  \textbf{-0.0002 (0.012)} &    0.01 (0.008) &  0.561 (0.022) &   0.43 (0.019) &   0.138 (0.01) &  0.634 (0.001) \\
0.4 &    \textbf{0.001 (0.012)} &    0.006 (0.01) &  0.552 (0.019) &  0.434 (0.018) &  0.142 (0.015) &  0.634 (0.001) \\
0.6 &   \textbf{-0.001 (0.013)} &   0.002 (0.007) &  0.514 (0.069) &  0.448 (0.028) &   0.168 (0.03) &  0.632 (0.001) \\
0.8 &   \textbf{-0.002 (0.011)} &  -0.004 (0.014) &  0.424 (0.113) &  0.437 (0.024) &  0.224 (0.077) &  0.627 (0.006) \\
\bottomrule
\end{tabular}}

\caption{Change of FPR gaps for various protected variables with the increase of bias mitigation weight ($\lambda$) regarding "race" when predicting the travel burden using BLR.}
\label{effect_mitigate_wfh_BLR}
\end{threeparttable}
\end{table}

\begin{table}[hbt!]
    \centering
    \begin{threeparttable}
\resizebox{\textwidth}{!}{\begin{tabular}{c|ccccc|c}
\toprule
$\lambda$ & \multicolumn{5}{c|}{FNR gap} &       Accuracy \\
\cline{2-6}
{} &           Race &         Gender &        Income &        Medical Condition & \textbf{Region}& \\

\midrule
0.0 &  0.002 (0.009) &   0.01 (0.009) &  0.003 (0.011) &  0.043 (0.025) &   \textbf{0.146 (0.028)} &  0.946 (0.001) \\
0.2 &  0.008 (0.008) &   0.015 (0.01) &  0.029 (0.011) &  0.048 (0.029) &   \textbf{0.048 (0.032)} &  0.906 (0.002) \\
0.4 &  0.026 (0.017) &    0.03 (0.01) &  0.083 (0.035) &  0.163 (0.039) &   \textbf{0.062 (0.031)} &  0.862 (0.003) \\
0.6 &  0.056 (0.019) &  0.044 (0.028) &  0.231 (0.038) &  0.308 (0.061) &   \textbf{0.038 (0.051)} &   0.81 (0.003) \\
0.8 &  0.071 (0.021) &  0.047 (0.038) &   0.288 (0.03) &  0.494 (0.069) &  \textbf{-0.009 (0.058)} &  0.778 (0.001) \\
\bottomrule
\end{tabular}}
\caption{Change of FNR gaps for various protected variables with the increase of bias mitigation weight ($\lambda$) regarding "region" when predicting the frequent rideshare usage using 3-layer DNN.}
\label{effect_mitigate_mode_DNN}
\end{threeparttable}
\end{table}

\begin{table}[hbt!]
    \centering
    \begin{threeparttable}
\resizebox{\textwidth}{!}{\begin{tabular}{c|ccccc|c}
\toprule
$\lambda$ & \multicolumn{5}{c|}{FPR gap} &       Accuracy \\
\cline{2-6}
{} &           \textbf{Race} &         Gender &        Income &        Medical Condition & Region& \\

\midrule
0.0 &   \textbf{0.071 (0.007)} &  0.008 (0.004) &  0.101 (0.011) &  0.117 (0.014) &  0.041 (0.005) &  0.854 (0.004) \\
0.2 &   \textbf{0.025 (0.007)} &  0.013 (0.005) &  0.234 (0.014) &  0.202 (0.016) &  0.086 (0.011) &  0.746 (0.007) \\
0.4 &  \textbf{-0.004 (0.021)} &  0.007 (0.011) &  0.469 (0.029) &  0.352 (0.014) &  0.148 (0.014) &  0.664 (0.003) \\
0.6 &  \textbf{-0.003 (0.02)} &  0.012 (0.011) &  0.561 (0.016) &  0.381 (0.031) &  0.168 (0.012) &  0.641 (0.001) \\
0.8 &  \textbf{-0.014 (0.015)} &  0.008 (0.009) &  0.459 (0.042) &   0.328 (0.02) &     0.2 (0.02) &  0.633 (0.001) \\
\bottomrule
\end{tabular}}

\caption{Change of FPR gaps for various protected variables with the increase of bias mitigation weight ($\lambda$) regarding "race" when predicting the travel burden using 3-layer DNN.}
\label{effect_mitigate_wfh_DNN}
\end{threeparttable}
\end{table}

\newpage
\section{Descriptive Statistics (the Chicago Travel Survey)}
\label{sec:descriptive_chicago}
\setcounter{table}{0}
\renewcommand{\thetable}{G\arabic{table}}
\begin{table}[hbt!]
    \centering
    \begin{threeparttable}
\begin{tabular}{lrrrrr}
\toprule
{} &    Mean &    Std. &   Min &  Median &     Max \\
\midrule
\textbf{\textit{Independent Variables:}}                                              &   &  &  &   &    \\
*Ethnic minority (Yes=1)                 &   0.221 &   0.415 &   0.0 &     0.0 &     1.0 \\
*Low income household (Yes=1)            &   0.056 &   0.230 &   0.0 &     0.0 &     1.0 \\
*Disability status (Yes=1)               &   0.024 &   0.152 &   0.0 &     0.0 &     1.0 \\
Age                                     &  41.712 &  14.247 &  16.0 &    40.0 &    87.0 \\
Education level (1-6: low-high)         &   4.410 &   1.480 &   1.0 &     5.0 &     6.0 \\
Number of jobs                          &   1.117 &   0.395 &   1.0 &     1.0 &     9.0 \\
Have a valid driver's license (Yes=1)   &   0.944 &   0.230 &   0.0 &     1.0 &     1.0 \\
Gender (female=1)                       &   0.529 &   0.499 &   0.0 &     1.0 &     1.0 \\
Count of household members              &   3.155 &   1.481 &   1.0 &     3.0 &    12.0 \\
Count of household vehicles             &   2.052 &   1.201 &   0.0 &     2.0 &    10.0 \\
Home ownership (Yes=1)                  &   0.687 &   0.464 &   0.0 &     1.0 &     1.0 \\
Count of years lived at current address &  11.040 &  10.100 &   1.0 &     8.0 &   120.0 \\
Distance to the work place (km)         &  17.191 &  33.700 &   0.0 &    11.2 &  1272.2 \\
Last year's household income (K \$)      &  98.023 &  52.445 &   7.5 &    87.5 &   175.0 \\
\textbf{\textit{Dependent Variable:}}                                              &   &  &  &   &    \\
Travel mode to work (auto=1, else=0)    &   0.731 &   0.444 &   0.0 &     1.0 &     1.0 \\

\textbf{Sample size:} 13,730                                             &   &  &  &   &    \\
\bottomrule
\end{tabular}
\begin{tablenotes}
      \small
      \item Note: the sample weights are used to compute the summary statistics; (*) denotes the independent variables that are also treated 
as the protected variables based on which we examine the prediction disparities.
    \end{tablenotes}
\caption{Summary statistics of the explanatory and dependent variables in the Chicago travel survey dataset}
\label{descriptive_chicago}
\end{threeparttable}
\end{table}

\end{document}